\documentclass{article}

\usepackage{microtype}
\usepackage[preprint]{icml2026}

\usepackage{amssymb}
\usepackage{mathtools}
\usepackage{amsthm}
\usepackage{url}
\usepackage[dvipsnames]{xcolor}       % colors
\usepackage{colortbl}     
\usepackage{amsmath}
\usepackage[utf8]{inputenc} % allow utf-8 input
\usepackage[T1]{fontenc}    % use 8-bit T1 fonts
\usepackage{url}            % simple URL typesetting
\usepackage{amsfonts}       % blackboard math symbols
\usepackage{nicefrac}       % compact symbols for 1/2, etc.
\usepackage{longtable}
\usepackage{pifont}

\usepackage{caption}
\usepackage{textcomp}

\usepackage{booktabs}
\usepackage{multirow}
\usepackage{wrapfig}
\usepackage{graphicx}
\usepackage{enumitem}
\usepackage{soul}
\usepackage{amsmath,amscd,amsbsy,amssymb,amsfonts,latexsym,url,bm,amsthm}
\def\BibTeX{{\rm B\kern-.05em{\sc i\kern-.025em b}\kern-.08em
		T\kern-.1667em\lower.7ex\hbox{E}\kern-.125emX}}
\usepackage{threeparttable}
\usepackage{listings}
\usepackage{array}   
\usepackage{amssymb}% 
\usepackage{multirow}
\usepackage{multicol}
\usepackage{subfigure}
\usepackage{algpseudocode}% \theoremstyle{definition}
\definecolor{customblue}{HTML}{2A52BE} 
\usepackage[colorlinks,
            linkcolor=customblue,
            anchorcolor=blue,
            citecolor=teal,
            ]{hyperref}
\usepackage{mathtools}
\usepackage{thmtools}
\usepackage{thm-restate}
\usepackage{xspace}
 
\definecolor{codegreen}{rgb}{0,0.6,0}
\definecolor{codegray}{rgb}{0.5,0.5,0.5}
\definecolor{codepurple}{rgb}{0.58,0,0.82}
\definecolor{backcolour}{rgb}{0.98,0.98,0.98}
\definecolor{promptblue}{rgb}{0.01,0.28,0.48}
\definecolor{promptbg}{rgb}{0.97, 0.97, 1.0}
\definecolor{placeholderred}{rgb}{0.8, 0.2, 0.2}
\usepackage{etoolbox} % Required for \gappto
\usepackage{thm-restate}
\gappto{\appendix}{%
  \addtocontents{toc}{\protect\setcounter{tocdepth}{1}}% Optional: Adjust ToC depth
}
\lstdefinestyle{promptstyle}{
    basicstyle=\small\ttfamily\color{promptblue},
    frame=tb, % Add top and bottom frame lines
    framesep=8pt,
    framerule=0.4pt,
    breaklines=true,
    breakatwhitespace=true,
    postbreak=\mbox{\textcolor{gray}{$\hookrightarrow$}\space},
    backgroundcolor=\color{promptbg},
    keywordstyle=\color{black}\bfseries,
    literate=
        {\{}{{\textcolor{placeholderred}{\{}}}{1}
        {\}}{{\textcolor{placeholderred}{\}}}}{1},
    morekeywords={You, Your, Now, Once, Prior, Below, Here}, % Keywords to highlight
}
\lstdefinestyle{mystyle}{
    backgroundcolor=\color{backcolour},   
    commentstyle=\color{codegreen},
    keywordstyle=\color{magenta},
    numberstyle=\tiny\color{codegray},
    stringstyle=\color{codepurple},
    basicstyle=\footnotesize\ttfamily,
    breakatwhitespace=false,         
    breaklines=true,                 
    captionpos=b,                    
    keepspaces=true,                 
    numbers=left,                    
    numbersep=5pt,                  
    showspaces=false,                
    showstringspaces=false,
    showtabs=false,                  
    tabsize=2,
    language=Python % 指定语言
}
\definecolor{mygray}{gray}{0.9}  % Define the mygray color (adjust the 0.9 value as needed)
\usepackage{arydshln}
\definecolor{first}{RGB}{178,24,43}
\definecolor{second}{RGB}{117,112,179}
\definecolor{third}{RGB}{189,189,189}
\newcommand{\thickhline}{\hline\hline}  % Define thickhline as double hline
\newcommand{\alfworld}{ALFWorld}
\newcommand{\sciworld}{ScienceWorld}
\newcommand{\react}{ReAct}
\newcommand{\agentgym}{AgentGym}
\newcommand{\gigpo}{GiGPO}
\newcommand{\rlvmr}{RLVMR}
\newcommand{\ours}{\textsc{EPO}\xspace}
\newcommand{\cfirst}[1]{\textcolor{first}{\textbf{#1}}}
\newcommand{\csecond}[1]{\textcolor{second}{\textbf{#1}}}
\newcommand{\cthird}[1]{\textcolor{third}{\textbf{#1}}}
\usepackage[capitalize,noabbrev]{cleveref}

\theoremstyle{plain}
\newtheorem{theorem}{Theorem}[section]

\newtheorem{lemma}[theorem]{Lemma}
\newtheorem{corollary}[theorem]{Corollary}
\theoremstyle{definition}
\newtheorem{definition}[theorem]{Definition}
\newtheorem{assumption}[theorem]{Assumption}
\theoremstyle{remark}
\newtheorem{remark}[theorem]{Remark}

\usepackage[textsize=tiny]{todonotes}

\icmltitlerunning{EPO: Entropy-regularized Policy Optimization for LLM Agents Reinforcement Learning}

\newcommand{\cem}[1]{\cellcolor[gray]{0.9}{#1}}
\newcommand{\lightcem}[1]{\cellcolor[gray]{0.95}{#1}}
\newcommand{\update}[1]{\textcolor{black}{#1}}
\begin{document}

\twocolumn[
  \icmltitle{EPO: Entropy-regularized Policy Optimization for LLM Agents Reinforcement Learning}




  \begin{icmlauthorlist}
    \icmlauthor{Wujiang Xu}{ru}
    \icmlauthor{Wentian Zhao}{adobe}
    \icmlauthor{Zhenting Wang}{ru}
    \icmlauthor{Yu-Jhe Li}{adobe} \\
    \icmlauthor{Can Jin}{ru} 
    \icmlauthor{Mingyu Jin}{ru}
    \icmlauthor{Kai Mei}{ru}
    \icmlauthor{Kun Wan}{adobe}
    \icmlauthor{Dimitris N. Metaxas}{ru}
\end{icmlauthorlist}
\icmlaffiliation{ru}{Rutgers University, USA}
\icmlaffiliation{adobe}{Adobe Inc., USA}
\icmlcorrespondingauthor{Wujiang Xu}{wujiang.xu@rutgers.edu}

  \icmlkeywords{Machine Learning, ICML}

  \vskip 0.3in
]



\printAffiliationsAndNotice{}  % no special notice (required even if empty)

\begin{abstract}
Reinforcement learning has enabled LLMs to acquire reasoning through single-turn verified rewards, yet extending this paradigm to multi-turn agents—where tasks span 30+ turns with only terminal rewards—introduces fundamental challenges. We identify a critical failure mode in this setting: the exploration-exploitation cascade failure. Unlike single-turn training, multi-turn agents share policy parameters across all turns, meaning entropy adjustments cannot independently control early-turn exploration and late-turn exploitation. Combined with sparse rewards that provide no intermediate correction, this coupling causes severe entropy oscillations that destabilize training. We propose Entropy-regularized Policy Optimization (EPO), a framework that stabilizes entropy dynamics through three synergistic mechanisms: (1) trajectory-level entropy regularization that captures multi-turn temporal structure, (2) an entropy smoothing regularizer that penalizes deviations from historical averages to dampen oscillations, and (3) adaptive phase-based weighting that transitions from conservative exploration to strong stabilization across training. EPO achieves up to 152\% improvement on ScienceWorld and 19.8\% on ALFWorld, transforming previously unstable sparse-reward scenarios into smoothly converging optimization. Our work demonstrates that multi-turn settings require fundamentally different entropy control than traditional RL.
The code is available at the URL~\footnote{\url{https://github.com/WujiangXu/EPO}}.
\end{abstract}

\section{Introduction}

% 1. LLM has provide 
Reinforcement Learning (RL)~\citep{grpo,ppo} has become an important approach  for post-training Large Language Models (LLMs), enabling them to acquire reasoning ability through single-turn verified rewards~\citep{dpskr1,qwen3}. 
Extending this paradigm to multi-turn LLM agents, which interact with environments across extended episodes, has achieved promising results in coding~\citep{anthropic_claude_code,openai_codex}, computer use~\citep{oaicua,osworld}, and web search~\citep{kimik2,GLM4.5}, yet introduces fundamental challenges.
In multi-turn settings, episodes can span 30+ turns~\citep{alfworld,sciworld} yet rewards arrive only at task completion. This means early-turn actions shape entire trajectories without receiving corrective feedback, making exploration-exploitation balance~\citep{tdlearning} a critical challenge that single-turn methods handle implicitly through immediate rewards.

% \begin{figure*}[tb!]
% \centering{
% \includegraphics[width=0.60\textwidth]{figures/exper/ppo_sciworld_PPO_vs_+EPO_combined_entropy_reward_curves_early0-10_late20-30.png}}
% % \captionsetup{font={footnotesize}}
% \vspace{-10pt}
% \caption{The exploration-exploitation cascade failure observed in standard PPO. \update{We identify two distinct failure phases. During Phase 1 of Excessive Early Exploration (steps 0 to 40) the PPO early trajectory steps (pink dashed line) exhibit uncontrolled entropy growth and stagnant rewards. This triggers Phase 2 of Uncertainty Propagation (steps 40 to 120) where instability cascades to late trajectory steps (red dotted line) causing persistent high entropy and reward plateaus.} In contrast, our EPO method maintains stable, controlled entropy levels across both early and late trajectory steps throughout training, achieving significantly lower final entropy values and consistent reward improvement, preventing the cascade failure.}
% \label{fig:main_figure}
% \vspace{-18pt}
% \end{figure*}

\begin{figure*}[tb]
\centering
\subfigure[Early Steps]{
\begin{minipage}[t]{0.32\linewidth}
\centering
\includegraphics[width=\linewidth]{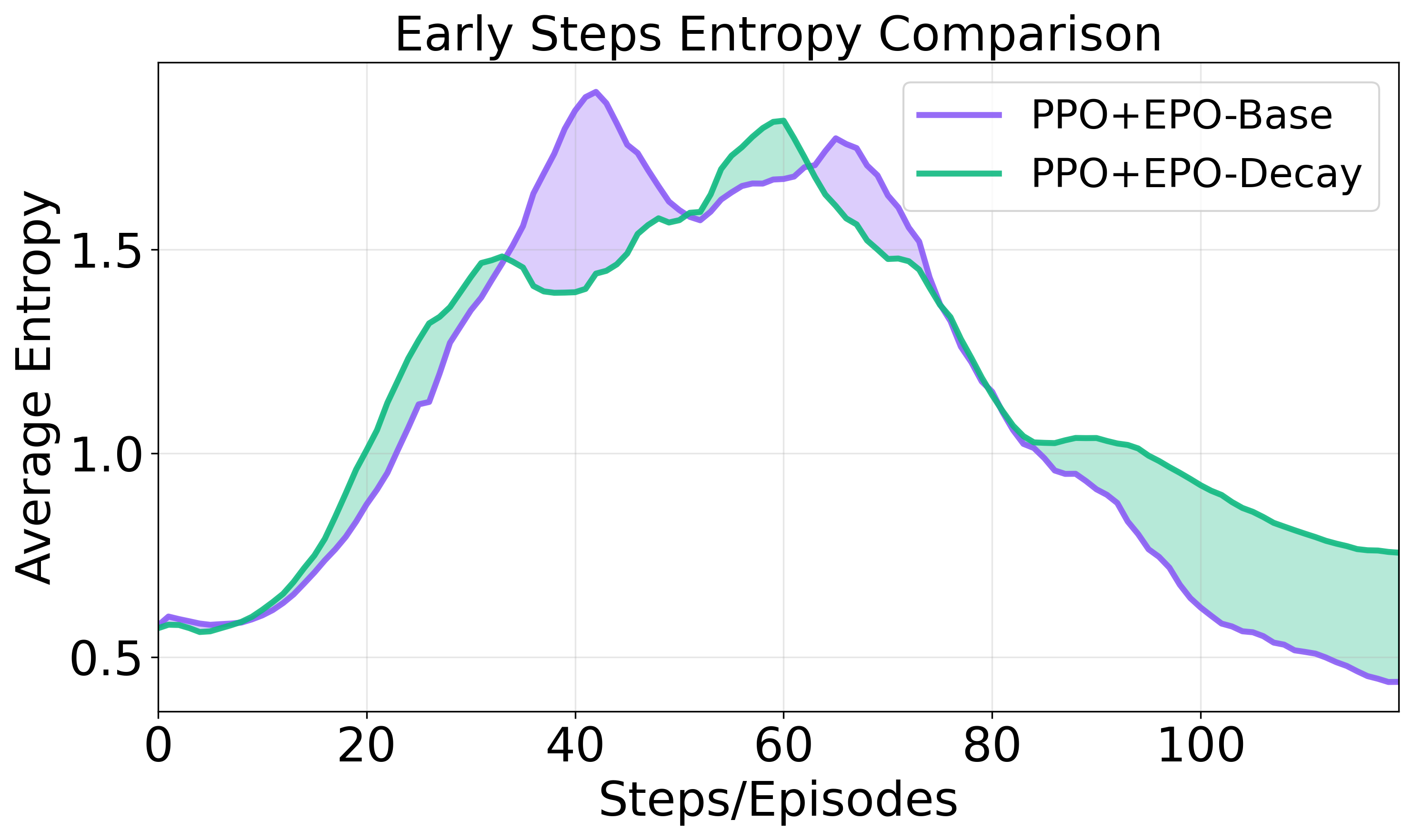}
\end{minipage}%
}%
\hfill
\subfigure[Late Steps]{
\begin{minipage}[t]{0.32\linewidth}
\centering
\includegraphics[width=\linewidth]{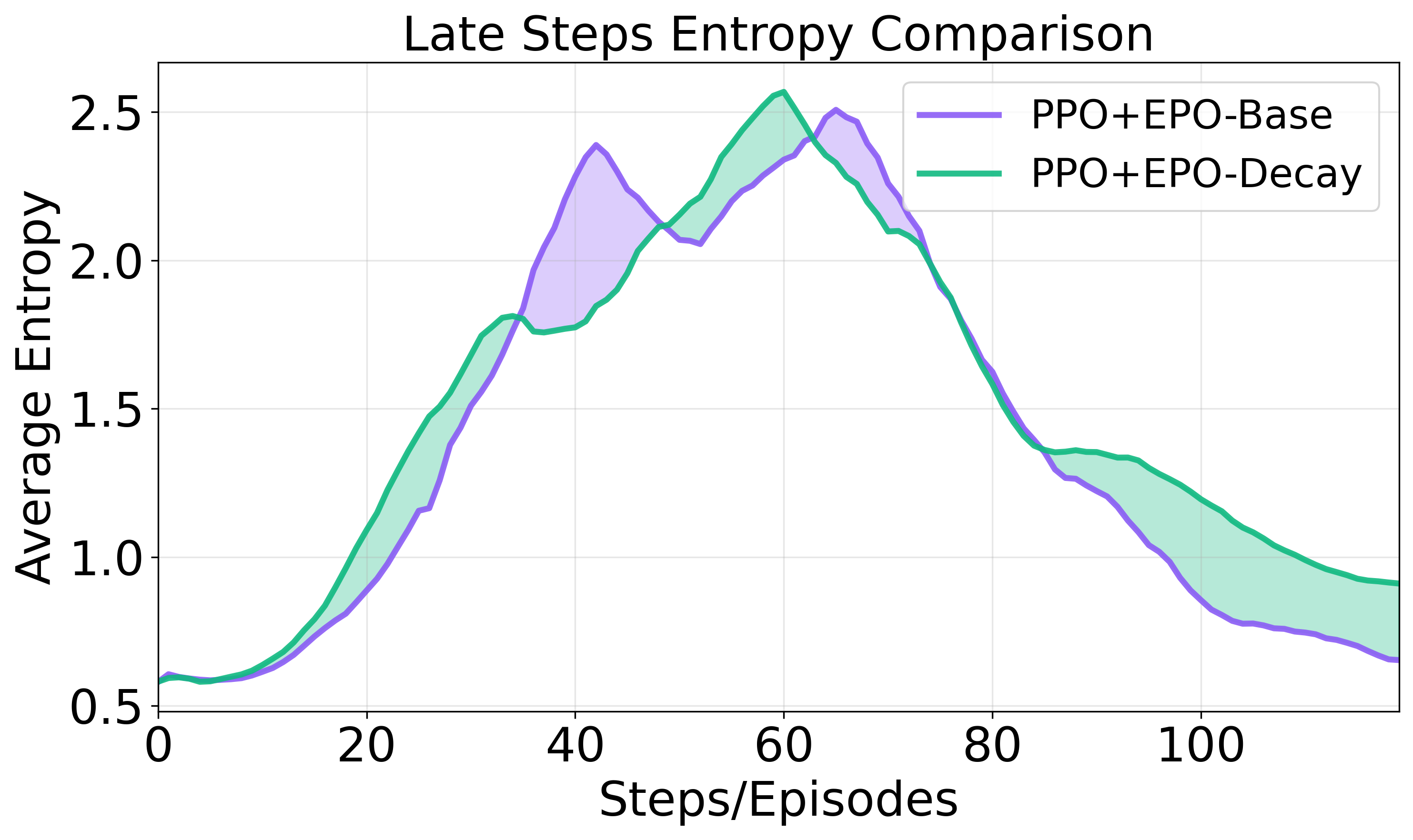}
\end{minipage}%
}%
\hfill
\subfigure[PPO vs PPO+EPO]{
\begin{minipage}[t]{0.32\linewidth}
\centering
\includegraphics[width=\linewidth]{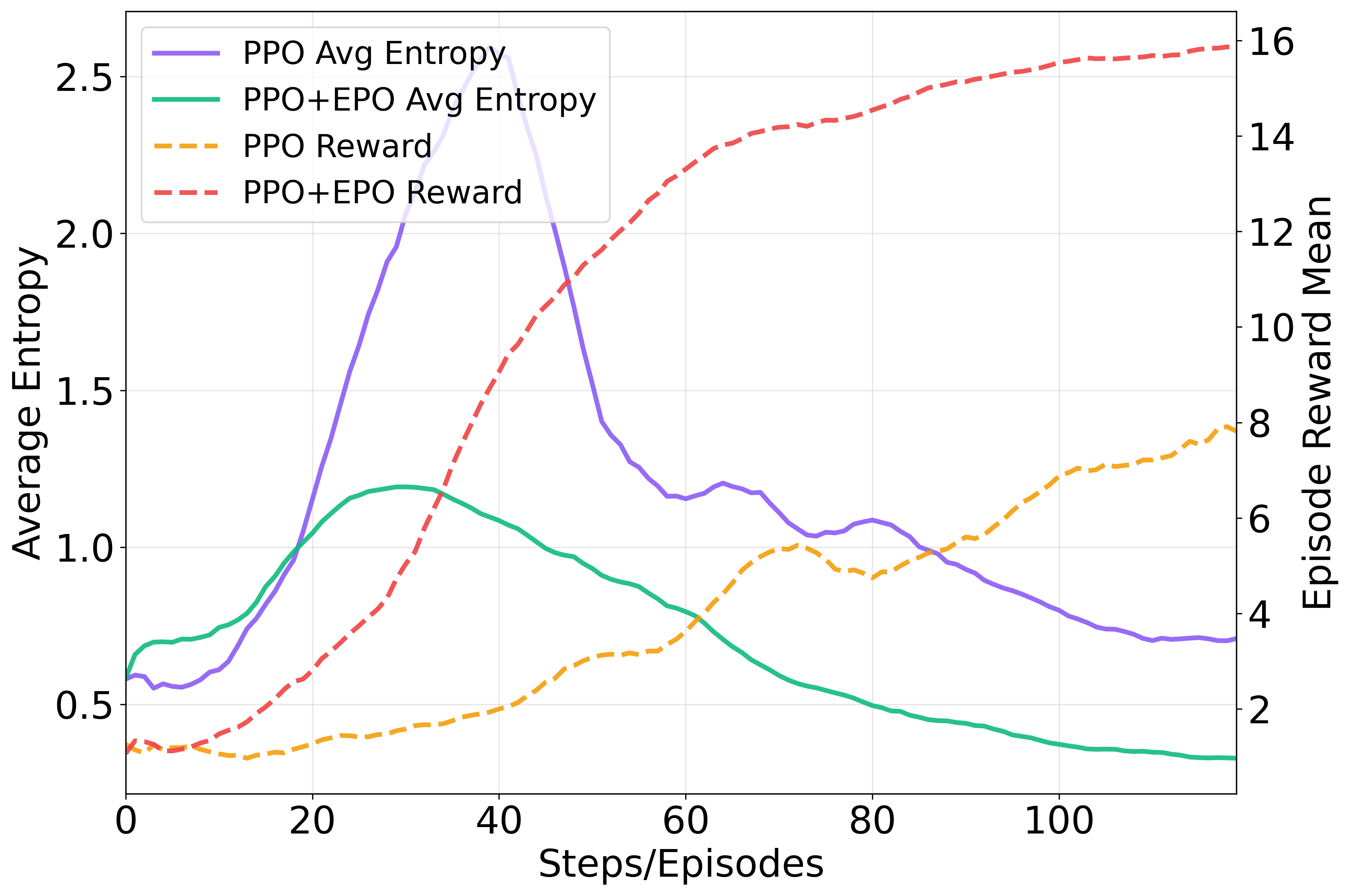}
\end{minipage}%
}%
\vspace{-10pt}
\caption{The exploration-exploitation cascade failure in multi-turn training. (a-b) Adjusting per-turn entropy weights fails due to shared parameters: EPO-Decay applies higher weights to early turns and lower weights to late turns, yet entropy curves remain nearly identical for both early steps (a) and late steps (b). (c) Standard PPO exhibits severe entropy fluctuation (purple) with stagnant rewards (yellow), while our proposed EPO maintains stable entropy (green, declining from 1.2 to 0.3) and achieves consistent reward improvement (red).}
\label{fig:intro}
\vspace{-15pt}
\end{figure*}

Traditional RL approaches~\citep{sac,a3c,91rlentropy} employ entropy regularization to address exploration, adding the policy's entropy to the objective function to discourage premature convergence to deterministic policies. In LLM training, researchers have adapted these mechanisms to reward high-entropy token generation~\citep{clip-conv,Rl-plus,skywork,8020entropy}, operating under the assumption that entropy collapse is the primary failure mode. 
Through extensive empirical analysis on standard multi-turn agent benchmarks, we uncover a critical limitation in multi-turn agent training settings: the challenge is not entropy collapse but rather uncontrolled entropy dynamics. In multi-turn training, the same policy parameters control all turns. Since all turns share parameters, adjusting entropy weights per turn cannot independently control exploration and exploitation: increasing entropy for early turns inevitably affects late turns, and vice versa. Combined with sparse rewards that provide no intermediate correction signal, this makes training dynamics inherently unstable. We term this the \textit{exploration-exploitation cascade failure}.

% As illustrated in \autoref{fig:main_figure}, this phenomenon consistently manifests in two distinct phases on standard benchmarks: \update{In Phase 1 (0-40 steps), without reward signals to provide corrective gradients, standard entropy regularization causes \textit{excessive early-stage exploration} with uncontrolled entropy growth (from 0.5 to 2.5) and chaotic early-turn actions while rewards remain stagnant. In Phase 2 (40-120 steps), because late turns inherit states shaped by early chaos, instability propagates as \textit{late-stage uncertainty}: entropy oscillations persist (1.0-2.0), preventing coherent strategy formation and causing reward plateaus. In contrast, our EPO method maintains stable, controlled entropy levels (declining from 1.2 to 0.3) throughout training, achieving consistent reward improvement from 2 to 16.} \update{This cascade failure is not an artifact of specific synthetic settings, but a fundamental failure mode that emerges naturally when standard RL objectives interact with multi-turn sparse reward environments.} Therefore, \textit{how to design exploration mechanisms that prevent this cascade failure while maintaining necessary exploration} remains an important open problem for multi-turn agent training.
As illustrated in \autoref{fig:intro}, we empirically validate this limitation. In Figure (a) and (b), we attempt a natural solution: applying higher entropy weights to early turns to promote exploration, and lower weights to late turns to encourage exploitation (EPO-Decay). However, due to shared policy parameters, early-turn and late-turn entropy curves remain nearly identical to the unweighted baseline (EPO-Base). This confirms that per-turn weight adjustment cannot break the coupling between turns. Figure (c) shows the consequence: PPO exhibits severe entropy fluctuation with stagnant rewards, while our proposed EPO maintains stable declining entropy and consistent reward growth. This cascade failure exposes a mismatch between existing entropy methods and multi-turn structure. Standard entropy regularization operates statelessly, treating each gradient update in isolation. However, in multi-turn training, shared parameters mean that each gradient update affects all turns simultaneously, so entropy at specific step depends on the entire training history, not just the current update. This raises a critical challenge: \textit{how can we stabilize entropy dynamics over time when shared parameters prevent independent per-turn control?} 

To address this cascade issue, we propose Entropy-regularized Policy Optimization (\textbf{\ours}), a novel framework that combines entropy regularization with specialized mechanisms for stable on-policy training under sparse reward conditions. 
% Our framework is guided by the central insight that standard entropy regularization is insufficient because it lacks temporal awareness. We find that anchoring the policy's entropy to a dynamically adjusted historical bound provides the necessary stability to halt the exploration-exploitation cascade failure without sacrificing essential exploration.
Our contributions are fourfold. \vspace{-4pt}

\ding{182} \update{\textbf{\textit{We empirically discover and formally characterize the exploration-exploitation cascade failure on standard multi-turn benchmarks}}, demonstrating that this novel failure mode, where entropy oscillations compound across turns under sparse rewards, is the primary cause of poor performance in standard RL baselines.}
\vspace{-4pt}

\ding{183} \textbf{\textit{We extend standard entropy regularization to multi-turn settings}}, where entropy accumulates across sequential turns through shared parameters, requiring trajectory-level aggregation rather than per-step computation.
\vspace{-4pt}

\ding{184} \textbf{\textit{We introduce an entropy smoothing regularizer that penalizes deviations from historical entropy averages}}, effectively reducing the oscillations between overconfidence and over-exploration that characterize cascade failure.
\vspace{-4pt}

\ding{185} \textbf{\textit{We develop an adaptive weighting scheme that dynamically balances exploration and exploitation across training phases}}, applying stronger smoothing early to establish controlled exploration, relaxing mid-training as the policy stabilizes, and maintaining sufficient regularization in later phases to ensure convergence.
\vspace{-4pt}

Together, these components create a theoretically grounded and general framework that \update{prevents the empirically observed cascade failure and ensures} optimal exploration-exploitation trade-offs while being compatible with any on-policy optimization method. We validate \ours on challenging benchmarks ScienceWorld~\citep{sciworld} and ALFWorld~\citep{alfworld}, achieving up to 152\% performance improvement with more stable training dynamics, transforming previously unstable sparse-reward scenarios into smoothly converging optimization problems.

\section{Related Work}

\subsection{Reinforcement Learning for LLMs}
RLHF~\citep{rlhf} and DPO~\citep{dpo} have become foundational approaches for aligning LLMs with human preferences, with both methods significantly improving model alignment and instruction-following capabilities. Recent RL methods such as GRPO~\citep{grpo} and DAPO~\citep{dapo} further enhance LLM reasoning abilities during post-training through verified rewards. In contrast to PPO~\citep{ppo}, these methods leverage batch-wise advantage computation from identical prompts, obviating the critic model and substantially improving the computational tractability of large-scale RL training for LLMs. However, a primary challenge in applying RL to LLMs is the phenomenon of policy entropy collapse: models rapidly reduce their stochasticity, converging on narrow, over-optimized behaviors~\citep{clip-conv,Rl-plus,8020entropy,deng2025trial,skywork,ReasonwithExplore,cheng2025reasoning}. In response, recent work has focused on integrating entropy control mechanisms into the optimization process to preserve policy diversity. For instance, Cui et al.~\citep{clip-conv} regulate the impact of high-covariance tokens by applying a clipping function and a KL penalty. Meanwhile, Cheng et al.~\citep{cheng2025reasoning} augment the advantage function with a clipped, gradient-detached entropy term, which encourages deeper reasoning chains changing the original policy optimization direction.

\subsection{Reinforcement Learning for LLM Agents}
To enhance their autonomy, LLM agents are designed to interact with external environments using diverse toolsets~\citep{oaicua,tongyidr}. However, training agents to complete multi-step tasks with these tools presents significant challenges for standard reinforcement learning, including sparse rewards and credit assignment problems. To address these issues, seminal works have introduced advanced training paradigms such as hierarchical RL~\citep{archer}, autonomous learning~\citep{digirl}, and off-policy Q-learning~\citep{digiq}. Meanwhile, another line of research employs supervised fine-tuning (SFT) to directly enhance the models’ decision-making abilities, training them on vast datasets of high-quality tool-use trajectories to master complex environments and APIs~\citep{agentgym,zhang2024codeagent,qin2023toolllm}. Recently, to leverage the training stability of GRPO~\citep{grpo}, researchers have extended single-turn GRPO to multi-turn training settings with various training techniques to improve performance~\citep{jin2025search,gigpo,ragen}. To guide learning in long-horizon scenarios, \rlvmr~\citep{zhang2025rlvmr} introduces verifiable meta-reasoning rewards that provide dense, intermediate feedback on the agent's reasoning process. However, these methods overlook the entropy dynamics unique to multi-turn settings. Because all turns share policy parameters, entropy adjustments cannot independently control early-turn exploration and late-turn exploitation. This coupling, combined with sparse rewards that provide no intermediate correction, causes the exploration-exploitation cascade failure, where severe entropy oscillations destabilize training and prevent effective long-horizon learning.

\section{Preliminary}

\subsection{On-policy Optimization}

On-policy optimization is a fundamental paradigm in reinforcement learning where the agent learns to improve its policy by directly optimizing the expected return using trajectories sampled from the current policy. Given a parameterized stochastic policy $\pi_\theta(a|s)$ with parameters $\theta \in \mathbb{R}^d$, the objective is to maximize the expected return,given by $J(\theta) = \mathbb{E}_{\tau \sim \pi_\theta}[R(\tau)]
$,
% \begin{equation}
% J(\theta) = \mathbb{E}_{\tau \sim \pi_\theta}[R(\tau)]
% \end{equation}
where $\tau$ denotes a trajectory and $R(\tau) = \sum_{t=0}^{T} \gamma^t r_t$ is the discounted return.
On-policy optimization methods build on:

\vspace{-10pt}\begin{equation}
\nabla_\theta J(\theta) = \mathbb{E}_{\tau \sim \pi_\theta}\left[\sum_{t=0}^{T} \nabla_\theta \log \pi_\theta(a_t|s_t) A^{\pi_\theta}(s_t, a_t)\right]
\end{equation}
where $A^{\pi_\theta}(s_t, a_t) = Q^{\pi_\theta}(s_t, a_t) - V^{\pi_\theta}(s_t)$ is the advantage function.
While the policy gradient provides an unbiased estimator of $\nabla_\theta J(\theta)$, directly optimizing this objective can lead to instability due to large policy updates. To address this, modern on-policy methods employ surrogate objective functions that approximate the policy gradient while ensuring stable learning.
The standard policy gradient can be reformulated as the surrogate objective $L^{PG}(\theta) = \mathbb{E}_{\tau \sim \pi_{\theta_{\text{old}}}}\left[\frac{\pi_\theta(a_t|s_t)}{\pi_{\theta_{\text{old}}}(a_t|s_t)} \hat{A}_t\right]$.
% \begin{equation}
% L^{PG}(\theta) = \mathbb{E}_{\tau \sim \pi_{\theta_{\text{old}}}}\left[\frac{\pi_\theta(a_t|s_t)}{\pi_{\theta_{\text{old}}}(a_t|s_t)} \hat{A}_t\right]
% \end{equation}
However, this surrogate objective is only valid for infinitesimally small updates. Proximal Policy Optimization (PPO)~\citep{ppo} addresses this limitation by constraining the policy ratio through a clipped surrogate objective:
% \begin{equation}
$
L^{CLIP}(\theta) = \mathbb{E}_{\tau \sim \pi_{\theta_{\text{old}}}}\left[\min\left(r_t(\theta)\hat{A}_t, \text{clip}(r_t(\theta), 1-\epsilon, 1+\epsilon)\hat{A}_t\right)\right]$
% \end{equation}
where $r_t(\theta) = \frac{\pi_\theta(a_t|s_t)}{\pi_{\theta_{\text{old}}}(a_t|s_t)}$ is the importance sampling ratio, $\mathbb{E}_{\tau \sim \pi_{\theta_{\text{old}}}}[\cdot]$ denotes expectation over trajectories sampled under $\pi_{\theta_{\text{old}}}$, and $\epsilon$ defines the trust region bounds.
Group Relative Policy Optimization (GRPO)~\citep{grpo} extends PPO by modifying the advantage function computation. Instead of using standard advantages $\hat{A}_t$, GRPO employs group-relative advantages, computed as $\tilde{A}_t = \frac{R_t - \mu_g}{\sigma_g + \delta}
$.
% \begin{equation}
% \tilde{A}_t = \frac{R_t - \mu_g}{\sigma_g + \delta}
% \end{equation}
where $R_t$ is the return from timestep $t$, $\mu_g$ and $\sigma_g$ are the mean and standard deviation of returns within group $g$, and $\delta$ is a small constant for numerical stability. This group-based normalization provides more stable gradient estimates.

\subsection{Problem formulation}
% \subsection{Multi-Turn RL Framework for Sequential Decision Tasks}

We formalize the multi-turn task as a sequential decision-making reinforcement learning problem. A single LLM agent $\pi_\theta$ executes a task through $T$ turns over a trajectory $\tau = (s_0, a_0, r_0, \ldots, s_T, a_T, r_T)$. The reward is sparse, with $r_t = 0$ for all intermediate turns and only the final turn receiving the task outcome reward, such that the total return is $R(\tau) = \sum_{t=0}^{T} \gamma^t r_t = r_T$,
% \begin{equation}
% R(\tau) = \sum_{t=0}^{T} \gamma^t r_t = r_T
% \end{equation}
where $r_T \in \{0, 1\}$ represents the binary task outcome. In our experimental settings, specifically ALFWorld~\citep{alfworld} and SciWorld~\citep{sciworld}, we assume an undiscounted formulation where $\gamma=1$.
All turns within the same task share the final outcome reward, creating a credit assignment challenge across sequential turns. The multi-turn policy optimization differs from standard RL in that losses accumulate across all $T$ turns before parameter updates:
% \begin{equation}~\label{eq:mt}
$L^{MT}(\theta) = \mathbb{E}_{\tau \sim \pi_{\theta_{\text{old}}}}\left[\mathbb{E}_{t \sim T}\left[\min\left(r_t(\theta)A_t, \text{clip}(r_t(\theta), 1-\epsilon, 1+\epsilon)A_t\right)\right]\right]$
% \end{equation}
where $r_t(\theta) = \frac{\pi_\theta(a_t|s_t)}{\pi_{\theta_{\text{old}}}(a_t|s_t)}$ and $A_t$ represents the advantage estimate at turn $t$ within the multi-turn trajectory. $\mathbb{E}_t[\cdot]$ denotes expectation over timesteps within turn $t$, while the outer expectation is over trajectories sampled under $\pi_{\theta_{\text{old}}}$.

\section{Methodology}

\subsection{Entropy Regularization}
To address the exploration-exploitation cascade failure, we first adapt entropy regularization to capture the temporal structure of multi-turn interactions. Unlike traditional RL where entropy is computed per-step, we recognize that in multi-turn environments, decisions compound across subsequent turns through shared policy parameters. Therefore, we compute entropy across all turns within each trajectory and average over the batch of trajectories. The entropy-regularized policy loss is formulated as $L^{ER}(\theta) = L^{MT}(\theta) - \lambda L^{H}(\theta)$,
where $\lambda$ is the entropy coefficient, and the entropy loss is averaged over the batch of trajectories:

\vspace{-10pt}
\begin{equation}~\label{eq:lh}
L^{H}(\theta) = \frac{1}{B} \sum_{j=1}^{B} \frac{1}{T} \sum_{t=0}^{T-1} \frac{1}{|\tau_{j,t}|} \sum_{i=1}^{|\tau_{j,t}|} \mathcal{H}_{j,t,i}
\end{equation}
where $B$ is the batch size (number of trajectories), $T$ is the number of turns per trajectory, $\mathcal{H}_{j,t,i}$ is the entropy at token position $i$ in turn $t$ of trajectory $j$, and $|\tau_{j,t}|$ represents the sequence length at turn $t$ of trajectory $j$. The token-level entropy $\mathcal{H}$ is computed from the model's probability distribution over the vocabulary at each position as $\mathcal{H} = -\sum_{v \in V} p(v|w_{<t}) \log p(v|w_{<t})$,
where $p(v|w_{<t})$ is the probability of token $v$ from vocabulary $V$ given the preceding context $w_{<t}$.

\subsection{Entropy Smoothing Regularizer}
While trajectory-level entropy regularization captures multi-turn structure, it cannot prevent the severe entropy oscillations caused by shared parameters and sparse rewards. To stabilize entropy dynamics, we introduce an entropy smoothing mechanism that anchors current entropy to historical averages. We maintain an entropy history window $\mathcal{W}_k = \{\bar{H}_{0}, \ldots, \bar{H}_{m}, \ldots, \bar{H}_{k-1}\}$ for RL step $k$, storing the average entropy $\bar{H}_m$ across all trajectories at the token level for each previous RL step $m$. The historical entropy reference is computed as the mean $\bar{H}^{\mathcal{W}_k} = \frac{1}{k} \sum_{m=0}^{k-1} \bar{H}_m$. This historical anchoring dampens the entropy fluctuations that destabilize training under sparse reward conditions. We apply a token-wise penalty based on acceptable entropy ranges relative to this historical average:
% \begin{equation}
% \mathcal{P}_{n,t,i} = \begin{cases}
% 0, & \text{if } \kappa_l \bar{H}^{\mathcal{W}_k}  \leq \mathcal{H}_{n,t,i} \leq \kappa_r \bar{H}^{\mathcal{W}_k},  \\
% \alpha, & \text{otherwise.}
% \end{cases}
% \end{equation}

\vspace{-7pt}
\begin{equation}
    % \mathcal{P}_{n,t,i} = \alpha \cdot \left[ \max\left(0, \kappa_l \bar{H}^{W_k} - H_{n,t,i}\right) + \max\left(0, H_{n,t,i} - \kappa_r \bar{H}^{W_k}\right) \right]
    \mathcal{P}_{n,t,i} = \left[\kappa_l \bar{H}^{W_k} - H_{n,t,i}\right]_+ + \left[H_{n,t,i} - \kappa_r \bar{H}^{W_k}\right]_+
\end{equation}

where $[x]_+ = \max(0, x)$ denotes the ReLU function, and $\kappa_l$, $\kappa_r$ define the acceptable entropy corridor. By bounding entropy within historical averages, we prevent the severe oscillations that arise from shared parameter coupling under sparse rewards. Aggregating these penalties across all tokens, turns, and trajectories yields the smoothing loss:

\vspace{-7pt}
\begin{equation}~\label{eq:lsmooth}
L^{smooth}(\theta) = \frac{1}{B} \sum_{n=1}^{B} \frac{1}{T} \sum_{t=0}^{T-1} \frac{1}{|\tau_{n,t}|} \sum_{i=1}^{|\tau_{n,t}|} \alpha \mathcal{P}_{n,t,i}
\end{equation}

where $\alpha$ provides the penalty weight for tokens with entropy outside the acceptable range. The complete entropy-smoothed policy optimization loss is then defined as $L^{EPO}(\theta) = L^{MT}(\theta) - \lambda [ L^{H}(\theta) - \beta_k L^{smooth}(\theta) ]$,
where the dynamic coefficient $\beta_k$ follows an exponential schedule that adapts smoothing strength across training:

\vspace{-7pt}
\begin{equation}
\beta_k = 1 + e^{-\gamma \frac{k}{k_{mid}}}
\label{eq:beta_k}
\end{equation}

This adaptive schedule stabilizes entropy dynamics throughout training: it begins with stronger smoothing to establish controlled exploration, gradually relaxes as the policy stabilizes around mid-training ($k_{mid} = \lfloor K/2 \rfloor$), and maintains sufficient regularization in later phases to ensure smooth convergence. The parameter $\gamma$ controls the decay rate of this transition. Overall, the policy is updated according to the following objective:

\vspace{-7pt}
\begin{equation}
    L^{EPO}(\theta) = L^{MT}(\theta) - \lambda [ L^{H}(\theta) - \beta_k L^{\text{smooth}}(\theta) ].
    \label{eq:epo_loss}
\end{equation}

where $\lambda$ is the entropy coefficient controlling the overall strength of entropy regularization, and $\beta_k$ is a dynamic coefficient that adapts the smoothing strength across training phases. By decoupling the entropy terms from the policy loss, this formulation provides direct gradient signals $\nabla_\theta L^{H}(\theta)$ to guide exploration while the smoothing regularizer $L^{\text{smooth}}(\theta)$ ensures stability by anchoring entropy within historical bounds.
% \subsection{EPO Training Procedure}
\autoref{algo:epo} presents the full optimization procedure, incorporating these components.

\begin{algorithm}[tb!]
\caption{Entropy-regularized Policy Optimization (EPO)}
\label{algo:epo}
\begin{algorithmic}[1]
\Require Policy $\pi_\theta$, entropy coefficient $\lambda$, penalty weight $\alpha$, corridor bounds $\kappa_l, \kappa_r$, decay rate $\gamma$, total steps $K$
\State Initialize $\theta_0$, entropy history $\mathcal{W}_0 \leftarrow \emptyset$
\For{$k = 0, 1, \ldots, K-1$}
    \State Collect trajectories $\mathcal{D}_k = \{\tau_j\}_{j=1}^B$ using $\pi_{\theta_{\text{old}}}$
    \State Compute advantages $\hat{A}_t$ and policy loss $L^{MT}(\theta)$
    \State Compute entropy loss $L^{H}(\theta)$ via \autoref{eq:lh}
    \If{$k > 0$}
        \State $\bar{H}^{\mathcal{W}_k} \leftarrow \frac{1}{k} \sum_{m=0}^{k-1} \bar{H}_m$ \Comment{Entropy mean}
        \State Compute loss $L^{\text{smooth}}(\theta)$ via \autoref{eq:lsmooth}
    \Else
        \State $L^{\text{smooth}}(\theta) \leftarrow 0$
    \EndIf
    \State $\beta_k \leftarrow 1 + \exp(-\gamma \cdot k / k_{mid})$ where $k_{mid} = \lfloor K/2 \rfloor$
    \State Update $\theta$ by minimizing \autoref{eq:epo_loss}
    \State $\mathcal{W}_{k+1} \leftarrow \mathcal{W}_k \cup \{\bar{H}_k\}$ \Comment{Append current batch entropy}
\EndFor
\Ensure Optimized policy parameters $\theta_K$
\end{algorithmic}
\end{algorithm}

% 1. we appear more wide stable epoch range, like get the average reward compared to the baseline. 
\begin{table*}[tb!]
    \centering
    \caption{For PPO and GRPO baselines with our EPO method: \cfirst{better performance} indicates improvement over baseline, \csecond{worse performance} indicates degradation. \cthird{Highlighted values} represent the best performance among other baseline methods. $\Delta$ shows the relative improvement (\%) when applying our method. Results for other baseline methods (\react, \agentgym, SFT, \gigpo, \rlvmr) are sourced from the \rlvmr~\citep{zhang2025rlvmr} paper. We ran our own implementations of PPO, GRPO, and EPO, tuning hyperparameters across multiple trials to obtain stable results.}
    \vspace{-7pt}
    \label{tab:quantitative_result}
    \footnotesize  
    \setlength{\tabcolsep}{3pt}  
    \renewcommand\arraystretch{0.98}  
    
    % The corrected tabular definition and header rows are below
    \begin{tabular}{l|ccccc|ccccc}
    \thickhline
    \rowcolor{mygray}
    % The empty cell corresponds to the 'Method' column
    & \multicolumn{5}{c|}{\textbf{ScienceWorld}} & \multicolumn{5}{c}{\textbf{ALFWorld}}\\
    \rowcolor{mygray}
    \textbf{Method} & \textbf{LLM} & \multicolumn{2}{c}{\textbf{IID}} & \multicolumn{2}{c|}{\textbf{OOD}} & \textbf{LLM} & \multicolumn{2}{c}{\textbf{IID}} & \multicolumn{2}{c}{\textbf{OOD}}\\
    \rowcolor{mygray}
    & & Succ.$^{*}$ & $\overline{Succ.}$ & Succ.$^{*}$ & $\overline{Succ.}$ & & Succ.$^{*}$ & $\overline{Succ.}$ & Succ.$^{*}$ & $\overline{Succ.}$ \\
    \hline
    \react\tiny{~\citep{react}} & \texttt{\scriptsize GPT-4o} & 45.4 & - & \textbf{\cthird{49.2}} &  - & \texttt{\scriptsize GPT-4o} & 57.3 & - & 66.0 & - \\ 
    \react\tiny{~\citep{react}} & \texttt{\scriptsize DeepSeek-R1} & 22.2 & - & 31.4 &  - & \texttt{\scriptsize DeepSeek-R1} & 68.8 & - & 70.2 & - \\ 
    \react\tiny{~\citep{react}} & \texttt{\scriptsize Qwen2.5-7B} & 7.8 & - & 11.3 &  - &\texttt{\scriptsize Qwen2.5-7B} & 23.1 & - & 28.5 & -\\ 
    \hline
    \agentgym\tiny{~\citep{agentgym}} & \texttt{\scriptsize LLaMa2-7B} & 46.9 & - & 33.6 &  - & \texttt{\scriptsize LLaMa2-7B} & 76.6 & - & 63.3 & - \\ 
    SFT & \texttt{\scriptsize Qwen2.5-7B} & 36.7 & - & 32.0 &  - & \texttt{\scriptsize Qwen2.5-7B} & 63.3 & - & 57.0 & - \\ 
    \hline
    \gigpo\tiny{~\citep{gigpo}} & \texttt{\scriptsize Qwen2.5-7B} & 53.4 & - & 35.2 &  - &\texttt{\scriptsize Qwen2.5-7B} & 89.5 & - & 90.2 & - \\ 
    \rlvmr\tiny{~\citep{zhang2025rlvmr}} & \texttt{\scriptsize Qwen2.5-7B} & \textbf{\cthird{67.2}} & - & 43.0 &  - & \texttt{\scriptsize Qwen2.5-7B} & \textbf{\cthird{91.4}} & - & \textbf{\cthird{91.8}} & - \\ 
    \hline
    PPO\tiny{~\citep{ppo}} &\lightcem{\texttt{\scriptsize Qwen2.5-7B}} & \lightcem{\csecond{64.6}} & \lightcem{\csecond{38.4}} & \lightcem{\csecond{58.3}} &  \lightcem{\csecond{39.1}} & \lightcem{\texttt{\scriptsize Qwen2.5-3B}} & \lightcem{\cfirst{95.8}} & \lightcem{\csecond{72.3}} & \lightcem{\csecond{87.5}} & \lightcem{\csecond{70.9}} \\
    
    +EPO & \cem{\texttt{\scriptsize Qwen2.5-7B}} & \cem{\cfirst{100.0}} & \cem{\cfirst{96.8}} & \cem{\cfirst{100.0}} & \cem{\cfirst{96.2}} & \cem{\texttt{\scriptsize Qwen2.5-3B}} &\cem{ \csecond{85.4}} &\cem{ \cfirst{73.4}} & \cem {\cfirst{91.7}} &\cem {\cfirst{74.3}} \\   
    $\Delta$ & & \bf{54.8\%} & \bf{152.1\%} & \bf{71.5\%} & \bf{146.0\%} & & -10.9\% & \bf{1.5\%} & \bf{4.8\%} & \bf{4.8\%} \\ 
    
    \hdashline
    GRPO\tiny{~\citep{grpo}} & \lightcem{\texttt{\scriptsize Qwen2.5-7B}} & \lightcem{\csecond{93.8}} & \lightcem{\csecond{81.6}} & \lightcem{\csecond{91.7}} & \lightcem{\csecond{80.9}} & \lightcem{\texttt{\scriptsize Qwen2.5-3B}} & \lightcem{\csecond{87.5}} & \lightcem{\csecond{63.3}} & \lightcem{\csecond{83.3}} & \lightcem{\csecond{63.5}} \\
    
    +EPO & \cem{\texttt{\scriptsize Qwen2.5-7B}} & \cem{\cfirst{95.8}} & \cem{\cfirst{83.8}} & \cem{\cfirst{95.8}} & \cem{\cfirst{81.3}} & \cem{\texttt{\scriptsize Qwen2.5-3B}} & \cem{\cfirst{91.7}} & \cem{\cfirst{75.8}} & \cem{\cfirst{89.6}} & \cem{\cfirst{75.4}} \\
    $\Delta$ & & \bf{2.1\%} & \bf{2.7\%} & \bf{4.5\%} & \bf{0.5\%} & & \bf{4.8\%} & \bf{19.8\%} & \bf{7.6\%} & \bf{18.7\%}  \\ 
    \thickhline
    \end{tabular}
    % \vspace{0.5em}
    \vspace{-7pt}
\end{table*}

\section{Experiments}\label{sec:exper}
We validate EPO on ScienceWorld and ALFWorld, two sparse-reward benchmarks requiring 30+ turn interactions. Our experiments examine: (1) performance gains over existing RL methods, (2) necessity of each proposed component via ablation, and (3) failure modes of alternative entropy control strategies. Results demonstrate that EPO effectively eliminates cascade failure, transforming unstable training into smooth convergence.

\subsection{Experiments Setup}
This section outlines our experimental setup, including the benchmarks, evaluation protocol, baselines, and implementation details. Further details can be found in~\autoref{app:exper}.

\textbf{Benchmark.} We evaluate on two challenging benchmarks that require different reasoning capabilities, ScienceWorld~\citep{sciworld} and ALFWorld~\citep{alfworld}. ScienceWorld  focuses on text-based scientific experimentation, demanding systematic hypothesis testing and structured exploration. ALFWorld  is an embodied environment containing 4,639 household task instances across six categories, requiring multi-step decision-making and spatial reasoning. To improve the generalizability of our approach across multiple scenarios, we finetune the foundation model directly on the environment using RL, rather than employing trajectory finetuning for initialization.

\textbf{Evaluation Setting.} To evaluate generalization capabilities, we focus on two key evaluation scenarios: \textbf{IID} (in-distribution) covers seen task variants and categories, while \textbf{OOD} (out-of-distribution) evaluates on unseen task variants within seen categories. This design allows us to measure both optimization effectiveness and generalization robustness, which are crucial for practical deployment. We employ dual success rate metrics to capture different aspects of performance: \textbf{Succ.$^{*}$} reports the average of maximum success rates across random seeds, while $\overline{\textbf{Succ.}}$ measures average performance after convergence, reflecting practical reliability. Given the high variance inherent in RL, final performance scores alone can be misleading. We therefore present averaged curves to provide a more robust comparison and illustrate the performance evolution throughout the training process.

\textbf{Baselines.} We conduct comprehensive comparisons across multiple paradigms: (1) \textbf{Prompting-based approaches} such as ReAct~\citep{react} that utilize in-context learning without parameter optimization; (2) \textbf{Trajectory-based and platform methods} including supervised fine-tuning (SFT) through expert trajectory imitation and \agentgym~\citep{agentgym} which provides a unified framework with behavioral cloning and self-evolution mechanisms; (3) \textbf{General reinforcement learning approaches} encompassing standard on-policy methods (PPO~\citep{ppo}, GRPO~\citep{grpo}); (4) \textbf{Agent RL approaches} including recent methods specifically designed for agent training (\gigpo~\citep{gigpo}, \rlvmr~\citep{zhang2025rlvmr}). Our proposed EPO method is architected as a general enhancement framework that can be seamlessly integrated with existing RL paradigms, as exemplified through our PPO+EPO and GRPO+EPO implementations.

\begin{figure*}[tb!]
\centering
% 第一行的两个子图
\subfigure[\scriptsize Training Rewards (ScienceWorld)]{
\begin{minipage}[t]{0.32\linewidth}
\centering
\includegraphics[width=0.99\linewidth]{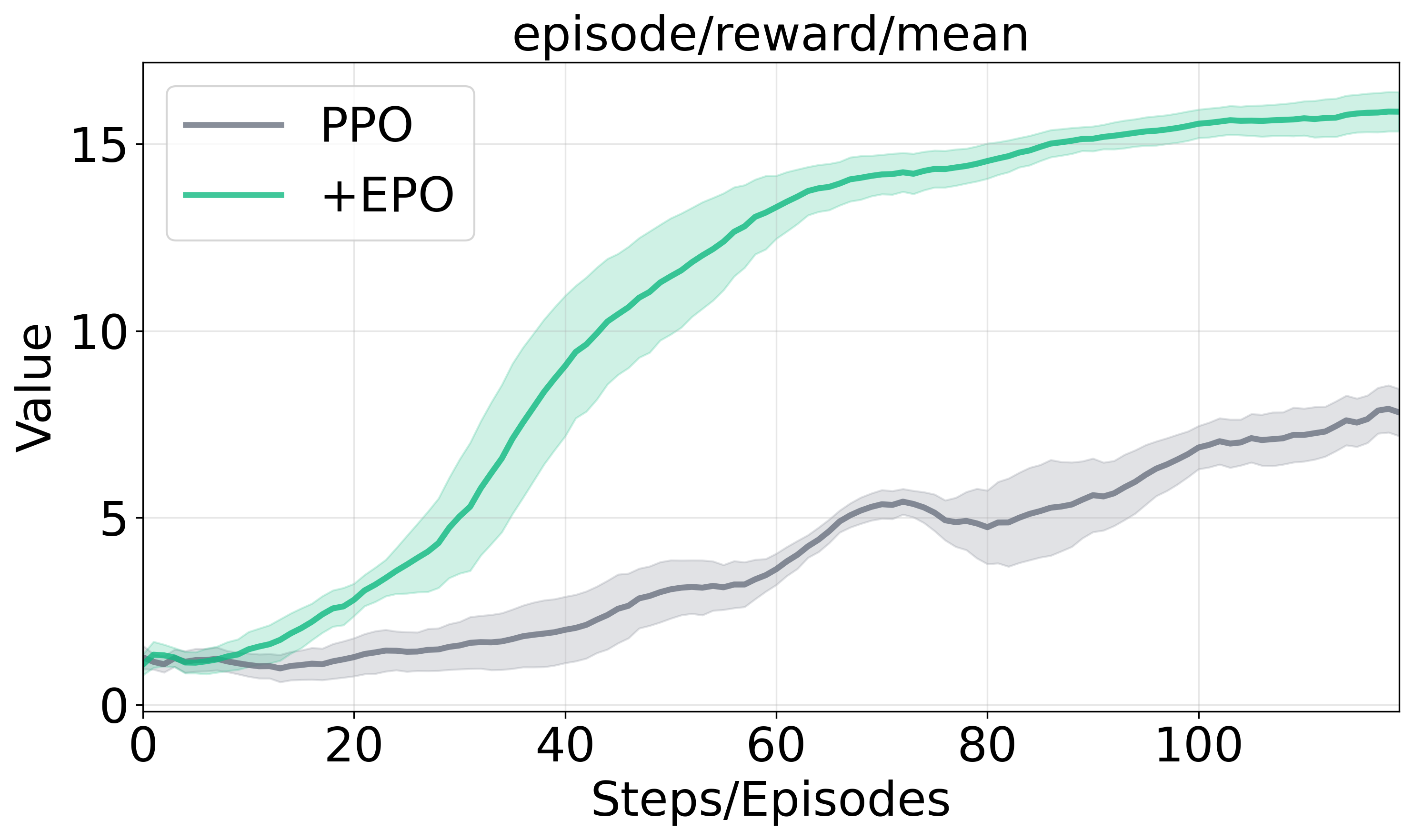}
%\caption{fig1}
\end{minipage}%
}%
\hfill % 填充所有可用的水平空间
\subfigure[\scriptsize IID Success Rate (ScienceWorld)]{
\begin{minipage}[t]{0.32\linewidth}
\centering
\includegraphics[width=0.99\linewidth]{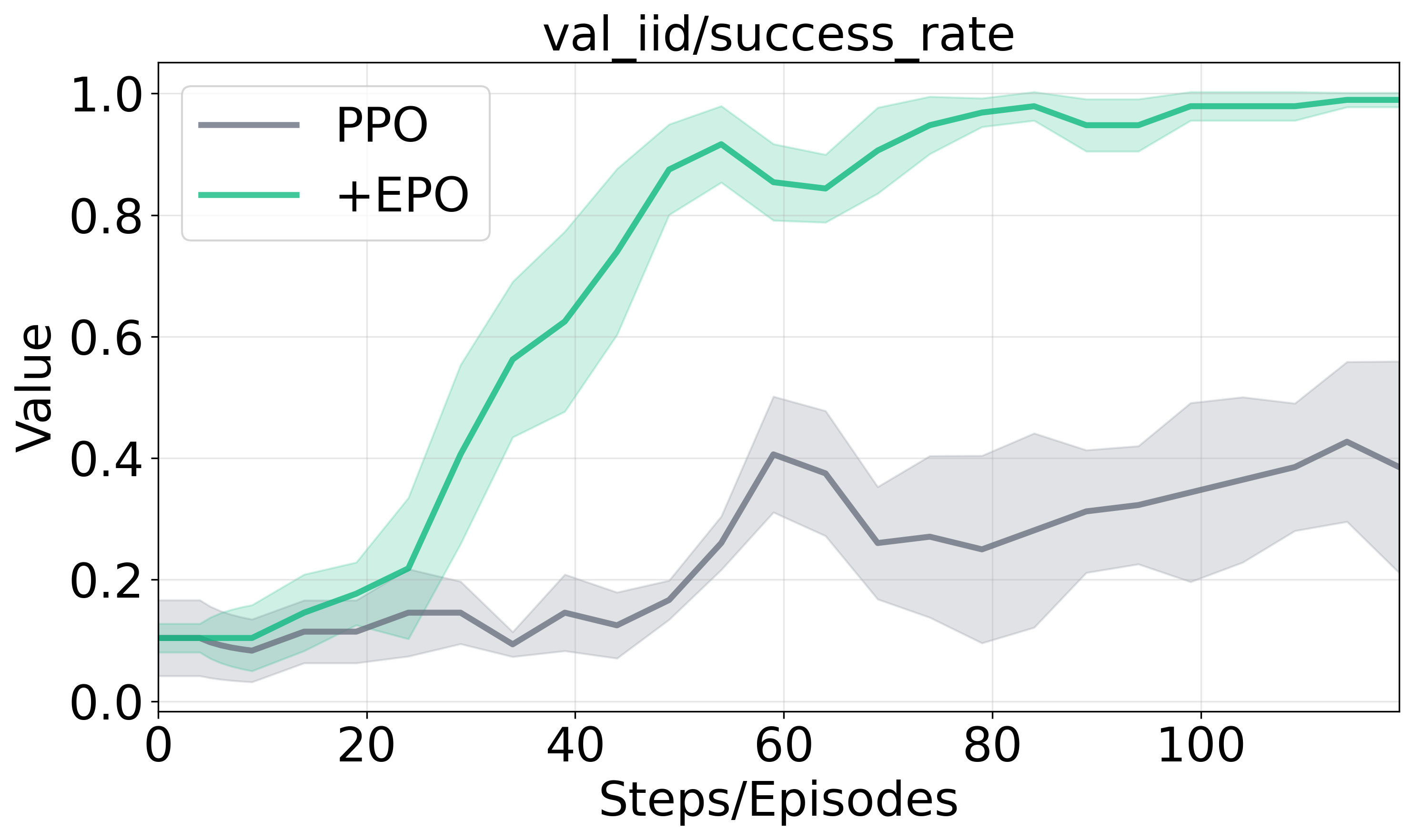}
%\caption{fig2}
\end{minipage}%
}% 
\hfill % 填充所有可用的水平空间
\subfigure[\scriptsize OOD Success Rate (ScienceWorld)]{
\begin{minipage}[t]{0.32\linewidth}
\centering
\includegraphics[width=0.99\linewidth]{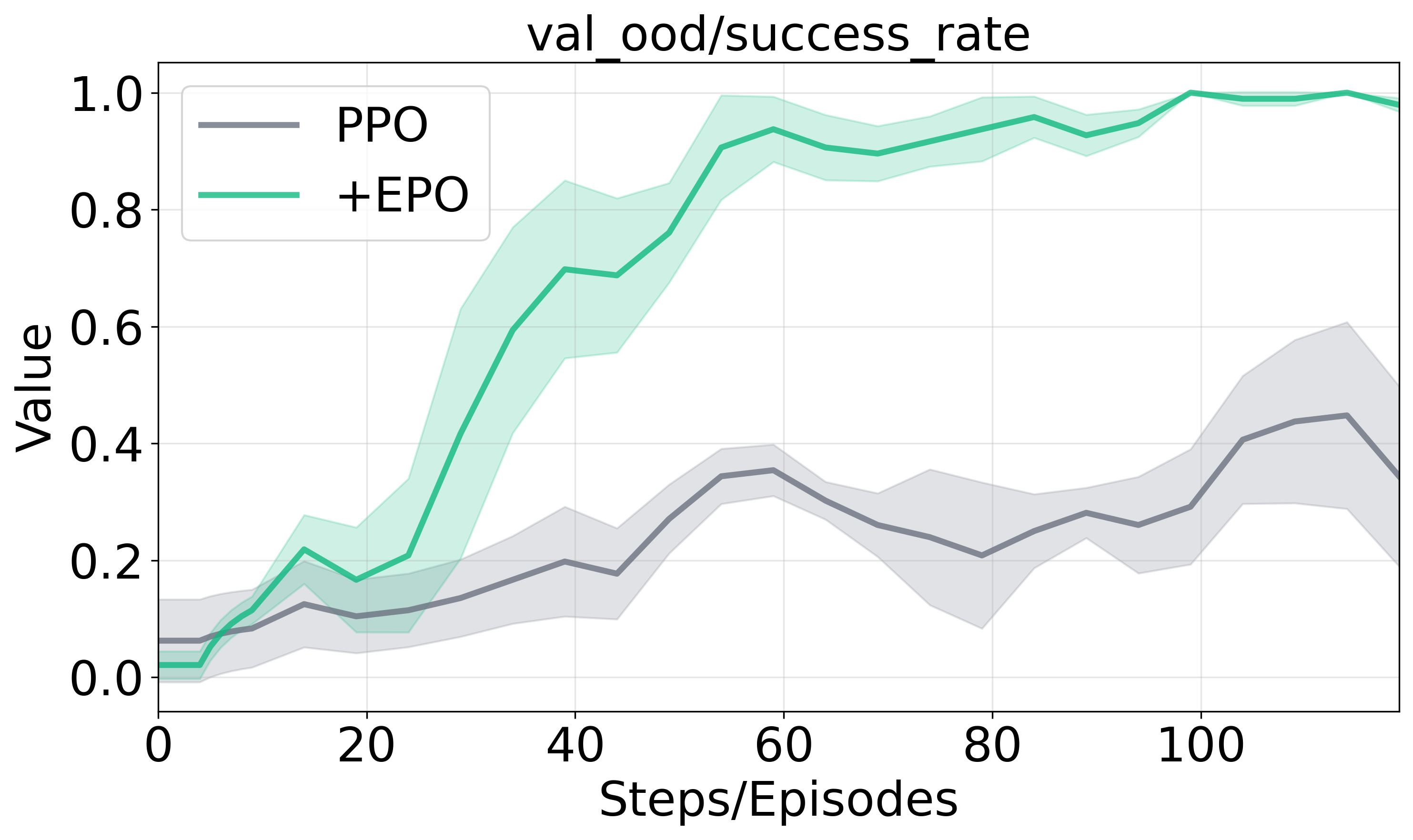}
%\caption{wfig2}
\end{minipage}%
}% 
\\ % 开始新的一行
\vspace{-7pt}
\subfigure[\scriptsize Training Rewards (ALFWorld)]{
\begin{minipage}[t]{0.32\linewidth}
\centering
\includegraphics[width=0.99\linewidth]{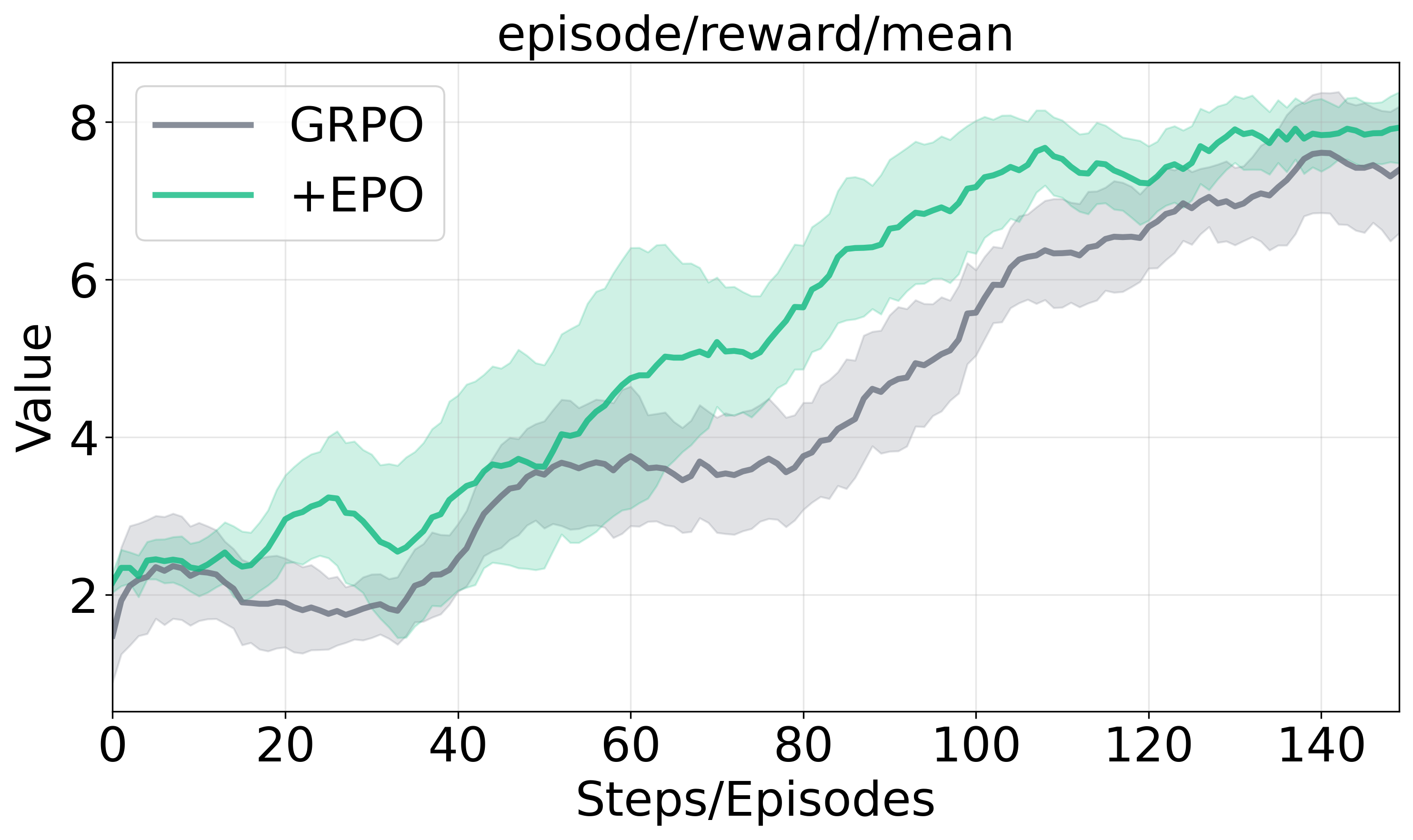}
%\caption{fig1}
\end{minipage}%
}%
\hfill % 填充所有可用的水平空间
\subfigure[\scriptsize IID Success Rate (ALFWorld)]{
\begin{minipage}[t]{0.32\linewidth}
\centering
\includegraphics[width=0.99\linewidth]{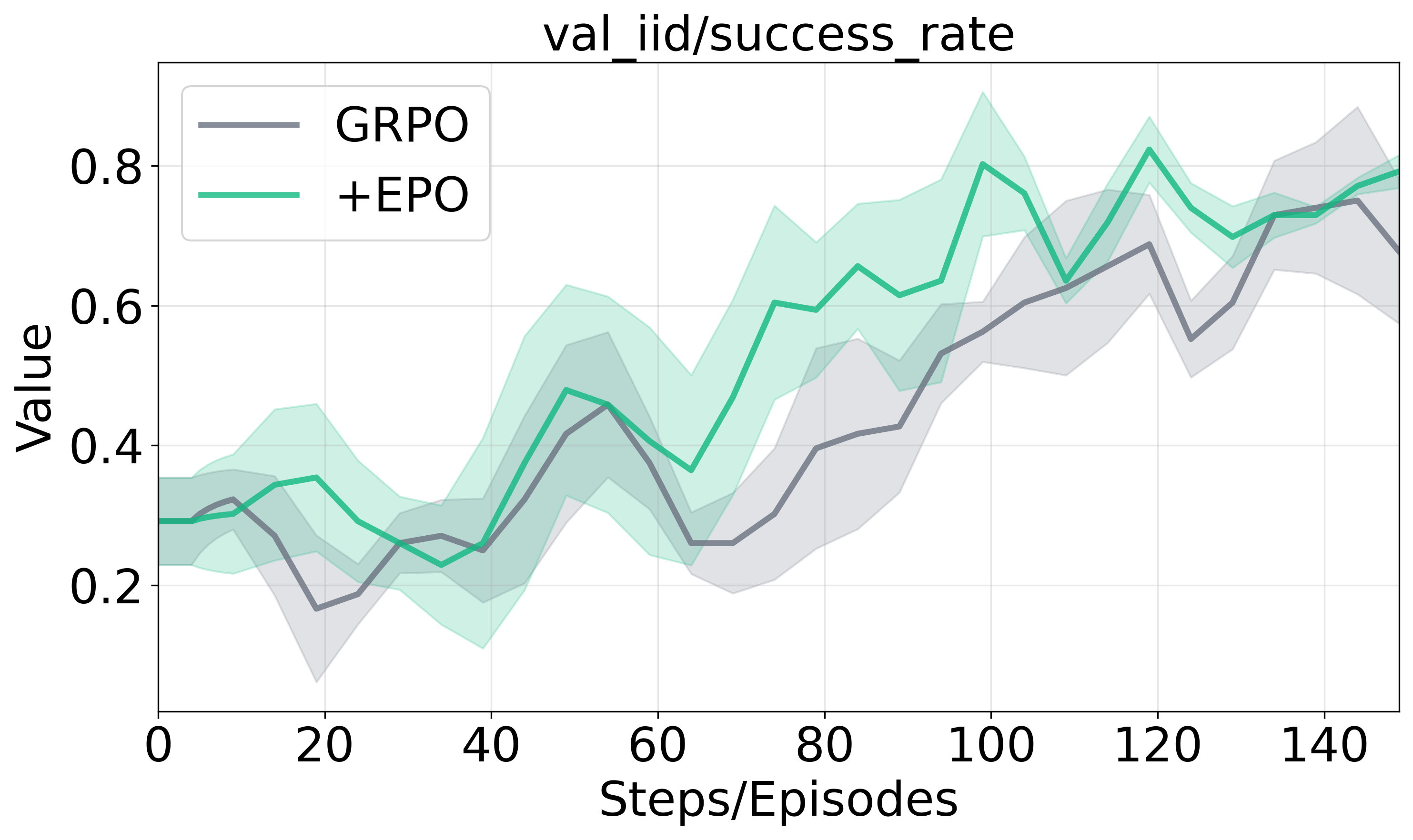}
%\caption{fig2}
\end{minipage}%
}% 
\hfill % 填充所有可用的水平空间
\subfigure[\scriptsize OOD Success Rate (ALFWorld)]{
\begin{minipage}[t]{0.32\linewidth}
\centering
\includegraphics[width=0.99\linewidth]{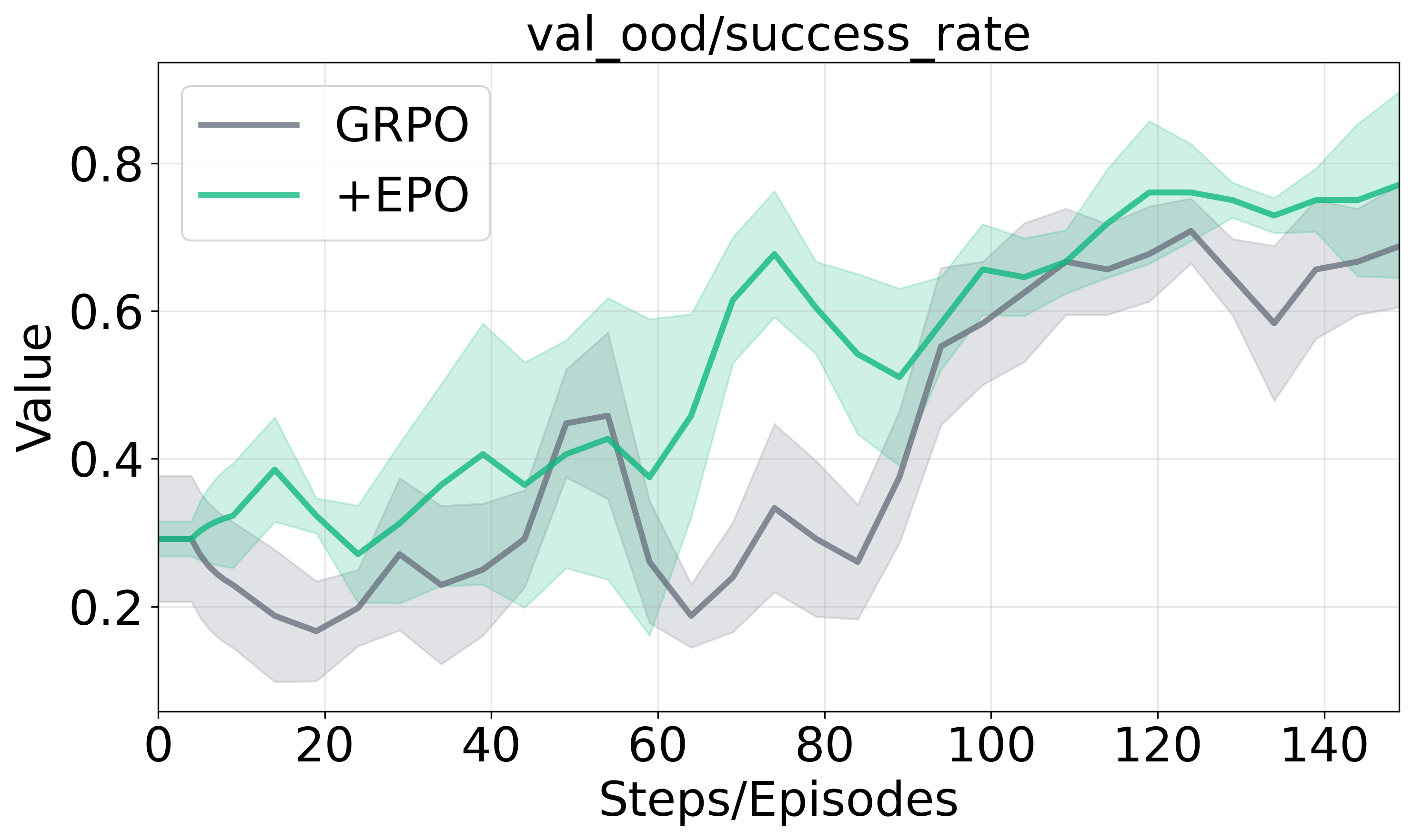}
%\caption{fig2}
\end{minipage}%
}% 

\centering
\vspace{-10pt}
% \captionsetup{font={small}}
\caption{Training dynamics and generalization performance. (a-c) ScienceWorld results comparing PPO and PPO+EPO across training rewards, IID success rate, and OOD success rate. (d-f) ALFWorld results comparing GRPO and GRPO+EPO under identical metrics. EPO eliminates the characteristic entropy oscillations of cascade failure, transforming unstable training into smooth convergence with substantial gains in both IID and OOD scenarios.}
\label{fig:training dynamic}
\vspace{-10pt}
\end{figure*}

\textbf{Implementation Details.}  To optimize performance for each benchmark's complexity, we employ Qwen2.5-3B-Instruct for ALFWorld  and the larger Qwen2.5-7B-Instruct for ScienceWorld's more complex scientific reasoning tasks. Constrained by computational resources, we adopt a single foundation model per task whose size is sufficient to ensure proper convergence, as smaller models consistently fail to converge.
We conduct our own implementations and experiments for our proposed method, PPO, and GRPO baselines across three random seeds to ensure statistical reliability. Results for other baseline methods (\react, \agentgym, SFT, \gigpo, \rlvmr) are sourced from the \rlvmr's~\citep{zhang2025rlvmr} paper.
To account for their different convergence characteristics, we trained the model for 120 RL steps on ScienceWorld  and 150 steps on ALFWorld.
% Learning rate schedules are optimized per method, with PPO on ScienceWorld using 3e-6 and all other configurations using 5e-6, incorporating cosine warmup to ensure stable training dynamics. 

\subsection{Performance Comparison}

\textbf{Quantitative Results Analysis.}
\autoref{tab:quantitative_result} presents comprehensive performance comparisons across both ScienceWorld and ALFWorld environments. Our EPO enhancement demonstrates substantial improvements when integrated with existing RL methods. Notably, PPO+EPO achieves a remarkable 152.1\% improvement in averaged success rates ($\overline{Succ.}$) on ScienceWorld IID tasks, significantly outperforming agent-specialized methods including \gigpo (53.4\%) and \rlvmr (67.2\%). This dramatic improvement stems from EPO's ability to address the exploration-exploitation cascade failure: PPO's aggressive policy updates amplify entropy oscillations under sparse rewards, which EPO's smoothing regularizer effectively dampens by anchoring entropy within historical bounds. In contrast, GRPO's group-relative advantage computation provides inherent stability against such oscillations, yielding more modest but consistent improvements (19.8\% on ALFWorld IID) when combined with EPO. We emphasize the $\overline{Succ.}$ metric as it represents performance averaged across multiple evaluation episodes, providing a more robust comparison by reducing variance from individual runs.

\textbf{Training Dynamics Analysis.} \autoref{fig:training dynamic} illustrates training dynamics and validation performance. The reward curves demonstrate that EPO-enhanced methods achieve substantially higher reward accumulation with superior stability. In ScienceWorld, PPO+EPO reaches approximately 2$\times$ higher training rewards (15 vs.\ 8) with smooth monotonic trajectories, while GRPO+EPO maintains steady upward trends on ALFWorld. The validation curves reveal rapid convergence: on ScienceWorld, EPO variants achieve high success rates (>0.8 for both IID and OOD) within 40 steps, compared to baselines that struggle to exceed 0.4 after 100 steps. On ALFWorld, EPO-enhanced approaches demonstrate consistent performance with reduced variance, particularly in OOD evaluation where baselines frequently drop below 0.2 while EPO variants maintain performance above 0.4.

\begin{figure*}[tb!]
\centering
\subfigure[\scriptsize Training Rewards (ScienceWorld)]{
\begin{minipage}[t]{0.32\linewidth}
\centering
\includegraphics[width=0.99\linewidth]{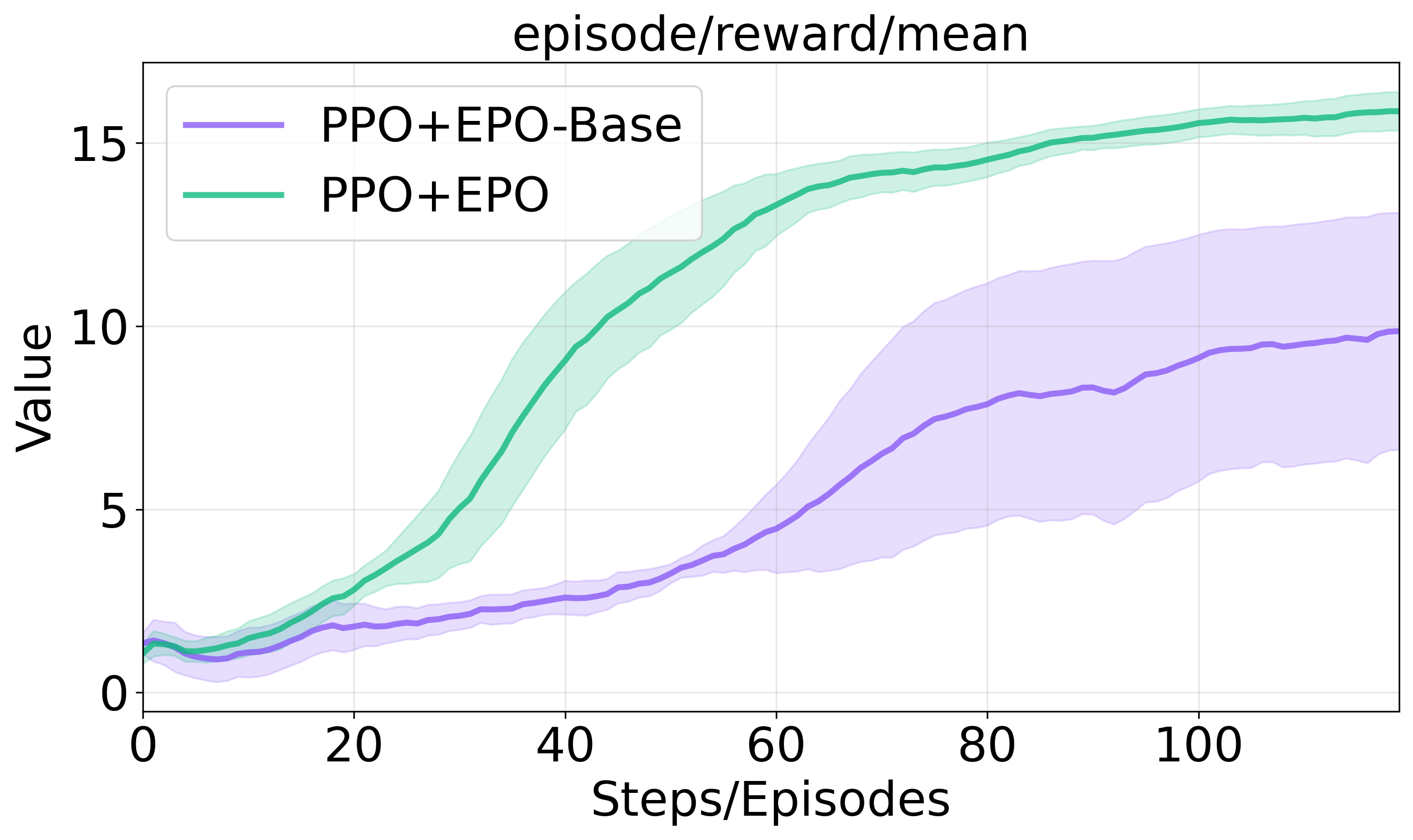}
%\caption{fig1}
\end{minipage}%
}%
\hfill
\subfigure[\scriptsize IID Success Rate (ScienceWorld)]{
\begin{minipage}[t]{0.32\linewidth}
\centering
\includegraphics[width=0.99\linewidth]{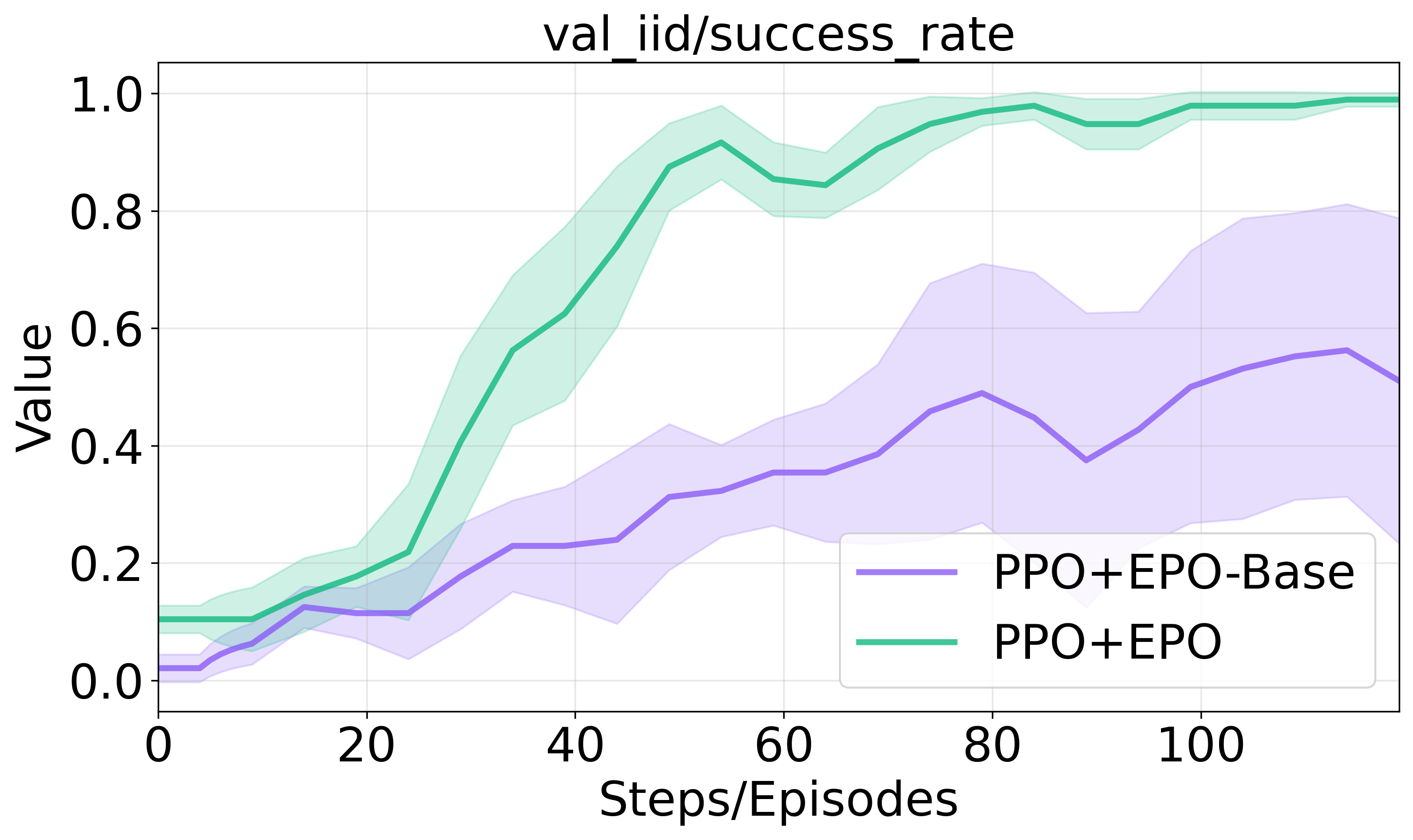}
%\caption{fig2}
\end{minipage}%
}% 
\hfill 
\subfigure[\scriptsize OOD Success Rate (ScienceWorld)]{
\begin{minipage}[t]{0.32\linewidth}
\centering
\includegraphics[width=0.99\linewidth]{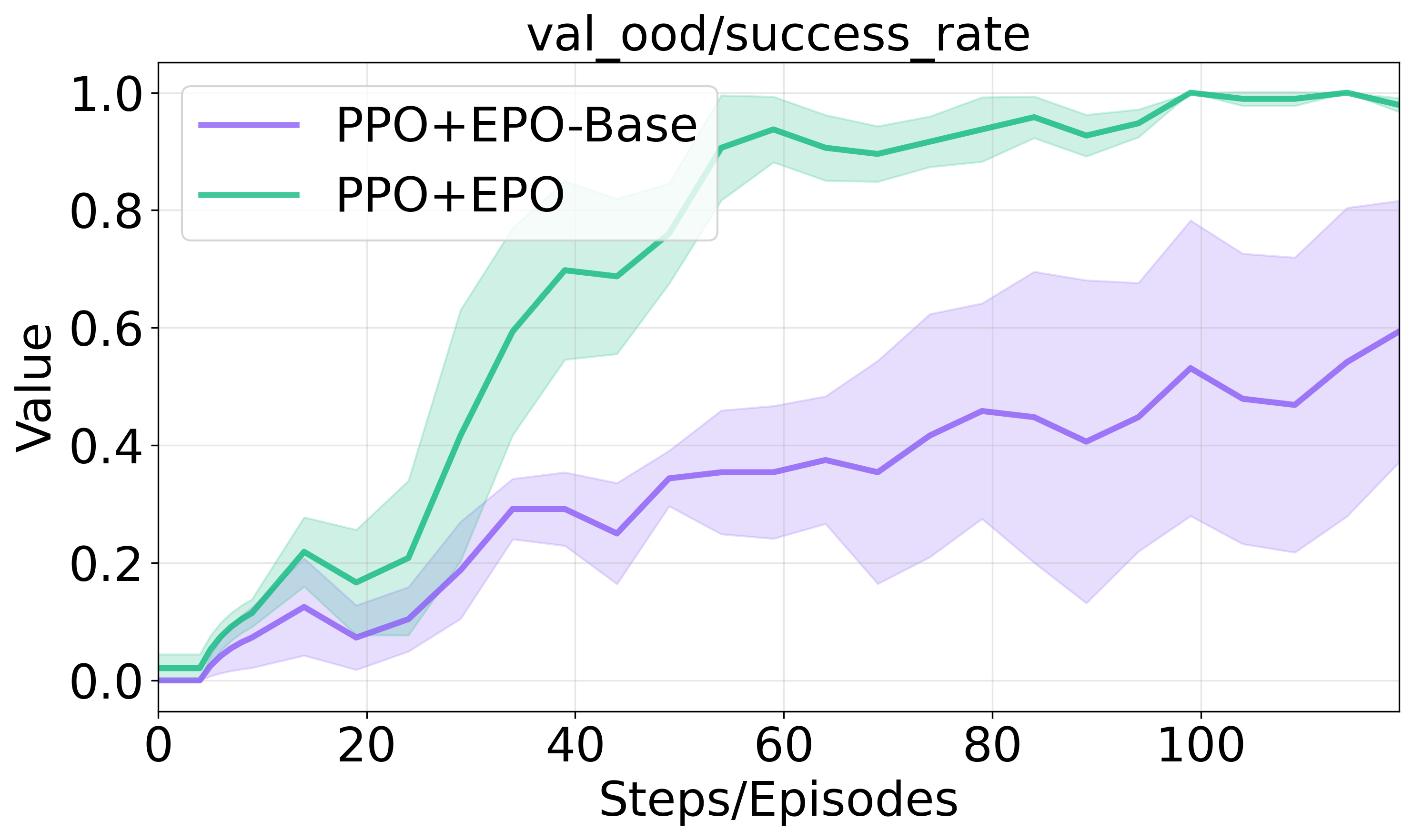}
%\caption{fig2}
\end{minipage}%
}% 
\\ \vspace{-5pt}
\subfigure[\scriptsize Training Rewards (ALFWorld)]{
\begin{minipage}[t]{0.32\linewidth}
\centering
\includegraphics[width=0.99\linewidth]{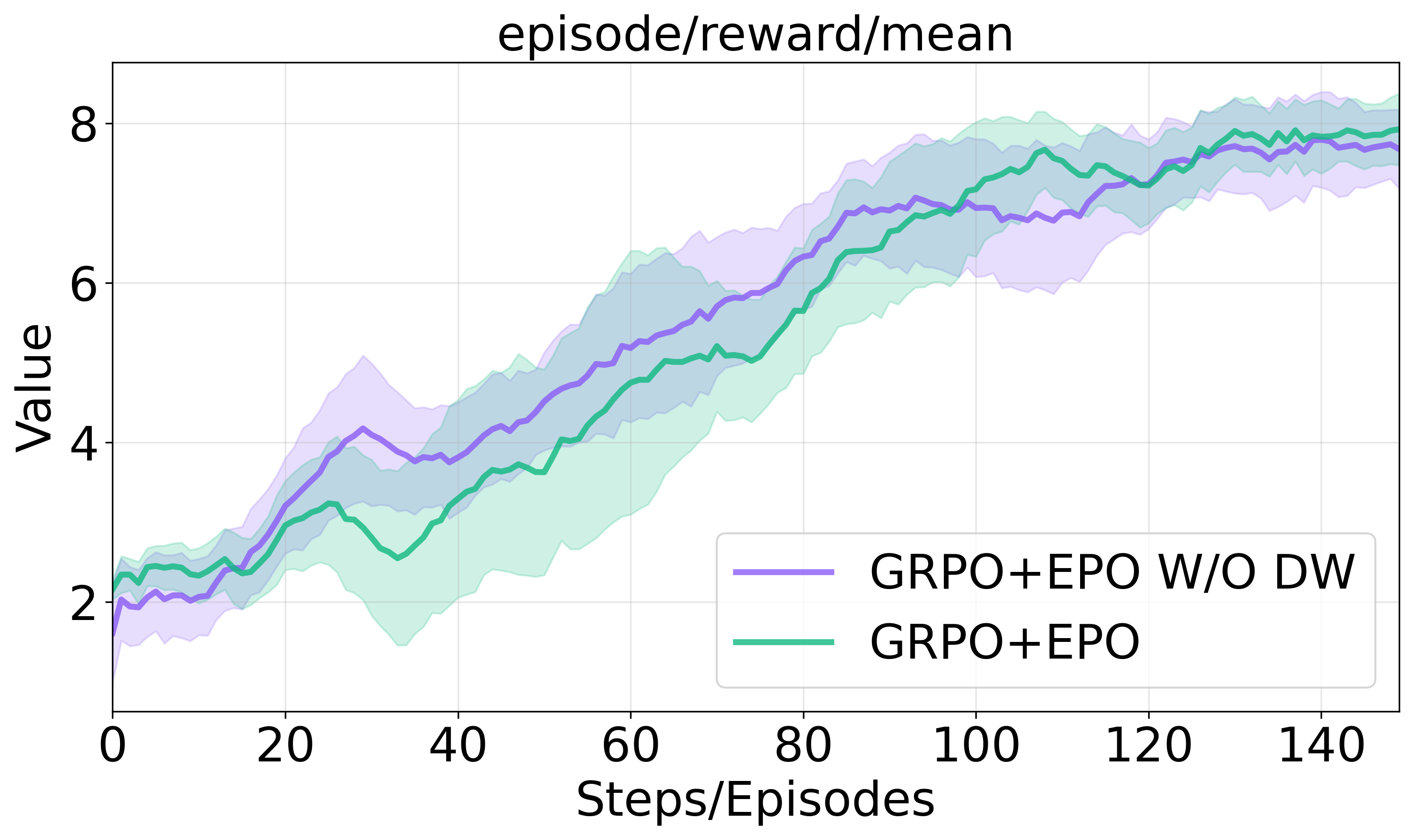}
%\caption{fig1}
\end{minipage}%
}%
\hfill 
\subfigure[\scriptsize IID Success Rate (ALFWorld)]{
\begin{minipage}[t]{0.32\linewidth}
\centering
\includegraphics[width=0.99\linewidth]{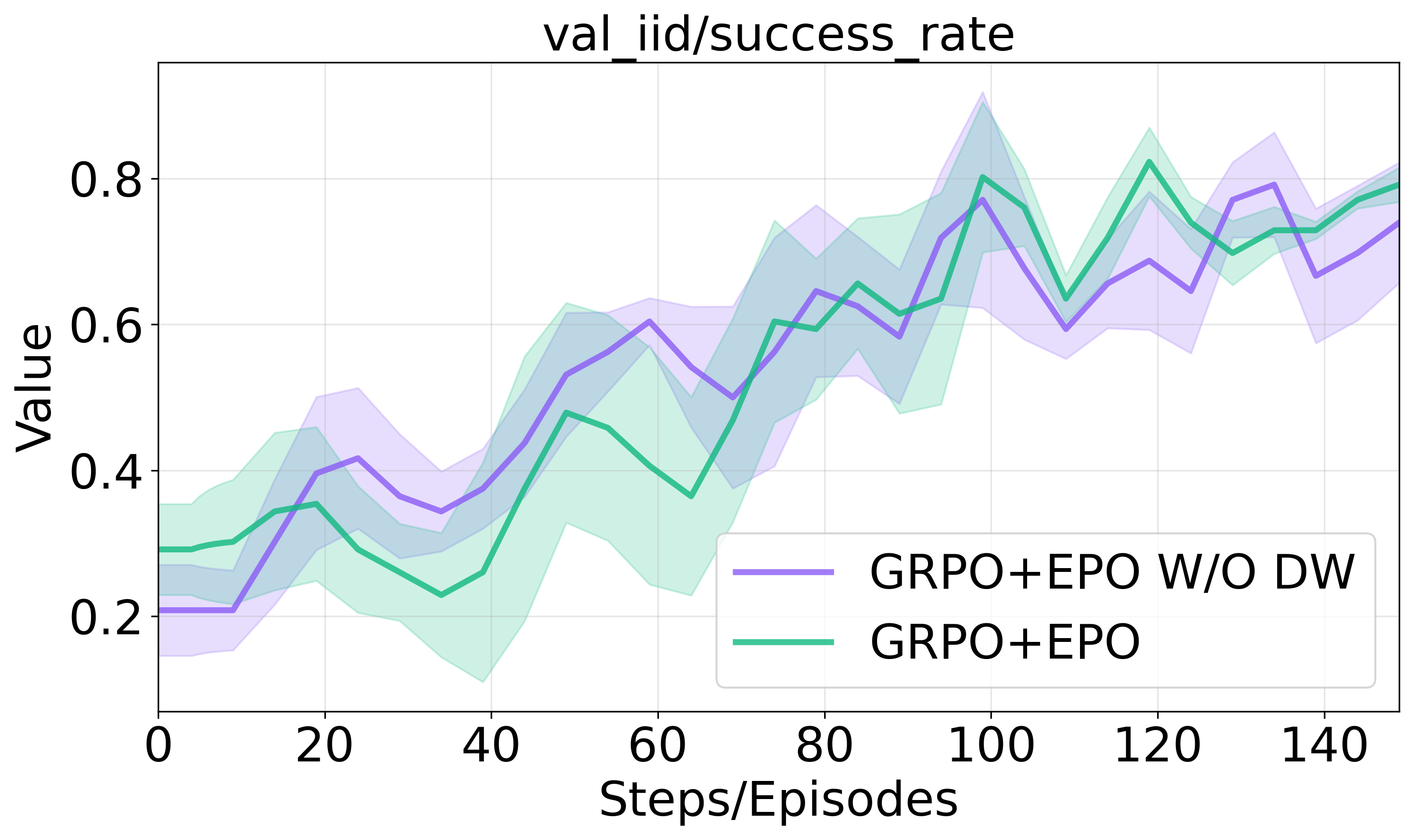}
%\caption{fig2}
\end{minipage}%
}% 
\hfill 
\subfigure[\scriptsize OOD Success Rate (ALFWorld)]{
\begin{minipage}[t]{0.32\linewidth}
\centering
\includegraphics[width=0.99\linewidth]{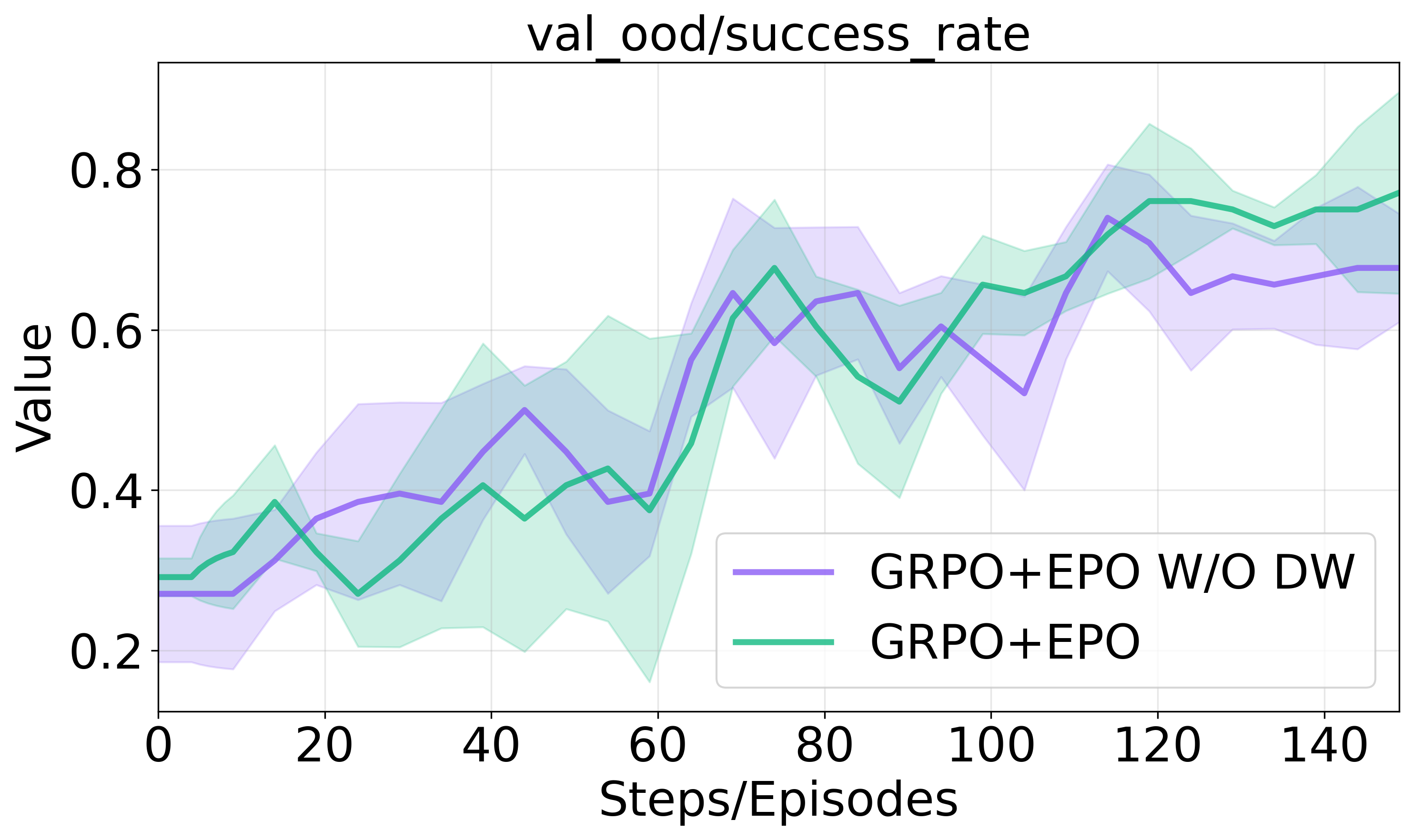}
%\caption{fig2}
\end{minipage}%
}% 

\centering
\vspace{-10pt}
% \captionsetup{font={small}}
\caption{Ablation studies on entropy regularization components. \textbf{(a-c)} ScienceWorld comparison of EPO versus EPO-Base without entropy smoothing, demonstrating that smoothing is essential for stable convergence in sparse reward settings. \textbf{(d-f)} ALFWorld comparison of EPO with dynamic $\beta_k$ versus EPO W/O DW using constant $\beta$, showing that adaptive weighting accelerates early training progress.}
\label{fig:abl study}
\vspace{-10pt}
\end{figure*}

\begin{figure*}[tb!]
\centering
\subfigure[EPO-Base vs EPO-Decay]{
\begin{minipage}[t]{0.32\linewidth}
\centering
\includegraphics[width=\linewidth]{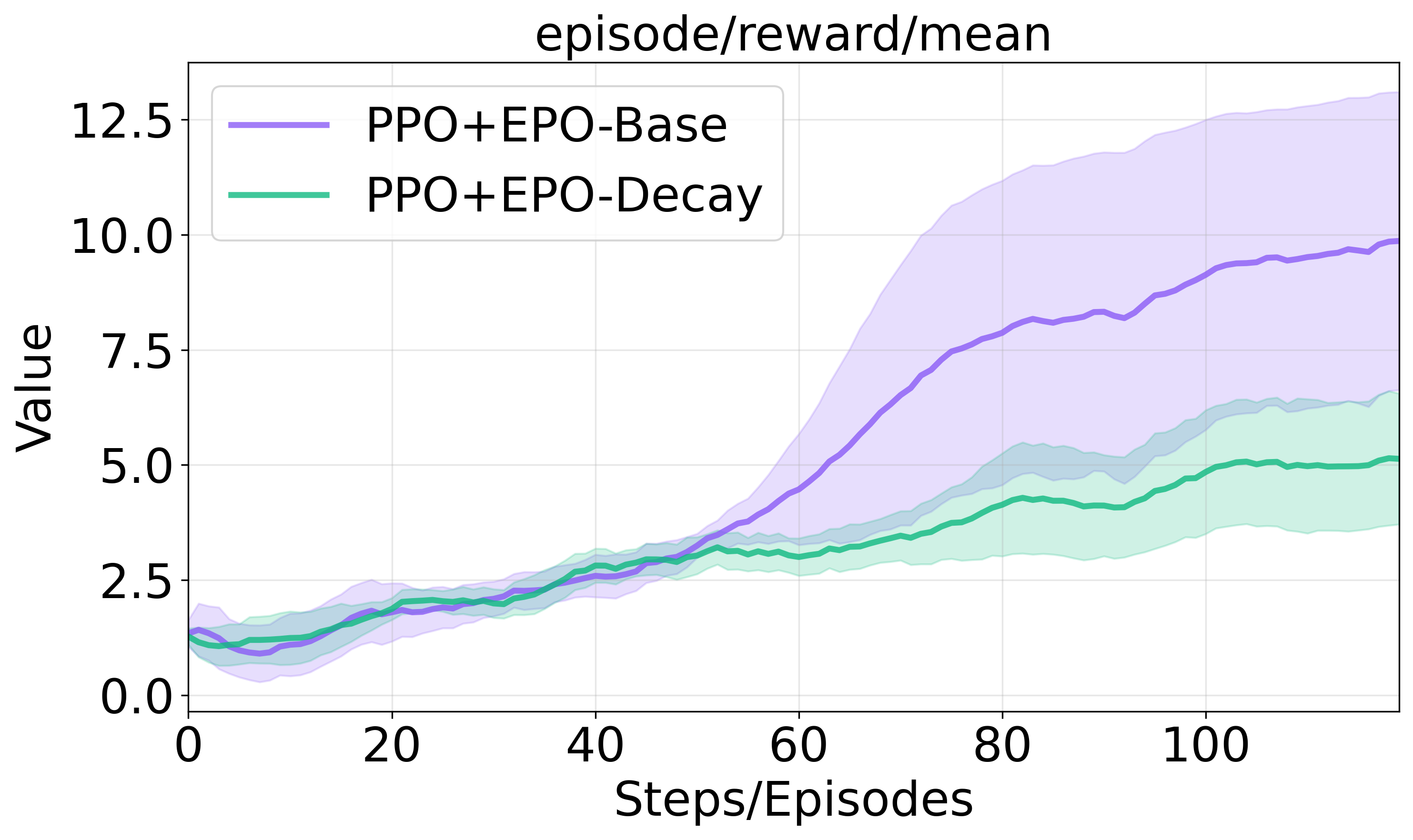}
\end{minipage}%
}%
\hfill
\subfigure[Early vs Late Entropy]{
\begin{minipage}[t]{0.32\linewidth}
\centering
\includegraphics[width=\linewidth]{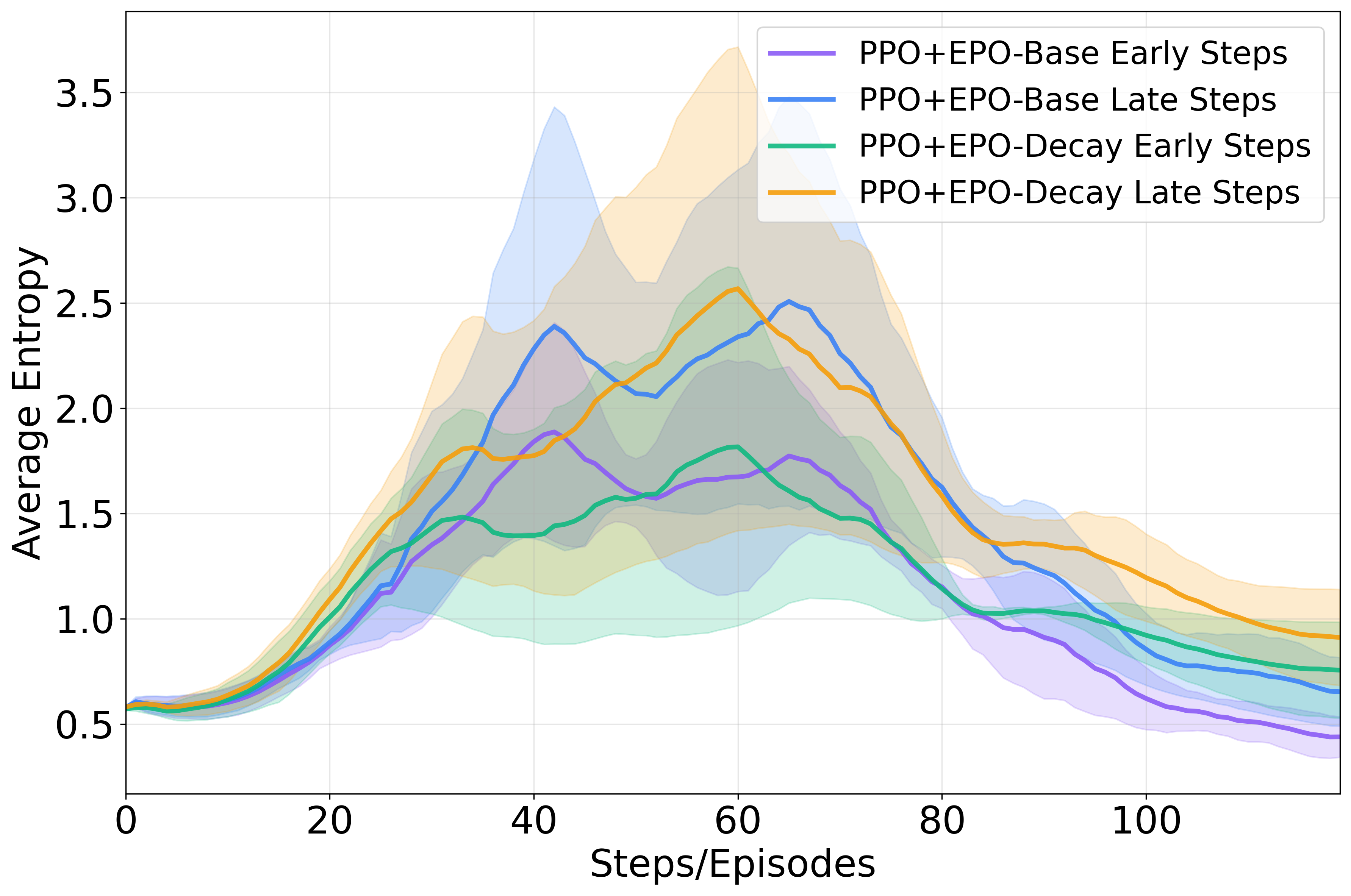}
\end{minipage}%
}%
\hfill
\subfigure[EPO vs EA Success Rate]{
\begin{minipage}[t]{0.32\linewidth}
\centering
\includegraphics[width=\linewidth]{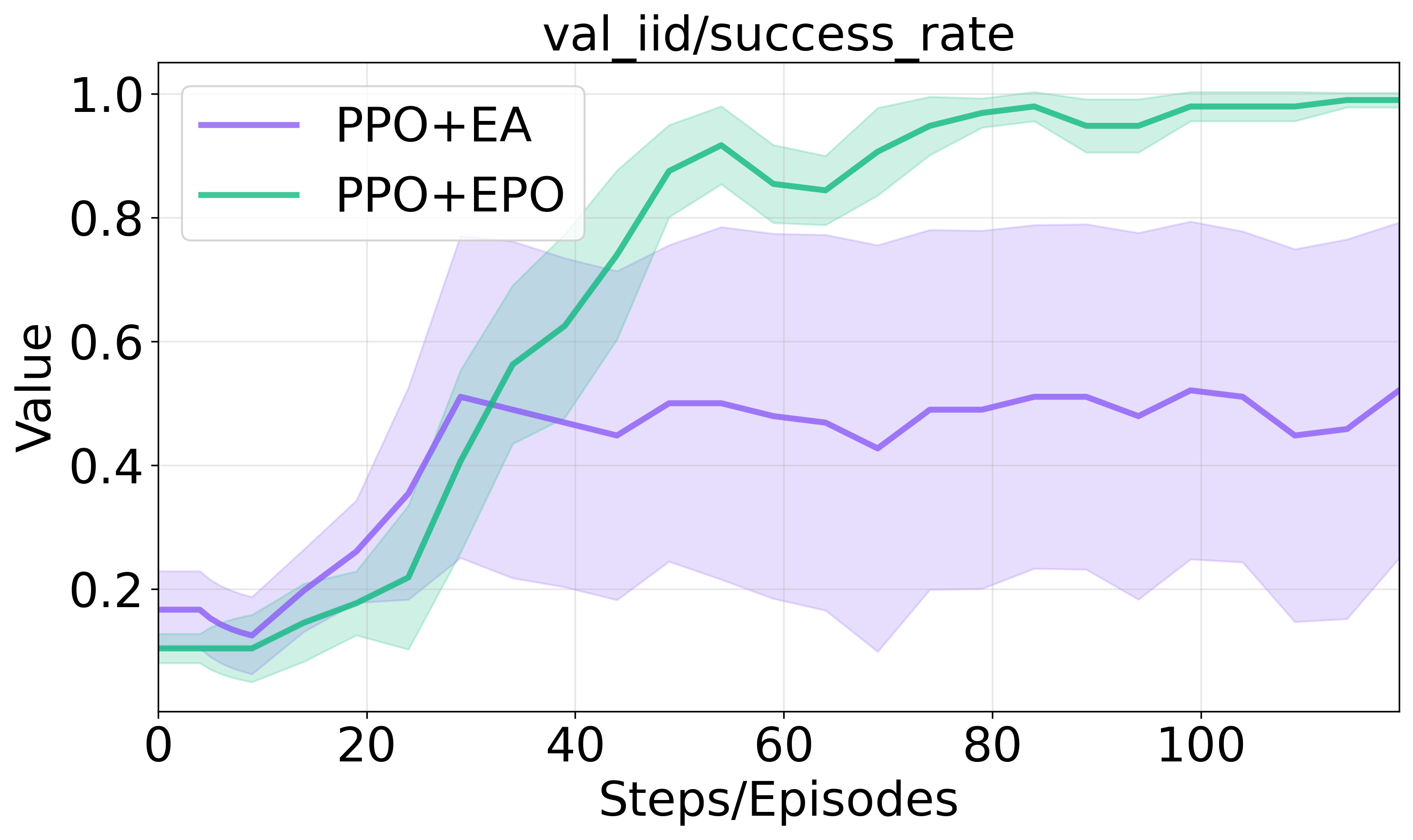}
\end{minipage}%
}%
\vspace{-10pt}
\caption{Model studies on ScienceWorld employing PPO with weighted entropy loss and entropy-based advantage shaping~\citep{cheng2025reasoning}.}
\label{fig:model study}
\vspace{-15pt}
\end{figure*}

\noindent The key pattern across both environments is the elimination of severe entropy oscillations, the characteristic signature of cascade failure (\autoref{fig:intro}(c)), that destabilize standard RL methods. By maintaining entropy within historical bounds through our smoothing regularizer, EPO transforms previously unstable sparse-reward scenarios into smoothly converging optimization, validating our framework's effectiveness in multi-turn LLM agent training.

\subsection{Ablation Study}\label{sec:abl}
\autoref{fig:abl study} presents ablation studies on two key components: the entropy smoothing regularizer and its dynamic weighting coefficient $\beta_k$, evaluated across two base methods (PPO, GRPO) and two environments (ScienceWorld, ALFWorld).

\noindent \textbf{Entropy Smoothing Regularizer.} On ScienceWorld, removing the smoothing regularizer (EPO-Base) severely degrades performance. EPO-Base exhibits delayed convergence with minimal rewards until step 40 and plateaus at 0.5--0.6 success rate, while full EPO achieves meaningful learning by step 20 and reaches 0.8--1.0 success rate, demonstrating a 50--60\% relative improvement. This gap directly reflects the cascade failure mechanism: ScienceWorld's sparse rewards induce severe entropy oscillations that compound across turns, which the smoothing regularizer effectively dampens by anchoring entropy within the corridor $[\kappa_l, \kappa_r]$.

\noindent \textbf{Dynamic Weighting Coefficient.} On ALFWorld, replacing the adaptive $\beta_k$ with a constant weight (EPO W/O DW) yields comparable final performance (0.7--0.8 success rate) but results in slower convergence during the initial 40--60 episodes and increased training variance. The adaptive schedule (\autoref{eq:beta_k}) automatically modulates regularization intensity across training phases, applying stronger smoothing early to establish controlled exploration and relaxing mid-training as the policy stabilizes. This improves convergence efficiency without sacrificing final performance. See \autoref{app:abl study} for detailed analysis.

% show the results without adaptive weights.

\subsection{Model Study}
% 1. early steps with more exploration, later steps with more exploitation doesn't work, with our oldest weights expers.

% 2. change the advantage function doesn't work for agent scenarios. like directly add entropy normalization.
% We compare consistent entropy regularization (PPO+EPO-Base) against a decaying schedule (PPO+EPO-Decay). This approach dynamically adjusts the exploration-exploitation balance by applying a high weight to the entropy loss in early steps and a progressively smaller weight in later steps to promote exploitation. Counter-intuitively, ~\autoref{fig:model study}(a) shows that decay consistently underperforms by prematurely suppressing early-turn exploration, locking agents into suboptimal strategies.
% We also compare EPO against the Entropy-based Advantage (EA) method from ~\citep{cheng2025reasoning}, ~\autoref{fig:model study}(b) shows EPO converges to near-perfect success rates ($\sim$1.0) while EA plateaus at 0.5--0.6. Unlike EA's detached entropy terms, EPO provides direct gradient signals $\nabla_\theta L^H(\theta)$ and temporal consistency through historical entropy smoothing.
% These studies reveal two key insights: (1) multi-turn sparse-reward tasks require sustained exploration rather than conventional exploration-to-exploitation scheduling, and (2) direct entropy optimization with temporal smoothing is superior to indirect advantage shaping for maintaining long-horizon policy stability. See ~\autoref{app:model study} for detailed analysis.
\textbf{Turn-wise vs. Uniform Entropy Weighting.}
We compare consistent entropy regularization (PPO+EPO-Base) against a decaying schedule (PPO+EPO-Decay), which applies higher entropy weights to early turns and lower weights to later turns. Counter-intuitively, \autoref{fig:model study}(a) shows that decay consistently underperforms by prematurely suppressing early-turn exploration, locking agents into suboptimal strategies. \autoref{fig:model study}(b) further reveals the mechanism: while both methods exhibit similar early-step entropy, the decay schedule causes late-step entropy to drop too rapidly, reducing the policy's ability to recover from suboptimal trajectories.

\textbf{Decoupled Regularization vs. Advantage Reshaping.}
We compare EPO against the Entropy-based Advantage (EA) method from~\citet{cheng2025reasoning}. As shown in \autoref{fig:model study}(c), EPO converges to near-perfect success rates ($\sim$1.0) while EA plateaus at 0.5--0.6. EA incorporates entropy directly into the advantage function, reshaping the policy gradient. In contrast, EPO decouples entropy as a separate regularization term, providing direct gradient signals while preserving the integrity of advantage-based credit assignment.

\textbf{Cumulative vs. Sliding Entropy Window.}
Our cumulative entropy window $\mathcal{W}_k$ is an intentional design choice for long-horizon training. With $K$ = 120-150 steps, the cumulative average adapts quickly when $k$ is small (weight $\sim 1/k$) and stabilizes later to anchor entropy against cascade failure. This design synergizes with the adaptive $\beta_k$: while $\beta_k$ modulates smoothing strength dynamically, a sliding window would create conflicting adaptation mechanisms.

\noindent These studies reveal two key insights: (1) multi-turn sparse-reward tasks require sustained exploration rather than conventional exploration-to-exploitation scheduling, and (2) decoupled entropy regularization with temporal smoothing is superior to advantage reshaping for maintaining long-horizon policy stability.

\section{Theoretical Analysis}\label{sec:theory}

We provide theoretical analysis to understand EPO's design and characterize when and why it outperforms standard entropy regularization. Our analysis reveals that EPO is a principled approach grounded in constrained optimization theory, with particular advantages in multi-turn settings where entropy 
instability compounds across turns. Full proofs are provided in \autoref{app:theory}.

\begin{restatable}[Exact Penalty Interpretation]{proposition}{propPenalty}\label{prop:penalty}
The EPO objective equals $V_{\lambda,\beta}^\pi = V^\pi + \lambda\mathcal{H}(\pi) - \rho_k \cdot \mathcal{P}(\pi)$,
where $\mathcal{P}(\pi)$ is the exact penalty for violating the entropy corridor $[\kappa_l\bar{\mathcal{H}}^{ref}, \kappa_r\bar{\mathcal{H}}^{ref}]$, and $\rho_k = \lambda\beta_k\alpha$ is an adaptive penalty weight.
\end{restatable}

This connects EPO to classical optimization: the smoothing loss is precisely the penalty for constraint violation, and the adaptive schedule implements curriculum-style enforcement—stronger early to guide entropy toward the corridor, relaxing later as stability is learned. While \autoref{prop:penalty} establishes EPO's principled foundation, it does not explain why EPO's benefits are disproportionately large in multi-turn settings. The key insight is how entropy errors propagate:

\begin{restatable}[Multi-Turn Entropy Stability]{proposition}{propMultiturn}\label{prop:multiturn}
Define cumulative entropy deviation $\Gamma_T(\pi) := \sum_{t=1}^{T} |\bar{\mathcal{H}}_t(\pi) - \bar{\mathcal{H}}^{ref}|$ over $T$ turns.

\noindent \textbf{EPO:} Under the corridor constraint, $\Gamma_T(\pi_\theta) = O(T)$.

\noindent \textbf{Standard entropy regularization:} Without corridor enforcement, $\Gamma_T(\pi) = O(T^2)$ in the worst case.
\end{restatable}

Without the corridor, entropy drift at turn $t$ affects turn $t+1$'s context, which affects turn $t+2$---errors compound quadratically. EPO's corridor acts as a guardrail, correcting deviations immediately. The advantage grows with dialogue length: $5.5\times$ for $T=10$, $10.5\times$ for $T=20$. The $O(T^2)$ worst-case arises under conditions common in LLM training: sparse end-of-dialogue rewards, entropy collapse dynamics, and autoregressive context dependence (\autoref{rem:cascade_conditions}).

\textbf{Why Stability Improves Convergence.}
The $O(T)$ vs $O(T^2)$ gap translates to convergence through two mechanisms: (i) \emph{gradient variance}: entropy collapse causes extreme importance weights, increasing gradient noise; and (ii) \emph{trajectory diversity}: collapsed entropy reduces the trajectories that reach informative rewards. We validate empirically that EPO achieves lower gradient variance and faster convergence (\autoref{sec:exper}).

\begin{restatable}[Performance Bound]{proposition}{propBound}\label{prop:main}
Under mild assumptions (\autoref{app:theory}), EPO achieves:
\begin{equation}
\begin{small}
V^{\pi^*} - V^{\pi_\theta} \leq \underbrace{\frac{|\mathcal{D}|^2\epsilon^2}{2\lambda(1-\alpha\beta_k) C}}_{\text{Optimization}} + \underbrace{\lambda H\log\frac{|\mathcal{A}|}{|\mathcal{A}_H^*|^{1/H}}}_{\text{Entropy Bias}} + \underbrace{\lambda\beta_k\Delta^{\text{smooth}}_{\pi^*}}_{\text{Stability Cost}}
\end{small}
\end{equation}
\end{restatable}

\vspace{-5pt}
The bound decomposes into optimization error (decreases with training), entropy bias (standard in entropy-regularized RL), and stability cost (price for corridor enforcement). Crucially, this cost often vanishes:

\begin{restatable}[Zero Stability Cost]{proposition}{propZeroCost}\label{prop:zero_cost}
If $\pi^*$ has moderate entropy ($\mathcal{H}(\pi^*|s) \in [\kappa_l\bar{\mathcal{H}}^{ref}, \kappa_r\bar{\mathcal{H}}^{ref}]$), then $\Delta^{\text{smooth}}_{\pi^*} = 0$---EPO matches standard bounds while providing stability guarantees for free.
\end{restatable}

In sparse-reward multi-turn tasks, optimal policies typically \emph{require} moderate entropy to discover good trajectories, naturally satisfying this condition. Thus EPO provides stability "for free" precisely when cascade failure threatens most.

\section{Conclusion}
In this work, we identified and addressed the exploration-exploitation cascade failure, a fundamental challenge unique to training LLM agents in multi-turn environments with sparse rewards. Our proposed EPO framework addresses this through three synergistic mechanisms: trajectory-aware entropy computation, entropy smoothing regularization, and adaptive phase-based weighting. Together, these components prevent the severe entropy oscillations that destabilize training under shared policy parameters. Empirical results demonstrate up to 152\% performance improvement on ScienceWorld and 19.8\% on ALFWorld, transforming previously unstable sparse-reward scenarios into smoothly converging optimization. This work establishes that multi-turn LLM agent training requires fundamentally different entropy control than traditional RL, opening new directions for effective training methods for agentic LLMs.

\clearpage

\section*{Impact Statement}
This paper presents work whose goal is to advance the field of machine learning, specifically improving the training stability of LLM agents in multi-turn environments. Our method enables more reliable training of agents for beneficial applications such as scientific experimentation, household task assistance, and interactive problem-solving. 

As with any advancement in autonomous agent capabilities, there exists potential for misuse if deployed without appropriate safeguards. However, our contribution is methodological in nature, focusing on training dynamics rather than enabling fundamentally new capabilities. We encourage practitioners to apply standard responsible AI practices when deploying agents trained with our method.

There are many potential societal consequences of our work, none of which we feel must be specifically highlighted here beyond the considerations common to research on LLM agents.

\nocite{langley00}

\bibliography{icml2026}
\bibliographystyle{icml2026}

%%%%%%%%%%%%%%%%%%%%%%%%%%%%%%%%%%%%%%%%%%%%%%%%%%%%%%%%%%%%%%%%%%%%%%%%%%%%%%%
%%%%%%%%%%%%%%%%%%%%%%%%%%%%%%%%%%%%%%%%%%%%%%%%%%%%%%%%%%%%%%%%%%%%%%%%%%%%%%%
% APPENDIX
%%%%%%%%%%%%%%%%%%%%%%%%%%%%%%%%%%%%%%%%%%%%%%%%%%%%%%%%%%%%%%%%%%%%%%%%%%%%%%%
%%%%%%%%%%%%%%%%%%%%%%%%%%%%%%%%%%%%%%%%%%%%%%%%%%%%%%%%%%%%%%%%%%%%%%%%%%%%%%%
\onecolumn
\clearpage
\tableofcontents

\clearpage
\appendix

\appendix
\section{Theoretical Analysis: Full Proofs}\label{app:theory}

\subsection{Notation and Setup}

We consider multi-turn dialogue generation as a finite-horizon MDP. Let $s_0$ denote the initial query, $H$ the maximum horizon (total tokens), and $T$ the number of dialogue turns. Each turn $t \in [T]$ generates $L_t$ tokens. The policy $\pi_\theta(a|s)$ maps states (conversation history) to distributions over tokens (actions).

\textbf{Standard Entropy-Regularized Objective:}
\begin{equation}
V_\lambda^\pi(s_0) = \mathbb{E}_\pi\left[\sum_{h=0}^{H-1} r_h + \lambda \mathcal{H}(\pi(\cdot|s_h)) \;\Big|\; s_0\right]
\end{equation}

\textbf{EPO Objective:}
\begin{equation}\label{eq:epo_objective}
V_{\lambda,\beta}^\pi(s_0) = V_\lambda^\pi(s_0) - \lambda\beta_k \cdot L^{\text{smooth}}_{\bar{\mathcal{H}}^{ref}}(\theta)
\end{equation}
where the smoothing loss is:
\begin{equation}
L^{\text{smooth}}_{\bar{\mathcal{H}}^{ref}}(\theta) = \frac{\alpha}{|\mathcal{B}|} \sum_{(s,a) \in \mathcal{B}} \mathcal{P}(\mathcal{H}(\pi_\theta(\cdot|s)); \bar{\mathcal{H}}^{ref})
\end{equation}
with corridor penalty:
\begin{equation}
\mathcal{P}(\mathcal{H}; \bar{\mathcal{H}}^{ref}) = [\kappa_l \bar{\mathcal{H}}^{ref} - \mathcal{H}]_+ + [\mathcal{H} - \kappa_r \bar{\mathcal{H}}^{ref}]_+
\end{equation}

\subsection{Assumptions}

\begin{assumption}[Fixed Entropy Reference]\label{assump:reference}
The smoothing loss is evaluated against a fixed reference $\bar{\mathcal{H}}^{ref} > 0$, representing the target entropy level.
\end{assumption}

\begin{assumption}[Entropy Corridor Constraint]\label{assump:corridor}
The learned policy $\pi_\theta$ maintains entropy within the corridor:
$$\kappa_l \bar{\mathcal{H}}^{ref} \leq \mathcal{H}(\pi_\theta(\cdot|s)) \leq \kappa_r \bar{\mathcal{H}}^{ref}$$
for all reachable states $s \in \mathcal{S}(s_0)$.
\end{assumption}

\begin{assumption}[Policy Regularity]\label{assump:regularity}
The policy gradient satisfies $\|\nabla V^{\pi_\theta}_{\lambda,\beta}(\mathcal{D})\| \leq \epsilon$ and $\min_{a,s} \pi_\theta(a|s) \geq \delta_\pi > 0$.
\end{assumption}

\begin{assumption}[Bounded Smoothing Strength]\label{assump:alpha}
The penalty weight satisfies $\alpha\beta_{\max} < 1$, where $\beta_{\max} = 1 + e^0 = 2$. Equivalently, $\alpha < 0.5$.
\end{assumption}

\begin{remark}[Justification]
\autoref{assump:corridor} captures EPO's design intent---the smoothing penalty enforces corridor compliance. We verify this empirically in \autoref{sec:exper}. \autoref{assump:alpha} ensures the EPO objective remains concave; our experiments use $\alpha = 0.1 < 0.5$.
\end{remark}

\subsection{Key Definitions}

\begin{definition}[Stability Cost]\label{def:stability_cost}
For any policy $\tilde{\pi}$:
$$\Delta^{\text{smooth}}_{\tilde{\pi}}(s_0) := \mathbb{E}_{\tilde{\pi}}\left[L^{\text{smooth}}_{\bar{\mathcal{H}}^{ref}}(\tilde{\theta}) \;\Big|\; s_0\right] \geq 0$$
Under \autoref{assump:corridor}, $\Delta^{\text{smooth}}_{\pi_\theta}(s_0) = 0$.
\end{definition}

\begin{definition}[Concentrability Coefficient]\label{def:concentrability}
For initial state $s_0$ and policy $\pi_\theta$:
\begin{equation}
C_{\lambda,\beta}^{\pi_\theta}(s_0) = C^2_d \cdot \min_{h,s \in \mathcal{S}(s_0)} \mathbb{P}^{\pi_\theta}_h(s) \left(\min_{a} \pi_\theta(a|s)\right)^2 \cdot \min_{h,s \in \mathcal{S}(s_0)} \frac{\mathbb{P}^{\pi_\theta}_h(s)}{\mathbb{P}^{\pi^*_{\lambda,\beta}}_h(s|s_0)}
\end{equation}
where $C_d = (|\mathcal{A}|H)^{-1/2}$ and $\mathbb{P}^{\pi}_h(s)$ is the state visitation probability at step $h$.
\end{definition}

\begin{definition}[EPO-Optimal Policy]\label{def:epo_optimal}
$$\pi_{\lambda,\beta}^* := \arg\max_{\pi \in \Pi} V_{\lambda,\beta}^\pi(\mathcal{D})$$
Under \autoref{assump:alpha}, the objective is concave, so this is well-defined.
\end{definition}

\begin{definition}[Cumulative Entropy Deviation]\label{def:cumulative_deviation}
For $T$-turn dialogue with average entropy $\bar{\mathcal{H}}_t$ at turn $t$:
$$\Gamma_T(\pi) := \sum_{t=1}^{T} |\bar{\mathcal{H}}_t(\pi) - \bar{\mathcal{H}}^{ref}|$$
\end{definition}

\subsection{Preliminary Lemmas}

\begin{lemma}[Penalty Function Properties]\label{lem:penalty_props}
The corridor penalty $\mathcal{P}(\mathcal{H}; \bar{\mathcal{H}}^{ref}) = [\kappa_l \bar{\mathcal{H}}^{ref} - \mathcal{H}]_+ + [\mathcal{H} - \kappa_r \bar{\mathcal{H}}^{ref}]_+$ satisfies:
\begin{enumerate}
    \item Non-negativity: $\mathcal{P} \geq 0$
    \item Corridor property: $\mathcal{P} = 0$ if and only if $\mathcal{H} \in [\kappa_l \bar{\mathcal{H}}^{ref}, \kappa_r \bar{\mathcal{H}}^{ref}]$
    \item Lipschitz continuity: $|\mathcal{P}(\mathcal{H}_1) - \mathcal{P}(\mathcal{H}_2)| \leq |\mathcal{H}_1 - \mathcal{H}_2|$
\end{enumerate}
\end{lemma}

\begin{proof}
Properties (1) and (2) follow from the definition of $[\cdot]_+ = \max(0, \cdot)$. For (3), both $[\kappa_l \bar{\mathcal{H}}^{ref} - \mathcal{H}]_+$ and $[\mathcal{H} - \kappa_r \bar{\mathcal{H}}^{ref}]_+$ are Lipschitz with constant 1, and at most one is active at any $\mathcal{H}$.
\end{proof}

\begin{lemma}[Gradient Towards Corridor]\label{lem:gradient}
For entropy outside the corridor:
$$\nabla_\theta L^{\text{smooth}} = \begin{cases}
-\alpha \nabla_\theta \mathcal{H}(\pi_\theta) & \text{if } \mathcal{H} < \kappa_l\bar{\mathcal{H}}^{ref} \quad \text{(increases entropy)} \\
0 & \text{if } \mathcal{H} \in [\kappa_l\bar{\mathcal{H}}^{ref}, \kappa_r\bar{\mathcal{H}}^{ref}] \\
+\alpha \nabla_\theta \mathcal{H}(\pi_\theta) & \text{if } \mathcal{H} > \kappa_r\bar{\mathcal{H}}^{ref} \quad \text{(decreases entropy)}
\end{cases}$$
\end{lemma}

\begin{proof}
Taking gradients of $\mathcal{P}(\mathcal{H}; \bar{\mathcal{H}}^{ref})$:
$$\nabla_\theta \mathcal{P} = \begin{cases}
-\nabla_\theta \mathcal{H} & \text{if } \mathcal{H} < \kappa_l\bar{\mathcal{H}}^{ref} \\
0 & \text{if } \mathcal{H} \in [\kappa_l\bar{\mathcal{H}}^{ref}, \kappa_r\bar{\mathcal{H}}^{ref}] \\
+\nabla_\theta \mathcal{H} & \text{if } \mathcal{H} > \kappa_r\bar{\mathcal{H}}^{ref}
\end{cases}$$
The result follows from $L^{\text{smooth}} = \alpha \cdot \mathbb{E}[\mathcal{P}]$.
\end{proof}

\begin{lemma}[Performance Difference~\citep{kakade2002approximately}]\label{lem:perf_diff}
For any policies $\pi, \pi'$:
$$V_h^\pi(s) - V_h^{\pi'}(s) = \mathbb{E}_{\pi}\left[\sum_{t=h}^{H-1} A_t^{\pi'}(s_t, a_t) \;\Big|\; s_h = s\right]$$
where $A_t^{\pi'}(s,a) = Q_t^{\pi'}(s,a) - V_t^{\pi'}(s)$ is the advantage function.
\end{lemma}

\begin{lemma}[Entropy Bias~\citep{shen2025entropy}]\label{lem:entropy_bias}
$$V^{\pi^*}(s_0) - V^{\pi_\theta}(s_0) \leq V_\lambda^{\pi_\lambda^*}(s_0) - V_\lambda^{\pi_\theta}(s_0) + \lambda H\log \frac{|\mathcal{A}|}{|\mathcal{A}_H^*(s_0)|^{1/H}}$$
where $|\mathcal{A}_H^*(s_0)|$ is the number of optimal action sequences from $s_0$.
\end{lemma}

\subsection{Main Results}

\subsubsection{Exact Penalty Interpretation}

\propPenalty*

\begin{proof}
Expanding the EPO objective:
\begin{align}
V_{\lambda,\beta}^\pi &= V^\pi + \lambda\mathcal{H}(\pi) - \lambda\beta_k \cdot L^{\text{smooth}}_{\bar{\mathcal{H}}^{ref}} \\
&= V^\pi + \lambda\mathcal{H}(\pi) - \lambda\beta_k \cdot \frac{\alpha}{|\mathcal{B}|} \sum_{(s,a) \in \mathcal{B}} \mathcal{P}(\mathcal{H}(\pi|s); \bar{\mathcal{H}}^{ref}) \\
&= V^\pi + \lambda\mathcal{H}(\pi) - \underbrace{\lambda\beta_k\alpha}_{\rho_k} \cdot \underbrace{\mathbb{E}_{s \sim \pi}[\mathcal{P}(\mathcal{H}(\pi|s); \bar{\mathcal{H}}^{ref})]}_{\mathcal{P}(\pi)}
\end{align}
This matches the exact penalty formulation for the constraint $\mathcal{H}(\pi|s) \in [\kappa_l\bar{\mathcal{H}}^{ref}, \kappa_r\bar{\mathcal{H}}^{ref}]$.
\end{proof}

\begin{remark}[Connection to Classical Optimization]
The exact penalty method~\citep{kakade2002approximately} guarantees that for sufficiently large $\rho$, the unconstrained optimum coincides with the constrained optimum. EPO's adaptive schedule $\rho_k = \lambda\beta_k\alpha$ provides curriculum-style enforcement: weaker early (exploration), stronger later (exploitation).
\end{remark}

\subsubsection{Multi-Turn Stability}

\propMultiturn*

\begin{proof}
\textbf{Part (i): EPO bound.}

Under \autoref{assump:corridor}, for all turns $t$ and states $s$:
$$\mathcal{H}(\pi_\theta(\cdot|s)) \in [\kappa_l\bar{\mathcal{H}}^{ref}, \kappa_r\bar{\mathcal{H}}^{ref}]$$

The average entropy at turn $t$ satisfies:
$$\bar{\mathcal{H}}_t(\pi_\theta) = \frac{1}{L_t}\sum_{h \in \text{turn } t} \mathcal{H}(\pi_\theta(\cdot|s_h)) \in [\kappa_l\bar{\mathcal{H}}^{ref}, \kappa_r\bar{\mathcal{H}}^{ref}]$$

Therefore:
$$|\bar{\mathcal{H}}_t(\pi_\theta) - \bar{\mathcal{H}}^{ref}| \leq \max\{|1 - \kappa_l|, |\kappa_r - 1|\} \cdot \bar{\mathcal{H}}^{ref} \leq \frac{\kappa_r - \kappa_l}{2} \cdot \bar{\mathcal{H}}^{ref} := \Delta_{\max}$$

Summing over $T$ turns:
$$\Gamma_T(\pi_\theta) = \sum_{t=1}^{T} |\bar{\mathcal{H}}_t(\pi_\theta) - \bar{\mathcal{H}}^{ref}| \leq T \cdot \Delta_{\max} = O(T)$$

\textbf{Part (ii): Entropy regularization worst case.}

Without corridor enforcement, entropy at turn $t$ depends on context from turns $1, \ldots, t-1$. Consider the cascade failure scenario where each turn introduces drift $\delta$:
$$\bar{\mathcal{H}}_t - \bar{\mathcal{H}}^{ref} = \sum_{s=1}^{t} \delta_s$$

If drifts are systematic (e.g., consistently decreasing entropy), $|\bar{\mathcal{H}}_t - \bar{\mathcal{H}}^{ref}| \leq t \cdot \delta_{\max}$.

Summing:
$$\Gamma_T(\pi) \leq \sum_{t=1}^{T} t \cdot \delta_{\max} = \frac{T(T+1)}{2} \cdot \delta_{\max} = O(T^2)$$
\end{proof}

\begin{remark}[Conditions for Cascade Failure]\label{rem:cascade_conditions}
The $O(T^2)$ worst-case arises under specific but common conditions in multi-turn LLM training:

\textbf{(i) Sparse rewards:} When rewards are only observed at dialogue end (e.g., task success/failure), there is no per-turn gradient signal to correct entropy drift early. Each turn's drift goes unchecked until final reward, by which point deviations have compounded.

\textbf{(ii) Entropy collapse dynamics:} Once entropy begins decreasing, the softmax policy assigns increasingly extreme probabilities. This reduces exploration, which reduces trajectory diversity, which reduces the probability of discovering reward signal, which further reduces entropy—a positive feedback loop.

\textbf{(iii) Context-dependent amplification:} In autoregressive generation, turn $t$'s output becomes turn $t+1$'s context. Low-entropy (confident) outputs at turn $t$ bias the model toward similar low-entropy behavior at turn $t+1$, causing drift to accumulate rather than cancel.

These conditions are prevalent in multi-turn LLM training, so the $O(T^2)$ bound captures realistic failure modes rather than pathological edge cases.
\end{remark}

\begin{remark}[Practical Implication]
The $O(T)$ vs $O(T^2)$ gap implies EPO's advantage grows with dialogue length. For $T=10$ turns, the worst-case ratio is $\frac{T(T+1)/2}{T} = \frac{T+1}{2} = 5.5\times$. For $T=20$, this becomes $10.5\times$. This explains why EPO's empirical benefits are most pronounced in long multi-turn tasks (\autoref{sec:exper}).
\end{remark}

\subsubsection{Connection Between Entropy Stability and Convergence}

\begin{remark}[Why Entropy Stability Improves Convergence]\label{rem:gamma_to_perf}
While we do not formally prove a bound relating $\Gamma_T$ to suboptimality, we identify two mechanisms through which entropy stability improves convergence:

\textbf{Mechanism 1: Importance Weight Variance.}
Policy gradient estimators use importance weights $w = \pi_\theta(a|s)/\pi_{\text{old}}(a|s)$. When entropy collapses ($\mathcal{H} \to 0$), policies become near-deterministic, causing $w \to \infty$ for some actions and $w \to 0$ for others. This increases gradient variance, slowing convergence. EPO's corridor constraint bounds entropy away from zero, limiting weight variance.

\textbf{Mechanism 2: Trajectory Diversity.}
In multi-turn generation with sparse rewards, gradient signal comes from trajectories that reach reward. Collapsed entropy reduces trajectory diversity exponentially with horizon: if per-step entropy drops by factor $c < 1$, trajectory diversity drops by $c^T$. The corridor constraint maintains per-step entropy, preserving trajectory diversity throughout training.

\textbf{Empirical Validation.}
Rather than formalizing these mechanisms into a bound with unverifiable constants, we directly measure the effect in \autoref{sec:exper}: EPO achieves lower gradient variance and faster convergence compared to standard entropy regularization, consistent with this intuition.
\end{remark}

\subsubsection{Performance Bound}

\propBound*

\begin{proof}
\textbf{Step 1: Decomposition through EPO-optimal policy.}
\begin{equation}
V^{\pi^*}(s_0) - V^{\pi_\theta}(s_0) = \underbrace{[V^{\pi^*} - V_{\lambda,\beta}^{\pi_{\lambda,\beta}^*}]}_{(I)} + \underbrace{[V_{\lambda,\beta}^{\pi_{\lambda,\beta}^*} - V_{\lambda,\beta}^{\pi_\theta}]}_{(II)} + \underbrace{[V_{\lambda,\beta}^{\pi_\theta} - V^{\pi_\theta}]}_{(III)}
\end{equation}

\textbf{Step 2: Bounding Term (II).}

Under \autoref{assump:alpha}, the EPO objective has effective entropy coefficient:
$$\lambda_{\text{eff}} = \lambda(1 - \alpha\beta_k) > 0$$

Since $\alpha\beta_k < 1$, the objective $V_{\lambda,\beta}$ remains concave in policy parameters. The gradient-to-performance bound from~\citet{shen2025entropy} applies:
\begin{equation}
(II) \leq \frac{|\mathcal{D}|^2}{2\lambda_{\text{eff}} \cdot C_{\lambda,\beta}^{\pi_\theta}(s_0)}\epsilon^2 = \frac{|\mathcal{D}|^2\epsilon^2}{2\lambda(1-\alpha\beta_k) C_{\lambda,\beta}^{\pi_\theta}(s_0)}
\end{equation}

\textbf{Step 3: Bounding Term (III).}

Under \autoref{assump:corridor}, $L^{\text{smooth}}(\theta) = 0$ by \autoref{lem:penalty_props}(2). Thus:
\begin{equation}
(III) = V_{\lambda,\beta}^{\pi_\theta}(s_0) - V^{\pi_\theta}(s_0) = \lambda\mathcal{H}(\pi_\theta|s_0) - 0 = \lambda\mathcal{H}(\pi_\theta|s_0)
\end{equation}

\textbf{Step 4: Bounding Term (I).}

By definition of $\pi_{\lambda,\beta}^*$ as the maximizer:
$$V_{\lambda,\beta}^{\pi_{\lambda,\beta}^*}(s_0) \geq V_{\lambda,\beta}^{\pi^*}(s_0)$$

Expanding $V_{\lambda,\beta}^{\pi^*}$:
$$V_{\lambda,\beta}^{\pi^*}(s_0) = V^{\pi^*}(s_0) + \lambda\mathcal{H}(\pi^*|s_0) - \lambda\beta_k\Delta^{\text{smooth}}_{\pi^*}(s_0)$$

Therefore:
\begin{equation}
(I) = V^{\pi^*} - V_{\lambda,\beta}^{\pi_{\lambda,\beta}^*} \leq V^{\pi^*} - V_{\lambda,\beta}^{\pi^*} = -\lambda\mathcal{H}(\pi^*|s_0) + \lambda\beta_k\Delta^{\text{smooth}}_{\pi^*}(s_0)
\end{equation}

\textbf{Step 5: Combining (I) + (III).}
\begin{align}
(I) + (III) &\leq \lambda\mathcal{H}(\pi_\theta|s_0) - \lambda\mathcal{H}(\pi^*|s_0) + \lambda\beta_k\Delta^{\text{smooth}}_{\pi^*}(s_0) \\
&= \lambda(\mathcal{H}(\pi_\theta|s_0) - \mathcal{H}(\pi^*|s_0)) + \lambda\beta_k\Delta^{\text{smooth}}_{\pi^*}(s_0)
\end{align}

Using entropy bounds: $\mathcal{H}(\pi_\theta|s_0) \leq H\log|\mathcal{A}|$ and $\mathcal{H}(\pi^*|s_0) \geq \log|\mathcal{A}_H^*(s_0)|$:
\begin{equation}
(I) + (III) \leq \lambda H\log\frac{|\mathcal{A}|}{|\mathcal{A}_H^*(s_0)|^{1/H}} + \lambda\beta_k\Delta^{\text{smooth}}_{\pi^*}(s_0)
\end{equation}

\textbf{Step 6: Final bound.}

Combining Steps 2 and 5:
\begin{equation}
V^{\pi^*}(s_0) - V^{\pi_\theta}(s_0) \leq \frac{|\mathcal{D}|^2\epsilon^2}{2\lambda(1-\alpha\beta_k) C_{\lambda,\beta}^{\pi_\theta}(s_0)} + \lambda H\log\frac{|\mathcal{A}|}{|\mathcal{A}_H^*(s_0)|^{1/H}} + \lambda\beta_k\Delta^{\text{smooth}}_{\pi^*}(s_0)
\end{equation}
\end{proof}

\subsubsection{Zero Stability Cost Condition}

\propZeroCost*

\begin{proof}
\textbf{Part 1: Zero stability cost.}

Under the stated condition, $\mathcal{H}(\pi^*(\cdot|s)) \in [\kappa_l\bar{\mathcal{H}}^{ref}, \kappa_r\bar{\mathcal{H}}^{ref}]$ for all $s \in \mathcal{S}(s_0)$.

By \autoref{lem:penalty_props}(2), $\mathcal{P}(\mathcal{H}(\pi^*|s); \bar{\mathcal{H}}^{ref}) = 0$ for all such states.

Therefore:
$$L^{\text{smooth}}_{\bar{\mathcal{H}}^{ref}}(\theta^*) = \alpha \cdot \mathbb{E}_{s \sim \pi^*}[\mathcal{P}(\mathcal{H}(\pi^*|s); \bar{\mathcal{H}}^{ref})] = 0$$

And:
$$\Delta^{\text{smooth}}_{\pi^*}(s_0) = \mathbb{E}_{\pi^*}[L^{\text{smooth}}|s_0] = 0$$

\textbf{Part 2: Matching bound.}

Substituting $\Delta^{\text{smooth}}_{\pi^*} = 0$ into \autoref{prop:main}:
$$V^{\pi^*} - V^{\pi_\theta} \leq \frac{|\mathcal{D}|^2\epsilon^2}{2\lambda(1-\alpha\beta_k) C} + \lambda H\log\frac{|\mathcal{A}|}{|\mathcal{A}_H^*|^{1/H}}$$

For small $\alpha$ (e.g., $\alpha = 0.1$), the factor $(1-\alpha\beta_k)^{-1} \approx 1.25$, so the optimization term is only slightly larger than the standard entropy-regularized bound.

\textbf{Part 3: Additional guarantees.}

Even with $\Delta^{\text{smooth}}_{\pi^*} = 0$, EPO provides stability via \autoref{lem:gradient}: deviations from the corridor are immediately corrected, preventing cascade failure (\autoref{prop:multiturn}).
\end{proof}

\begin{remark}[When Does the Condition Hold?]
The condition $\mathcal{H}(\pi^*|s) \in [\kappa_l\bar{\mathcal{H}}^{ref}, \kappa_r\bar{\mathcal{H}}^{ref}]$ holds when:
\begin{enumerate}
    \item The task admits multiple near-optimal solutions, requiring stochastic $\pi^*$
    \item The corridor $[\kappa_l, \kappa_r]$ is calibrated to include typical optimal entropy levels
    \item Multi-turn tasks where exploration remains valuable throughout the dialogue
\end{enumerate}
In sparse-reward multi-turn settings, deterministic policies often fail to discover good trajectories, so optimal policies naturally maintain moderate entropy.
\end{remark}

\begin{remark}[Corridor Width Trade-off]\label{rem:corridor_width}
The corridor bounds $[\kappa_l, \kappa_r]$ control a precision-stability trade-off:

\textbf{Narrow corridor} ($\kappa_r - \kappa_l$ small):
\begin{itemize}
    \item Strong entropy enforcement—deviations corrected quickly
    \item Higher stability cost if $\pi^*$ lies outside corridor
    \item Risk: over-constraining may prevent learning optimal behavior
\end{itemize}

\textbf{Wide corridor} ($\kappa_r - \kappa_l$ large):
\begin{itemize}
    \item Weak entropy enforcement—allows more variation
    \item Lower stability cost—most policies satisfy constraint
    \item Risk: may not prevent cascade failure if corridor is too permissive
\end{itemize}

\textbf{Practical guidance:} Set $[\kappa_l, \kappa_r]$ to contain the entropy range of good policies while excluding collapse/explosion regimes. Based on experiments across multiple tasks (\autoref{sec:abl}), we find $\kappa_l = 0.5, \kappa_r = 1.5$ (allowing entropy to vary by $\pm 50\%$ from reference) provides a robust default.

\textbf{Sensitivity:} Let $w = \kappa_r - \kappa_l$ denote corridor width. The stability cost scales as:
$$\Delta^{\text{smooth}}_{\pi^*} \propto \max\left(0, \frac{|\mathcal{H}(\pi^*) - \bar{\mathcal{H}}^{ref}|}{\bar{\mathcal{H}}^{ref}} - \frac{w}{2}\right)$$
Widening the corridor by $\delta$ reduces stability cost by $O(\delta)$ when $\pi^*$ is near the boundary.
\end{remark}

\subsection{Comparison with Standard Entropy Regularization}

\begin{corollary}[EPO vs Standard Entropy Regularization]\label{cor:comparison}
The standard entropy-regularized bound~\citep{shen2025entropy}:
$$V^{\pi^*} - V^{\pi_\theta} \leq \frac{|\mathcal{D}|^2\epsilon^2}{2\lambda C_\lambda^{\pi_\theta}} + \lambda H\log\frac{|\mathcal{A}|}{|\mathcal{A}_H^*|^{1/H}}$$

EPO's bound differs by:
\begin{enumerate}
    \item A factor of $(1-\alpha\beta_k)^{-1} \in [1.11, 1.25]$ in optimization error (for $\alpha = 0.1$)
    \item The stability cost $\lambda\beta_k\Delta^{\text{smooth}}_{\pi^*}$ (zero under \autoref{prop:zero_cost})
\end{enumerate}
\end{corollary}

\begin{remark}[Quantifying Multi-Turn Advantage]\label{rem:quantify_multiturn}
For the multi-turn stability advantage (\autoref{prop:multiturn}) to outweigh EPO's per-step overhead, we require:
$$\frac{O(T^2)}{O(T)} = O(T) > \frac{1}{1-\alpha\beta_k}$$
With $\alpha = 0.1$ and $\beta_k \in [1, 2]$, the right-hand side is at most $1.25$. Thus for $T \geq 2$, the multi-turn stability advantage dominates the per-step overhead. For typical dialogue lengths ($T \geq 5$), the advantage is substantial: $5\times$ to $10\times$ reduction in cumulative entropy deviation.
\end{remark}

\section{Experiments}~\label{app:exper}

\subsection{Detailed Experiment Setup}
\subsubsection{Benchmark} 

We evaluate our method on two challenging and complementary benchmarks, ScienceWorld and ALFWorld, which test distinct yet crucial reasoning capabilities.

ScienceWorld~\citep{sciworld} is a dynamic, text-based environment simulating a grade-school science lab, where the agent must solve open-ended scientific tasks. Success demands systematic hypothesis testing, common-sense reasoning about object properties, and a deep understanding of cause and effect. Its curriculum is divided into over 30 task types sourced from official study guides, primarily spanning: Physics, with tasks such as powering electrical circuits, testing the conductivity of materials, and manipulating states of matter; Chemistry, including identifying acids and bases or observing chemical reactions; and Life Science, which involves classifying organisms based on their traits. These tasks rigorously test an agent's abstract knowledge and procedural reasoning.

In contrast to the abstract challenges in ScienceWorld, ALFWorld~\citep{alfworld} tests embodied reasoning in a visually-rich environment. It requires an agent to interpret high-level natural language instructions and decompose them into long sequences of low-level actions within simulated household settings. The benchmark is structured around seven main task categories designed to test long-horizon planning and language grounding: (1) Pick \& Place, for simple object relocation (e.g., "Put a mug in the coffee maker"); (2) Pick Two \& Place, for handling multiple objects; (3) Pick \& Cool/Heat, requiring state changes using appliances; (4) Pick \& Clean, involving interaction with sinks; and more complex compositional tasks like (5) Look At Obj \& Pick and (6) Examine In Light. Success in ALFWorld hinges on multi-step task planning, spatial awareness, and the ability to ground language in a physical context.

Together, these two benchmarks provide a comprehensive testbed for our agent's capabilities, spanning from abstract knowledge application in ScienceWorld to embodied task execution in ALFWorld.

\subsubsection{Evaluation Setting} 
We employ dual success rate metrics to capture different aspects of performance: \textbf{Succ.$^{*}$} reports the average of maximum success rates across random seeds, while $\overline{\textbf{Succ.}}$ measures average performance after convergence, reflecting practical reliability. To calculate the average converged success rate ($\overline{Succ.}$), we first identify a convergence period where performance stabilizes. 

In the \alfworld environment, we observed that all methods exhibit similar convergence trends, with success rates plateauing after step 125. Therefore, we compute $\overline{Succ.}$ by averaging the success rates from step 125 onward (inclusive) across three random seeds. 
In contrast, the \sciworld environment exhibited more varied convergence behaviors across different random seeds, necessitating a per-run analysis. Specifically, the epoch ranges for computing the converged success rate in the \sciworld environment were determined as follows:
In our comparison with GRPO, the evaluation windows for the three random seeds of the GRPO baseline were set to epochs 70-120, 90-120, and 90-120. In stark contrast, for our method (GRPO with EPO), these windows began significantly earlier, spanning epochs 60-80, 80-120, and 25-120, respectively.
A similar trend was observed in the PPO comparison. The PPO baseline's convergence was identified late in training, with windows of 100-120, 105-120, and 105-120. When enhanced with EPO, the model converged much faster, with its evaluation periods set to 70-120, 60-120, and 60-120 for the three seeds. This detailed breakdown confirms that EPO consistently accelerates convergence across different algorithms and seeds.

Given the high variance inherent in RL, final performance scores alone can be misleading. We therefore present averaged curves to provide a more robust comparison and illustrate the performance evolution throughout the training process. We apply wandb's default running average (window size of 10) to smooth all training curves. This standard practice avoids visualization-specific tuning and ensures a fair comparison of the underlying learning trends.  Additionally, we scale the variance by a factor of 0.8 for better visual clarity.

\subsubsection{Baselines} 
We conduct comprehensive comparisons across multiple paradigms to evaluate the effectiveness of our proposed EPO methodology. These baselines are grouped into four categories, each representing a distinct approach to training large language model (LLM) agents.

\textbf{Prompting-based Approaches}
This paradigm focuses on leveraging the in-context learning capabilities of LLMs without any parameter optimization.

The ReAct framework~\citep{react} synergizes reasoning and acting in language models. Its core innovation is the interleaving of textual reasoning traces with actions that interact with an external environment. Unlike prior methods that treated reasoning and acting as separate processes, ReAct allows the model to create and adjust high-level plans while grounding them in reality by gathering information from the environment. As a prompting-based method, ReAct's performance relies on the quality of the in-context examples and the inherent capabilities of the base model. Its operational scope is confined to single-pass inference, without mechanisms for parameter optimization or learning from experiences across multiple episodes to discover novel policies.

\textbf{Trajectory-based and Platform Methods}
This category includes methods that rely on imitating expert trajectories and platforms designed for agent development.

SFT is a fundamental approach for adapting pre-trained language models to specific tasks by imitating expert-provided trajectories. The effectiveness of SFT is contingent upon the availability of a comprehensive dataset of high-quality expert demonstrations. The resulting agent's policy is inherently bounded by the scope of behaviors observed within this dataset, constraining its ability to explore novel strategies beyond the demonstrated examples.

AgentGym~\citep{agentgym} is a comprehensive framework for building and evaluating generalist LLM-based agents, introducing a self-evolution method, AgentEvol. While AgentGym provides a valuable framework for evaluation, its AgentEvol method operates through a form of behavioral cloning. The evolution of the agent is thus guided by the quality and diversity of the initial trajectory data, which influences its sample efficiency in exploring the environment.

\textbf{General Reinforcement Learning Approaches}
This group consists of well-established reinforcement learning algorithms that are not specifically designed for LLM agents but are widely used in the field.

PPO~\citep{ppo} is a state-of-the-art on-policy reinforcement learning algorithm known for its stability and ease of implementation. It uses a clipped surrogate objective function to constrain policy updates. When applied to multi-turn, sparse-reward environments, credit assignment in standard PPO is performed based on the terminal reward signal. This structure can present challenges in distributing credit across a long sequence of actions, potentially leading to instabilities such as the ``entropy oscillation'' phenomenon.

GRPO~\citep{grpo} is a critic-free reinforcement learning algorithm. Instead of relying on a learned value function, it compares the performance of a group of trajectories generated from the same initial state. GRPO performs credit assignment at the trajectory level, evaluating the collective outcome of an entire episode. This design provides a holistic signal for policy updates, rather than turn-by-turn feedback, which is a consideration for learning complex, multi-step tasks.

\textbf{Agent RL Approaches}
This category includes recent reinforcement learning methods that are specifically designed for training LLM agents.

GIGPO~\citep{gigpo} extends GRPO by introducing a two-level hierarchical structure for advantage estimation. It refines the trajectory-level credit assignment of GRPO by introducing a micro-level analysis based on ``anchor states.'' The utility of this mechanism is related to the frequency with which these anchor states are revisited, and learning is guided by the single reward signal provided at the conclusion of each episode.

RLVMR~\citep{zhang2025rlvmr} is a framework that integrates dense, process-level supervision into the reinforcement learning loop by rewarding verifiable, meta-reasoning behaviors. RLVMR shapes the agent's behavior by leveraging a ``teacher'' model (e.g., GPT-4) to annotate expert trajectories with meta-reasoning tags. The learning process is thus guided by the quality and potential biases inherent in these teacher-provided annotations.

\subsubsection{Implementation Details}  
All of our experiments were conducted on a single server node equipped with eight NVIDIA H100 or A100 GPUs to ensure consistent and reproducible results.
The training times varied based on the environment's complexity, the RL algorithm, and the GPU architecture. On the computationally intensive SciWorld benchmark, a full PPO training run required approximately 23 hours on H100s and 31 hours on A100s. The more efficient GRPO baseline was faster, completing in 16 hours on H100s and 20 hours on A100s. For the ALFWorld environment, the PPO baseline took 23 hours with H100s and 31 hours with A100s, while the GRPO baseline required 18 hours and 25 hours, respectively. Crucially, our proposed EPO method is designed for efficiency and introduces negligible computational overhead. The training times for our EPO-enhanced variants (PPO+EPO and GRPO+EPO) are effectively identical to their corresponding PPO and GRPO baselines under the same hardware configuration. The detailed hyperparameter configurations for each setup are presented in \autoref{tab:hyperparameters_alfworld} and \autoref{tab:hyperparameters_sciworld}.

\begin{table}[h!]
\centering
\footnotesize
\caption{Hyperparameter settings for the \textbf{SciWorld} environment.}
\label{tab:hyperparameters_sciworld}
\begin{tabular}{@{}lcccc@{}}
\toprule
\textbf{Hyperparameter} & PPO & GRPO & PPO+EPO & GRPO+EPO \\
\midrule
\multicolumn{5}{c}{\textit{General Setup}} \\
\midrule
Foundation Model & \multicolumn{4}{c}{Qwen2.5-7B-Instruct} \\
Total RL Steps ($K$) & \multicolumn{4}{c}{125} \\
Max Prompt Length & \multicolumn{4}{c}{2048} \\
Max Response Length & \multicolumn{4}{c}{512} \\
Test Frequency (steps) & \multicolumn{4}{c}{5} \\
\midrule
\multicolumn{5}{c}{\textit{Optimizer \& LR Scheduler}} \\
\midrule
Optimizer & \multicolumn{4}{c}{AdamW} \\
LR Scheduler & \multicolumn{4}{c}{Cosine ($num\_cycles=0.5$)} \\
Learning Rate & $3 \times 10^{-6}$ & $5 \times 10^{-6}$ & $3 \times 10^{-6}$ & $5 \times 10^{-6}$ \\
Warmup / Min LR Ratio & \multicolumn{4}{c}{0.1 / 0.2} \\
\midrule
\multicolumn{5}{c}{\textit{Batch Sizes \& PPO-Specific Setup}} \\
\midrule
Mini-batch Size & 64 & 128 & 64 & 128 \\
Micro-batch Size & 8 & 16 & 8 & 16 \\
Critic Micro-batch Size & 4 & --- & 4 & --- \\
\midrule
\multicolumn{5}{c}{\textit{EPO}} \\
\midrule
$\lambda$ & --- & --- & \multicolumn{2}{c}{0.001} \\
% $\beta_{start}$ & --- & --- & \multicolumn{2}{c}{2.0} \\
% $\beta_{end}$ & --- & --- & \multicolumn{2}{c}{1.0} \\
$\kappa_l, \kappa_r$ & --- & --- & \multicolumn{2}{c}{0.0, 2.0} \\
$\gamma$ & --- & --- & \multicolumn{2}{c}{3.0} \\
$\lambda_k$ & --- & --- & \multicolumn{2}{c}{0.05} \\
\bottomrule
\end{tabular}
\end{table}

\begin{table}[h!]
\centering
\footnotesize
\caption{Hyperparameter settings for the \textbf{ALFWorld} environment.}
\label{tab:hyperparameters_alfworld}
\begin{tabular}{@{}lcccc@{}}
\toprule
\textbf{Hyperparameter} & PPO & GRPO & PPO+EPO & GRPO+EPO \\
\midrule
\multicolumn{5}{c}{\textit{General Setup}} \\
\midrule
Foundation Model & \multicolumn{4}{c}{Qwen2.5-3B-Instruct} \\
Total RL Steps ($K$) & \multicolumn{4}{c}{150} \\
Max Prompt Length & \multicolumn{4}{c}{2048} \\
Max Response Length & \multicolumn{4}{c}{512} \\
Test Frequency (steps) & \multicolumn{4}{c}{5} \\
\midrule
\multicolumn{5}{c}{\textit{Optimizer \& LR Scheduler}} \\
\midrule
Optimizer & \multicolumn{4}{c}{AdamW} \\
LR Scheduler & \multicolumn{4}{c}{Cosine ($num\_cycles=0.5$)} \\
Learning Rate & \multicolumn{4}{c}{$5 \times 10^{-6}$} \\
Warmup / Min LR Ratio & \multicolumn{4}{c}{0.1 / 0.2} \\
\midrule
\multicolumn{5}{c}{\textit{Batch Sizes \& PPO-Specific Setup}} \\
\midrule
Mini-batch Size & 256 & --- & 256 & --- \\
Micro-batch Size & 32 & --- & 32 & --- \\
Critic Micro-batch Size & 16 & --- & 16 & --- \\
\midrule
\multicolumn{5}{c}{\textit{EPO}} \\
\midrule
$\lambda$ & --- & --- & \multicolumn{2}{c}{0.001} \\
% $\beta_{start}$ & --- & --- & \multicolumn{2}{c}{2.0} \\
% $\beta_{end}$ & --- & --- & \multicolumn{2}{c}{1.0} \\
$\kappa_l, \kappa_r$ & --- & --- & \multicolumn{2}{c}{0.0, 2.0} \\
$\gamma$ & --- & --- & \multicolumn{2}{c}{3.0} \\
$\lambda_k$ & --- & --- & \multicolumn{2}{c}{0.1} \\
\bottomrule
\end{tabular}
\end{table}

\subsection{Performance Comparison}
\begin{figure*}[tb!]
\centering
% 第一行的两个子图
\subfigure[\scriptsize Training Rewards (ScienceWorld)]{
\begin{minipage}[t]{0.49\linewidth}
\centering
\includegraphics[width=0.99\linewidth]{figures/exper/grpo_alfworld_GRPO_vs_+EPO_episode_reward_mean.png}
%\caption{fig1}
\end{minipage}%
}%
\hfill % 填充所有可用的水平空间
\subfigure[\scriptsize Training Rewards (ALFWorld)]{
\begin{minipage}[t]{0.49\linewidth}
\centering
\includegraphics[width=0.99\linewidth]{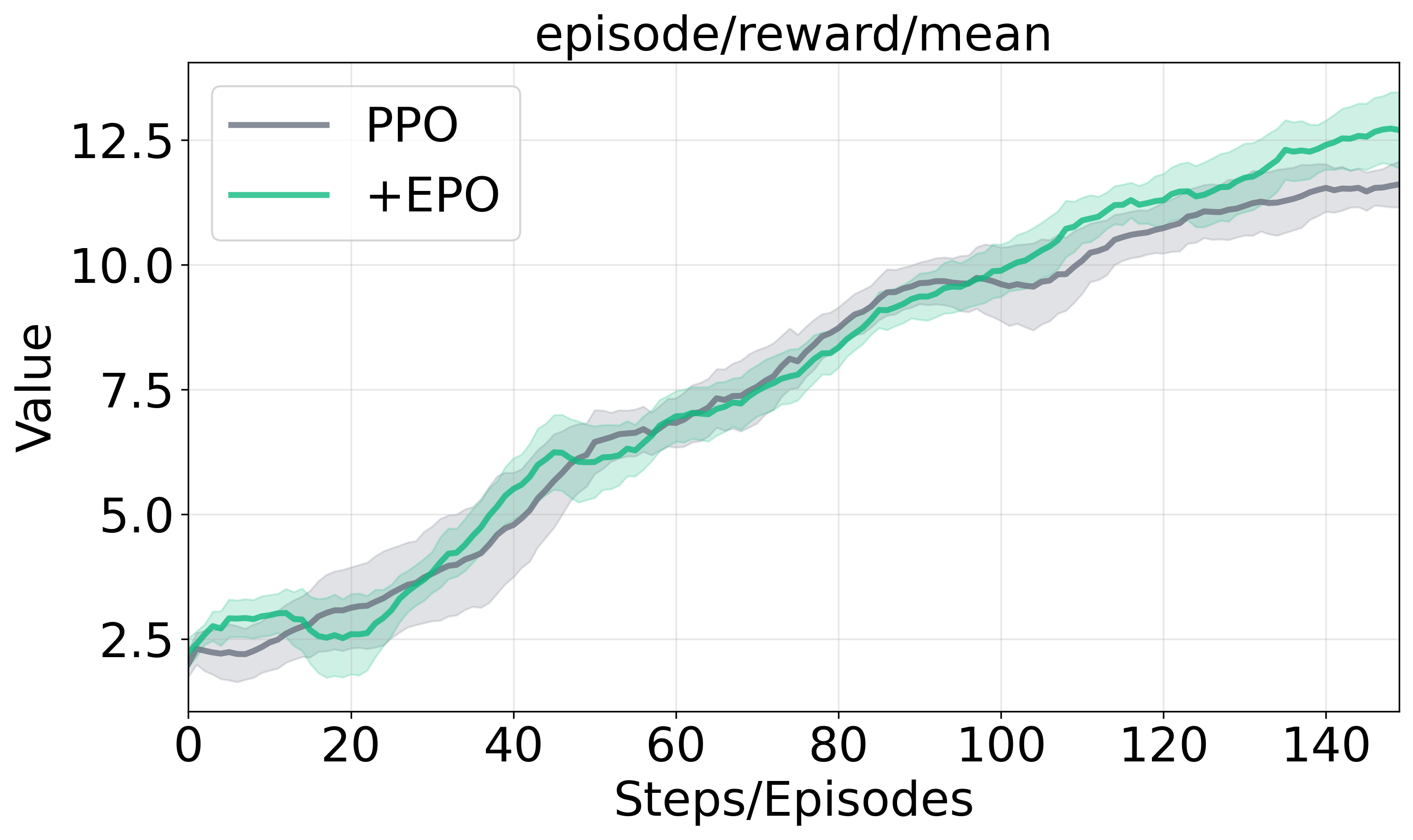}
%\caption{fig2}
\end{minipage}%
}% 
\\
\subfigure[\scriptsize IID Success Rate (ScienceWorld)]{
\begin{minipage}[t]{0.49\linewidth}
\centering
\includegraphics[width=0.99\linewidth]{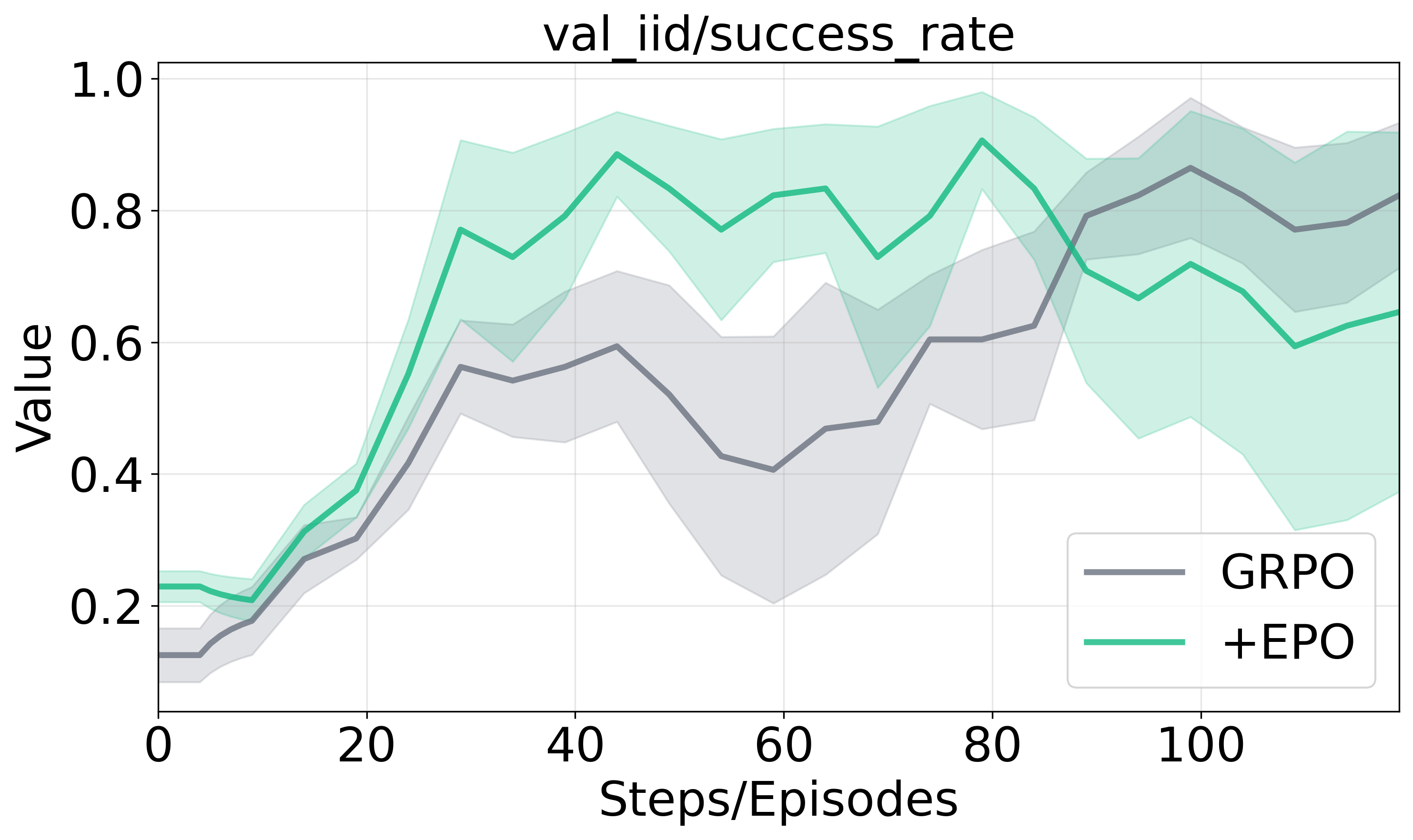}
%\caption{fig2}
\end{minipage}%
}% 
\hfill
\subfigure[\scriptsize IID Success Rate (ALFWorld)]{
\begin{minipage}[t]{0.49\linewidth}
\centering
\includegraphics[width=0.99\linewidth]{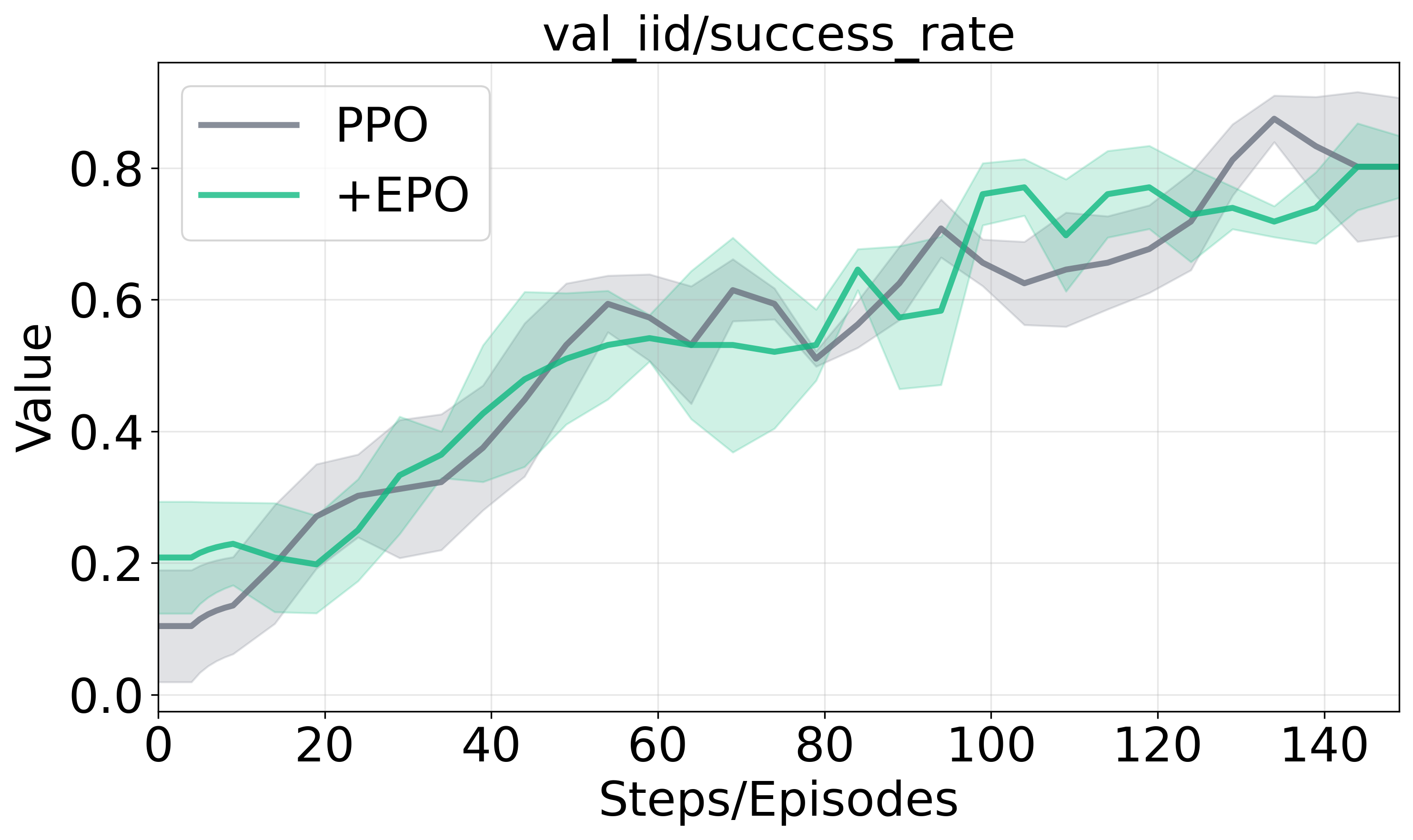}
%\caption{fig1}
\end{minipage}%
}%
\\
\subfigure[\scriptsize OOD Success Rate (ScienceWorld)]{
\begin{minipage}[t]{0.49\linewidth}
\centering
\includegraphics[width=0.99\linewidth]{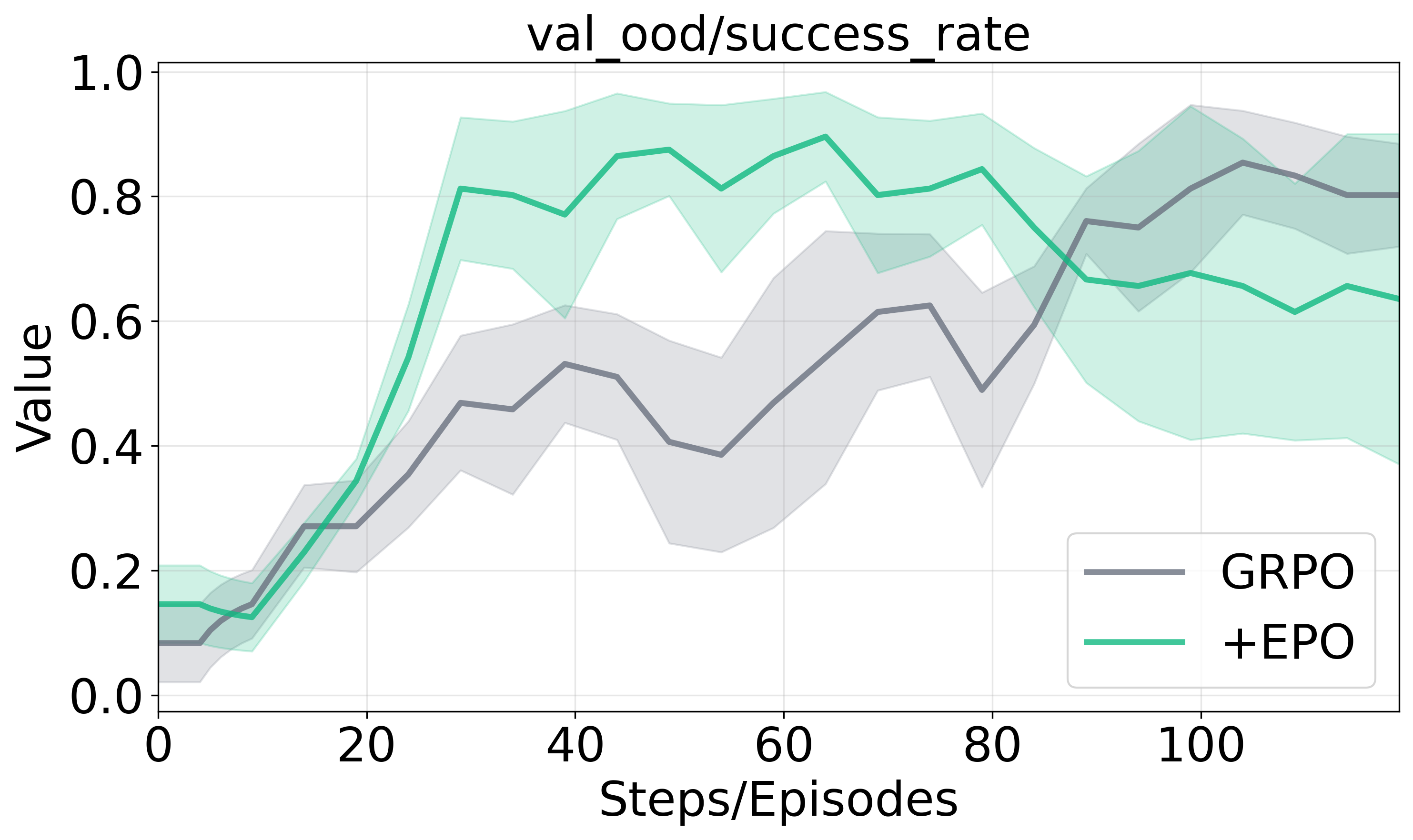}
%\caption{fig2}
\end{minipage}%
}% 
\hfill % 填充所有可用的水平空间
\subfigure[\scriptsize OOD Success Rate (ALFWorld)]{
\begin{minipage}[t]{0.49\linewidth}
\centering
\includegraphics[width=0.99\linewidth]{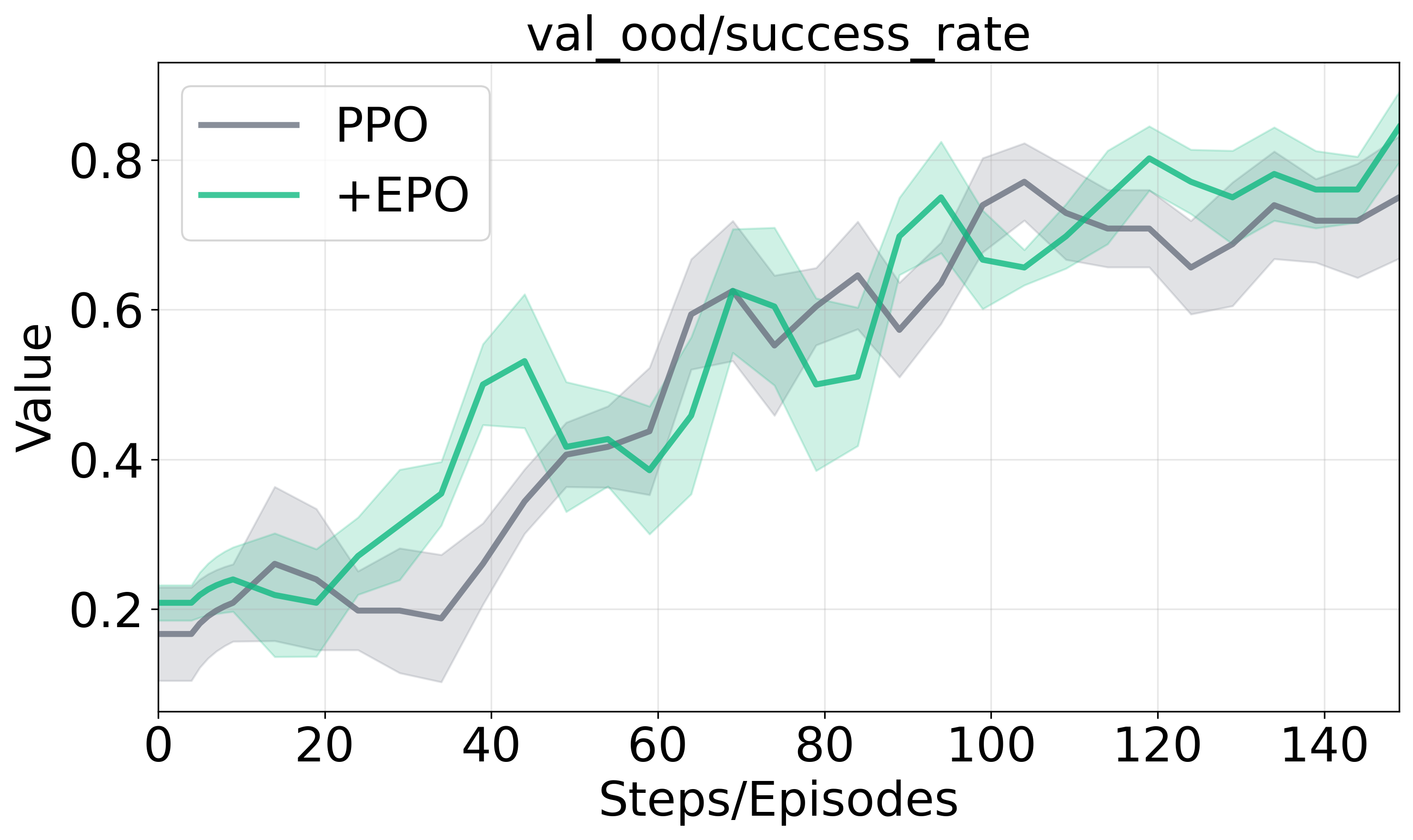}
%\caption{fig2}
\end{minipage}%
}% 

\centering
\vspace{-10pt}
% \captionsetup{font={small}}
\caption{Training dynamics and generalization performance analysis. We present the evolution of training rewards and validation success rates across both in-distribution (IID) and out-of-distribution (OOD) evaluation settings. (a,c,e) ScienceWorld experimental results contrasting GRPO and GRPO+EPO performance across training reward accumulation, IID validation, and OOD validation metrics. (b,d,f) ALFWorld  experimental results contrasting PPO and PPO+EPO under identical evaluation criteria. Our EPO enhancement exhibits significantly improved training stability and substantial performance gains across both IID and OOD evaluation scenarios against baseline methods.}
\label{app:fig:comparison}
\vspace{-10pt}
\end{figure*}

~\autoref{app:fig:comparison} demonstrates the training dynamics and generalization performance of our EPO enhancement across two environments and algorithms. Our EPO variants exhibit superior convergence characteristics in both settings. In ScienceWorld, GRPO+EPO achieves early convergence around step 60 with higher peak rewards than baseline GRPO (~\autoref{app:fig:comparison}(a)). Similarly, PPO+EPO in ALFWorld maintains more consistent reward accumulation with reduced oscillations (~\autoref{app:fig:comparison}(b)). This improved stability stems from EPO's entropy regularization guiding exploration toward productive policy regions.

For ScienceWorld, GRPO+EPO demonstrates clear advantages across both IID and OOD settings, achieving success rates exceeding 0.8 within 40 steps while baseline GRPO struggles to surpass 0.6 (~\autoref{app:fig:comparison}(c),(e)). In ALFWorld, PPO+EPO prioritizes OOD robustness over IID performance. While showing comparable IID results (~\autoref{app:fig:comparison}(d)), PPO+EPO maintains consistent OOD success rates above 0.6 compared to baseline PPO's frequent drops below 0.4 (~\autoref{app:fig:comparison}(f)).

The key advantage of EPO lies in variance reduction and elimination of training oscillations. Across both environments, EPO variants show tighter confidence intervals and more reliable convergence patterns. This stability particularly benefits OOD scenarios where baseline methods exhibit substantial performance degradation. These results validate our entropy regularization approach for addressing exploration-exploitation challenges in multi-turn LLM agent training, demonstrating simultaneous improvements in generalization capability and convergence reliability.

\subsection{Ablation Study}\label{app:abl study}
\begin{figure*}[tb!]
\centering
% 第一行的两个子图
\subfigure[\scriptsize Training Rewards (ScienceWorld)]{
\begin{minipage}[t]{0.32\linewidth}
\centering
\includegraphics[width=0.99\linewidth]{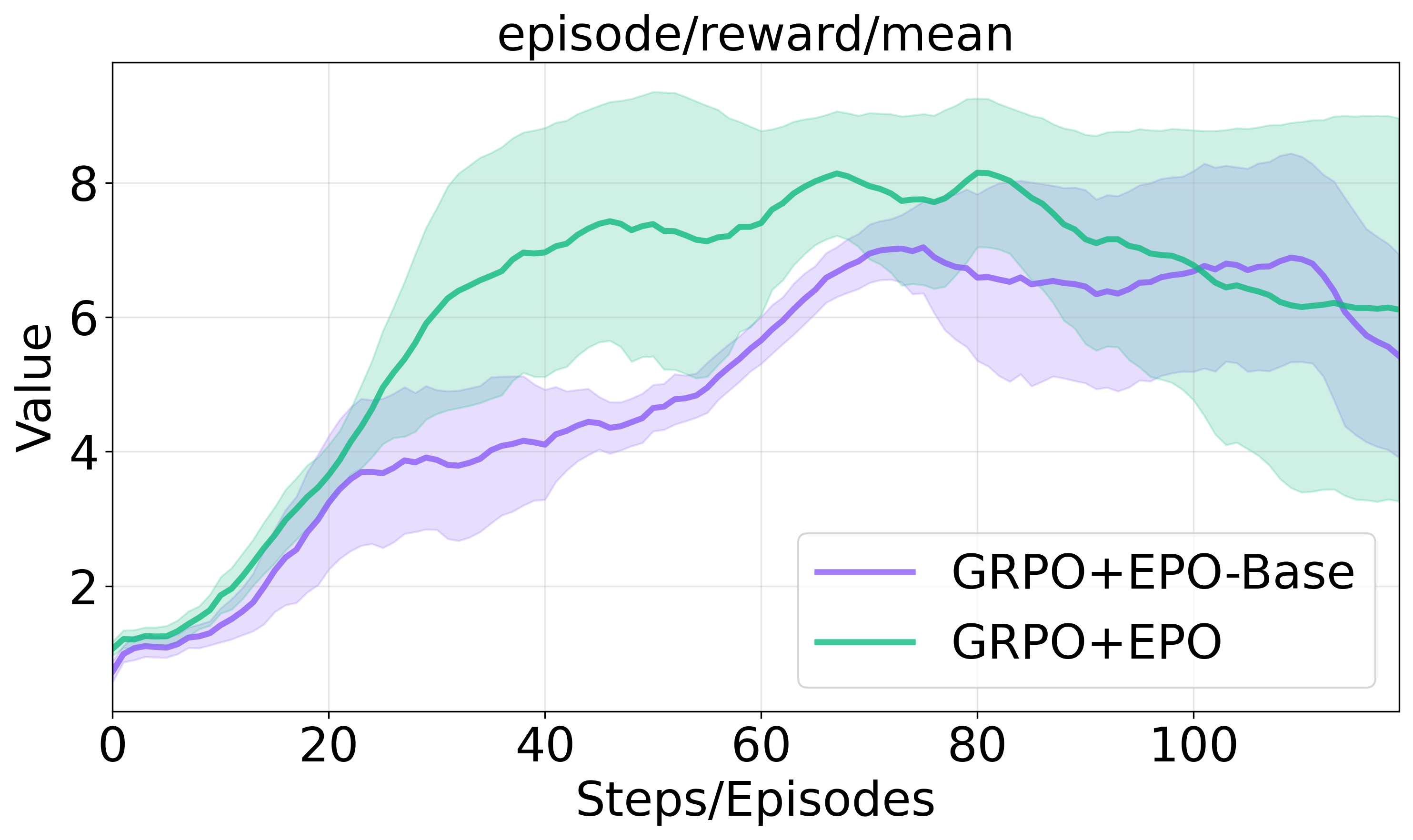}
%\caption{fig1}
\end{minipage}%
}%
\hfill % 填充所有可用的水平空间
\subfigure[\scriptsize Training Rewards (ALFWorld)]{
\begin{minipage}[t]{0.32\linewidth}
\centering
\includegraphics[width=0.99\linewidth]{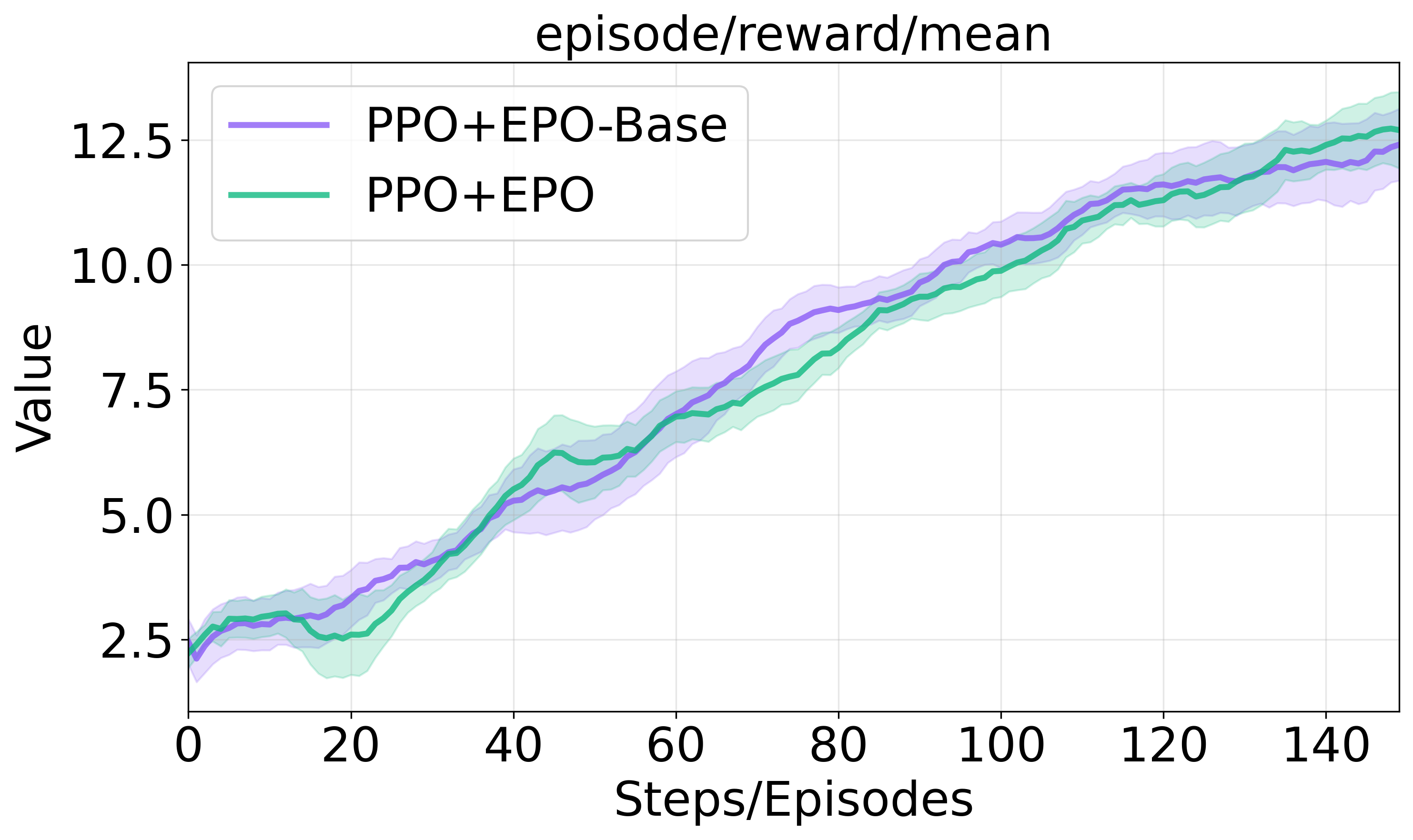}
%\caption{fig2}
\end{minipage}%
}% 
\hfill
\subfigure[\scriptsize IID Success Rate (ScienceWorld)]{
\begin{minipage}[t]{0.32\linewidth}
\centering
\includegraphics[width=0.99\linewidth]{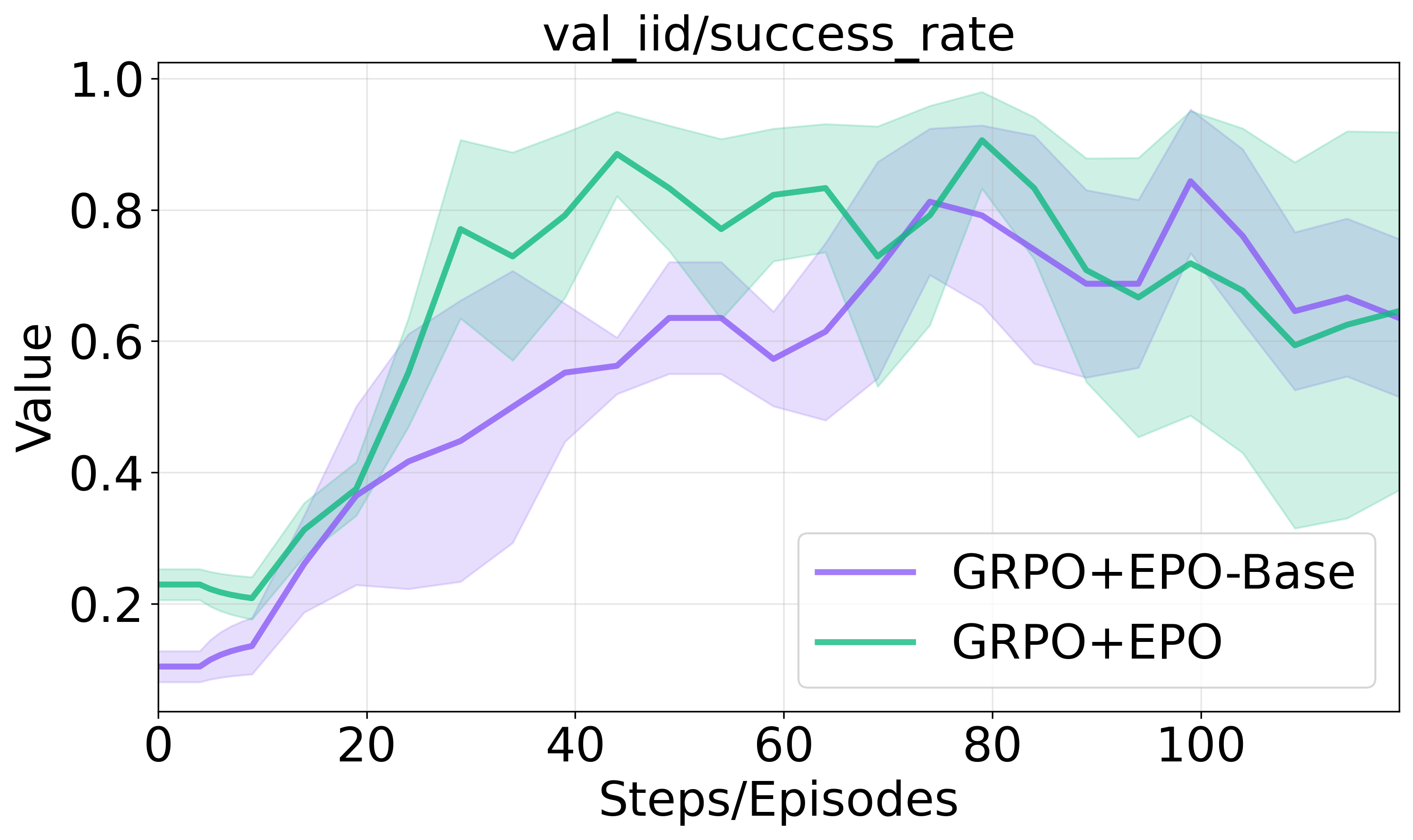}
%\caption{fig2}
\end{minipage}%
}% 
\\
\subfigure[\scriptsize IID Success Rate (ALFWorld) - PPO]{
\begin{minipage}[t]{0.32\linewidth}
\centering
\includegraphics[width=0.99\linewidth]{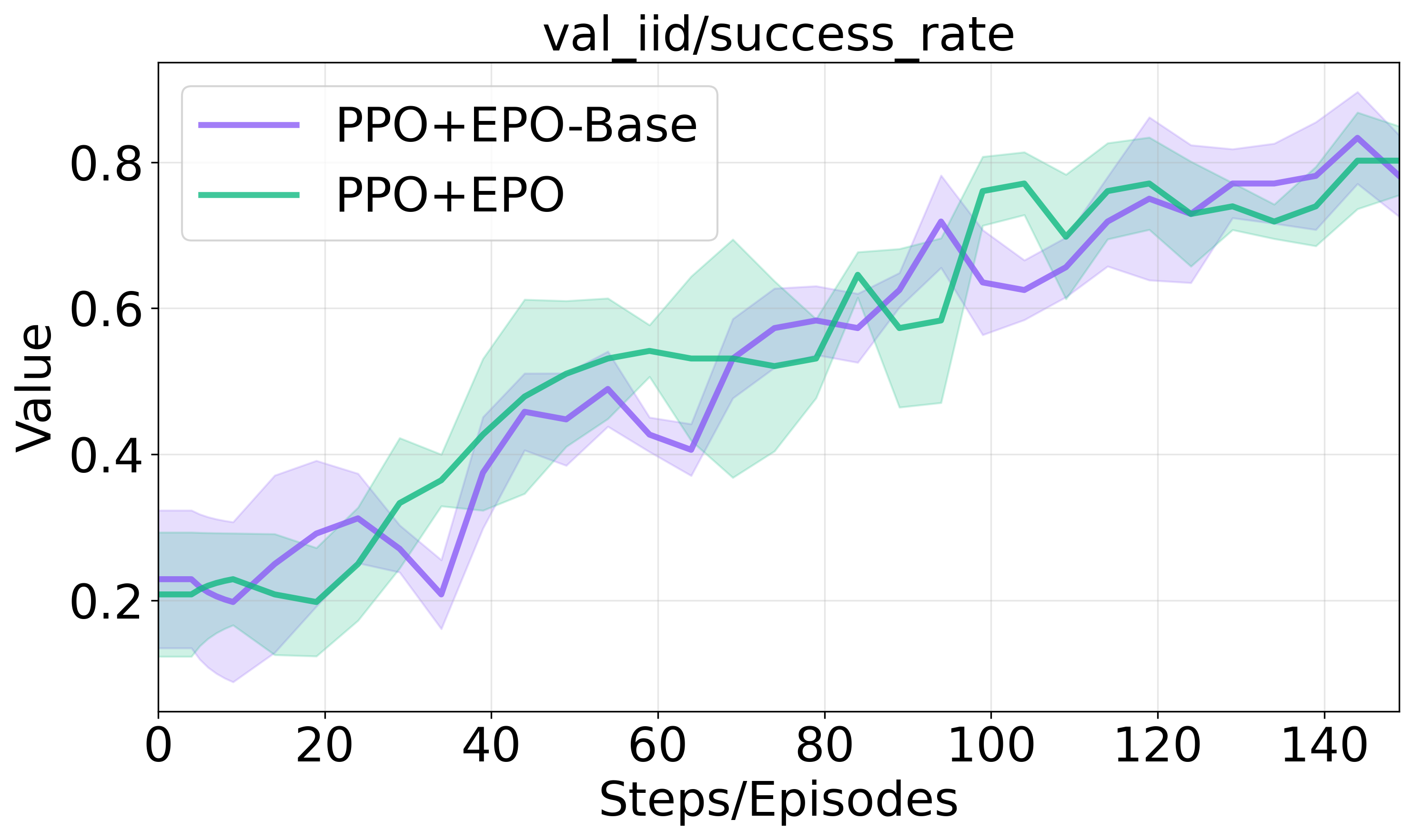}
%\caption{fig1}
\end{minipage}%
}%
\subfigure[\scriptsize OOD Success Rate (ScienceWorld) - PPO]{
\begin{minipage}[t]{0.32\linewidth}
\centering
\includegraphics[width=0.99\linewidth]{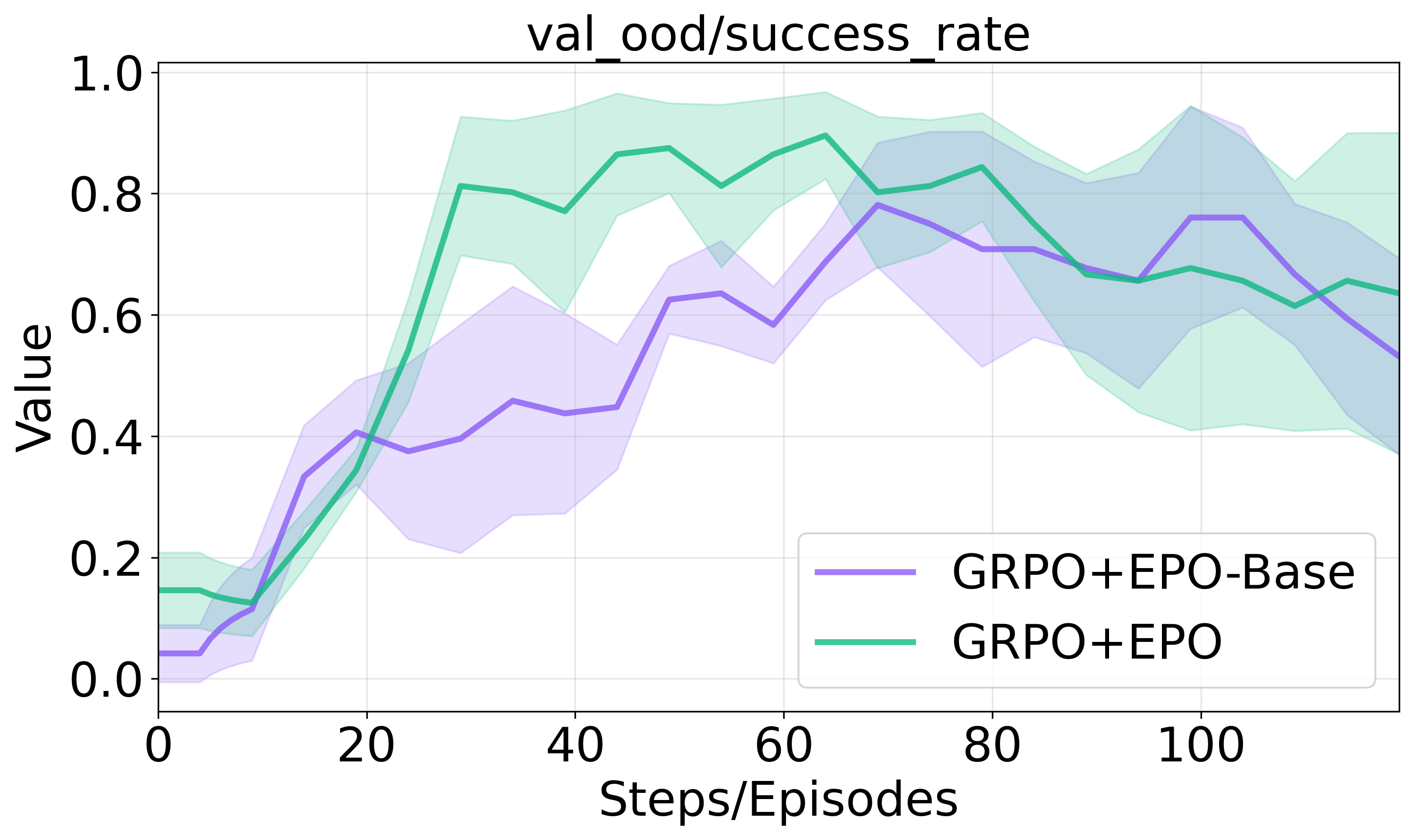}
%\caption{fig2}
\end{minipage}%
}% 
\hfill % 填充所有可用的水平空间
\subfigure[\scriptsize OOD Success Rate (ALFWorld) - PPO]{
\begin{minipage}[t]{0.32\linewidth}
\centering
\includegraphics[width=0.99\linewidth]{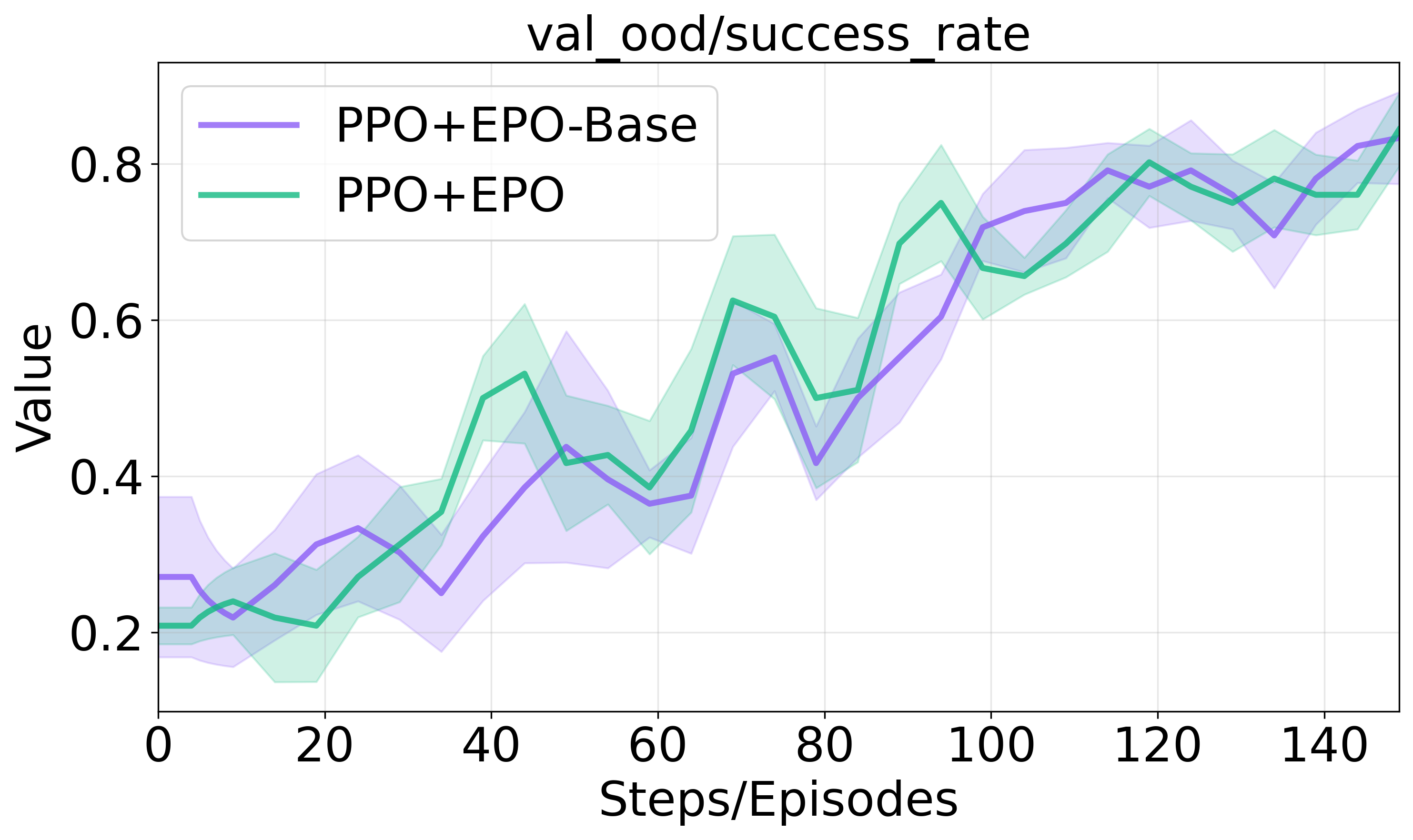}
%\caption{fig2}
\end{minipage}%
}% 
\\
\subfigure[\scriptsize Training Rewards (ALFWorld) - GRPO]{
\begin{minipage}[t]{0.32\linewidth}
\centering
\includegraphics[width=0.99\linewidth]{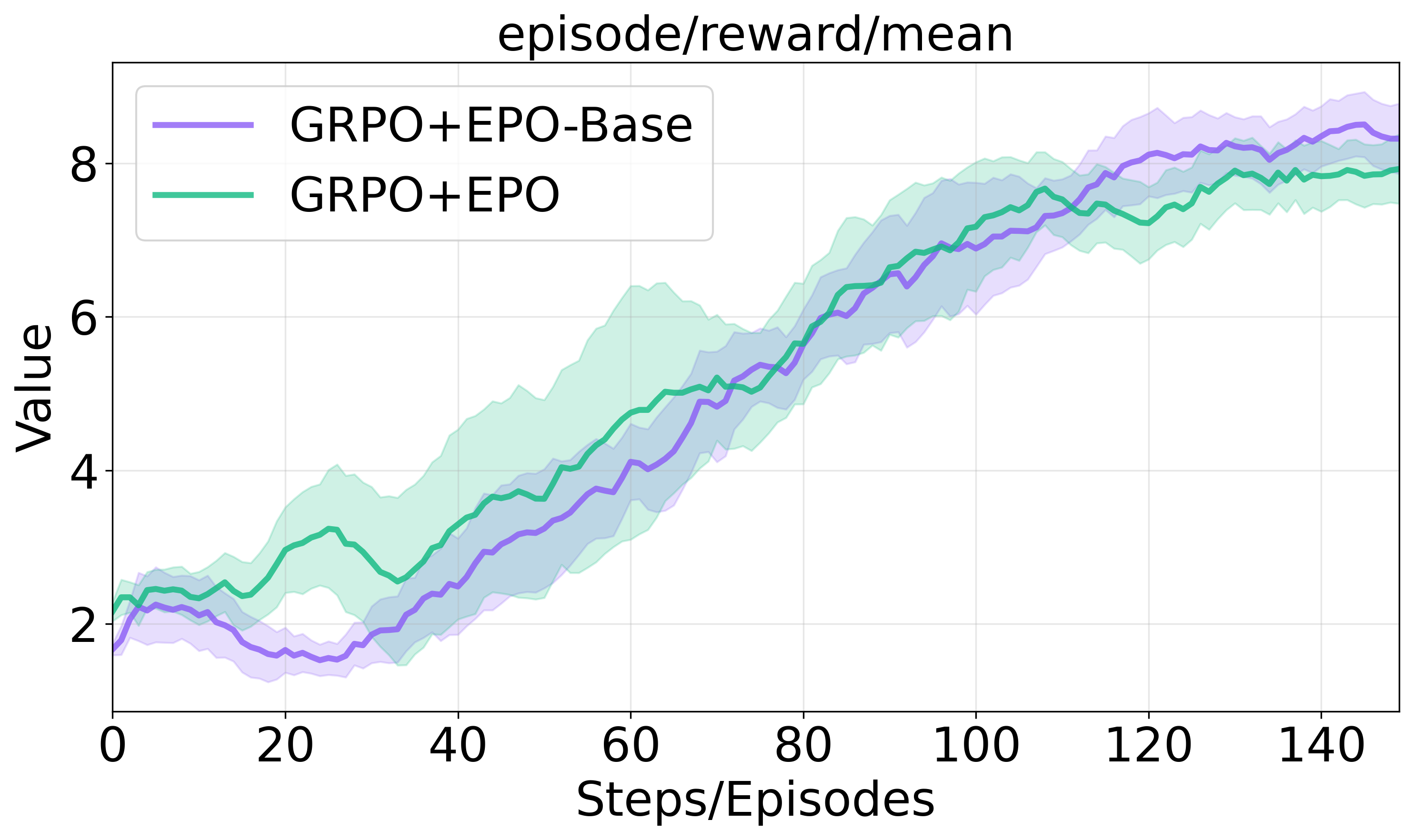}
%\caption{fig1}
\end{minipage}%
}%
\hfill 
\subfigure[\scriptsize IID Success Rate (ALFWorld) - GRPO]{
\begin{minipage}[t]{0.32\linewidth}
\centering
\includegraphics[width=0.99\linewidth]{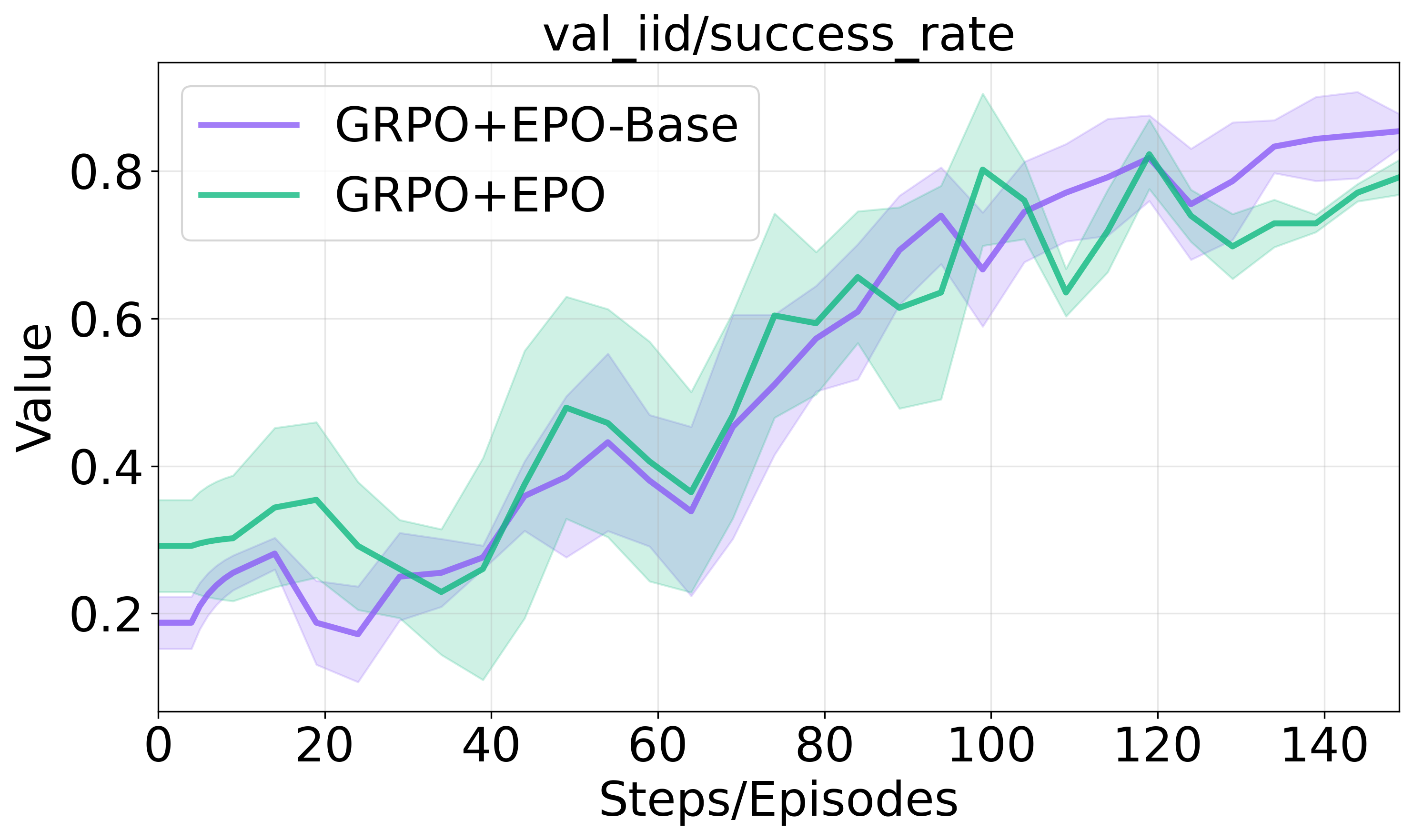}
%\caption{fig2}
\end{minipage}%
}% 
\hfill 
\subfigure[\scriptsize OOD Success Rate (ALFWorld) - GRPO]{
\begin{minipage}[t]{0.32\linewidth}
\centering
\includegraphics[width=0.99\linewidth]{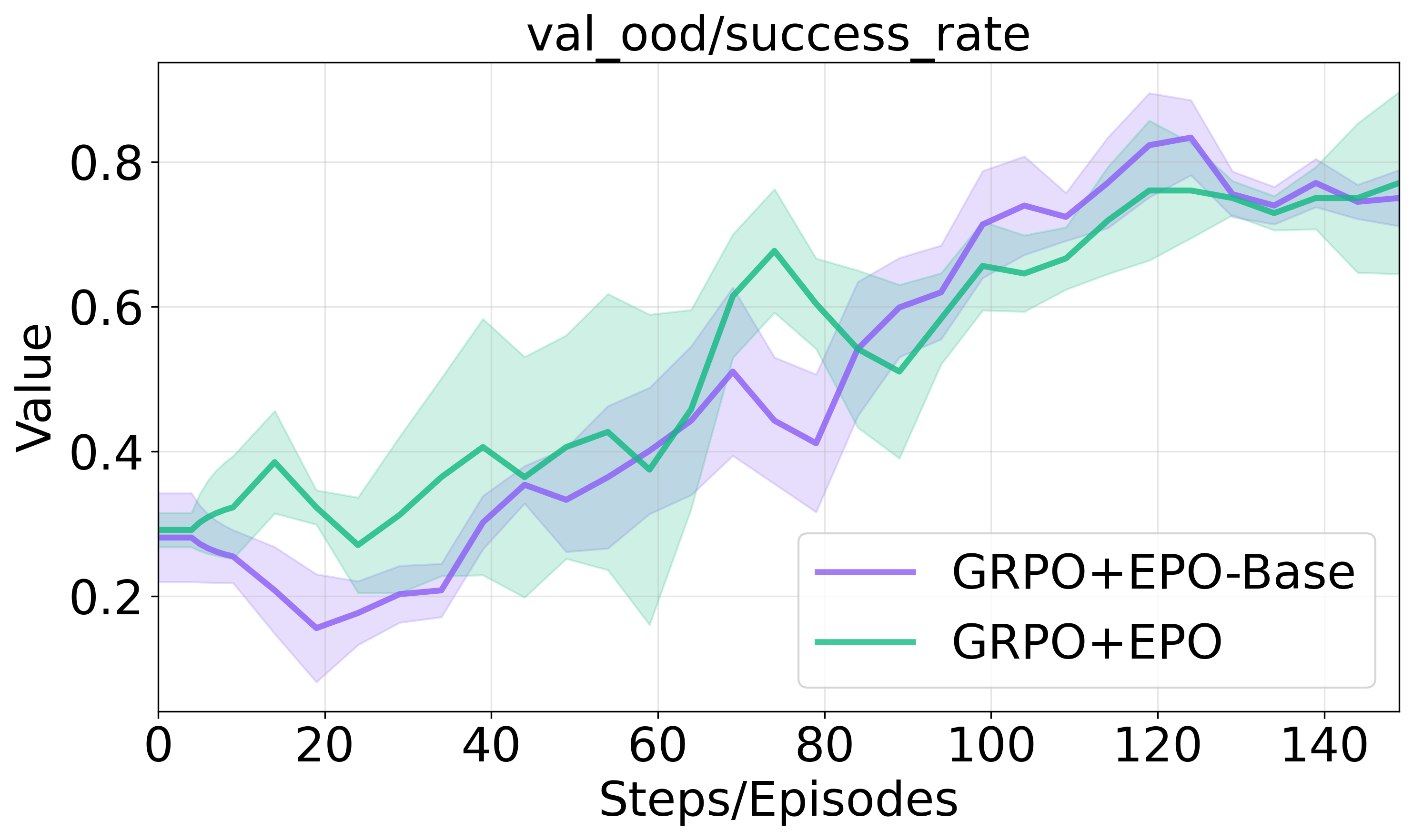}
%\caption{fig2}
\end{minipage}%
}% 
\\
\centering
\vspace{-10pt}
% \captionsetup{font={small}}
\caption{Impact of the entropy smoothing regularizer on training dynamics and performance. This ablation study contrasts our full method (EPO) with a variant that excludes the entropy smoothing regularizer (EPO-base). The comparison on ScienceWorld (a,c,e) and ALFWorld (b,d,f) demonstrates that the smoothing mechanism is essential for stable RL training progression. }
\label{app:fig:abl}
\vspace{-10pt}
\end{figure*}

\autoref{app:fig:abl} extends our ablation analysis to both GRPO and PPO variants across ScienceWorld and ALFWorld. In \textbf{ScienceWorld} (a,c,e), the entropy smoothing regularizer proves essential: GRPO+EPO-Base exhibits severely delayed learning with rewards remaining near 2 until step 40 and success rates plateauing at 0.6, while GRPO+EPO achieves immediate progress reaching 7-8 rewards and 0.7-0.85 success rates with a 40-50\% relative improvement that persists throughout training. This pattern holds across both GRPO and PPO, confirming the mechanism's algorithm-agnostic benefits. \textbf{ALFWorld} (b,d,f) shows markedly different dynamics: both PPO variants converge to similar final performance (~12.5 reward, 0.8 success rate), with PPO+EPO primarily demonstrating 20-episode faster convergence. This differential impact validates our theoretical framework—ScienceWorld's extreme sparsity (30+ actions before feedback) creates pathological exploration-exploitation oscillations that the smoothing regularizer effectively breaks by maintaining entropy within historical bounds. ALFWorld's structured feedback naturally prevents such oscillations, making smoothing beneficial for speed but not essential for convergence.

The consistent improvements across both GRPO and PPO in sparse settings confirm that entropy smoothing addresses a fundamental challenge in multi-turn optimization rather than algorithm-specific weaknesses. The adaptive $\beta_k$ weighting enables this by providing strong stabilization when oscillations are detected while relaxing constraints during natural convergence, transforming intractable sparse reward problems into smoothly converging optimization processes.

\subsection{Model Study}~\label{app:model study}

\subsubsection{Study of Entropy Regularization}
To analyze the exploration-exploitation trade-off, we compare our standard method, \textbf{PPO+EPO-Base}, which applies a consistent entropy regularization coefficient throughout training, against an experimental variant, \textbf{PPO+EPO-Decay}. This variant was designed to test the hypothesis that a dynamic schedule could improve performance. It employs a formula to modulate the entropy weight over time: assigning a higher weight during initial training phases to promote exploration, and systematically reducing the weight in later phases to encourage exploitation.

\begin{figure*}[tb!]
\centering
% 第一行的两个子图
\subfigure[\scriptsize Training Rewards (ScienceWorld)]{
\begin{minipage}[t]{0.49\linewidth}
\centering
\includegraphics[width=0.99\linewidth]{figures/exper/ppo_sciworld_PPO+EPO-Base_vs_PPO+EPO-Decay_episode_reward_mean.png}
%\caption{fig1}
\end{minipage}%
}%
\hfill % 填充所有可用的水平空间
\subfigure[\scriptsize IID Success Rate (ScienceWorld)]{
\begin{minipage}[t]{0.49\linewidth}
\centering
\includegraphics[width=0.99\linewidth]{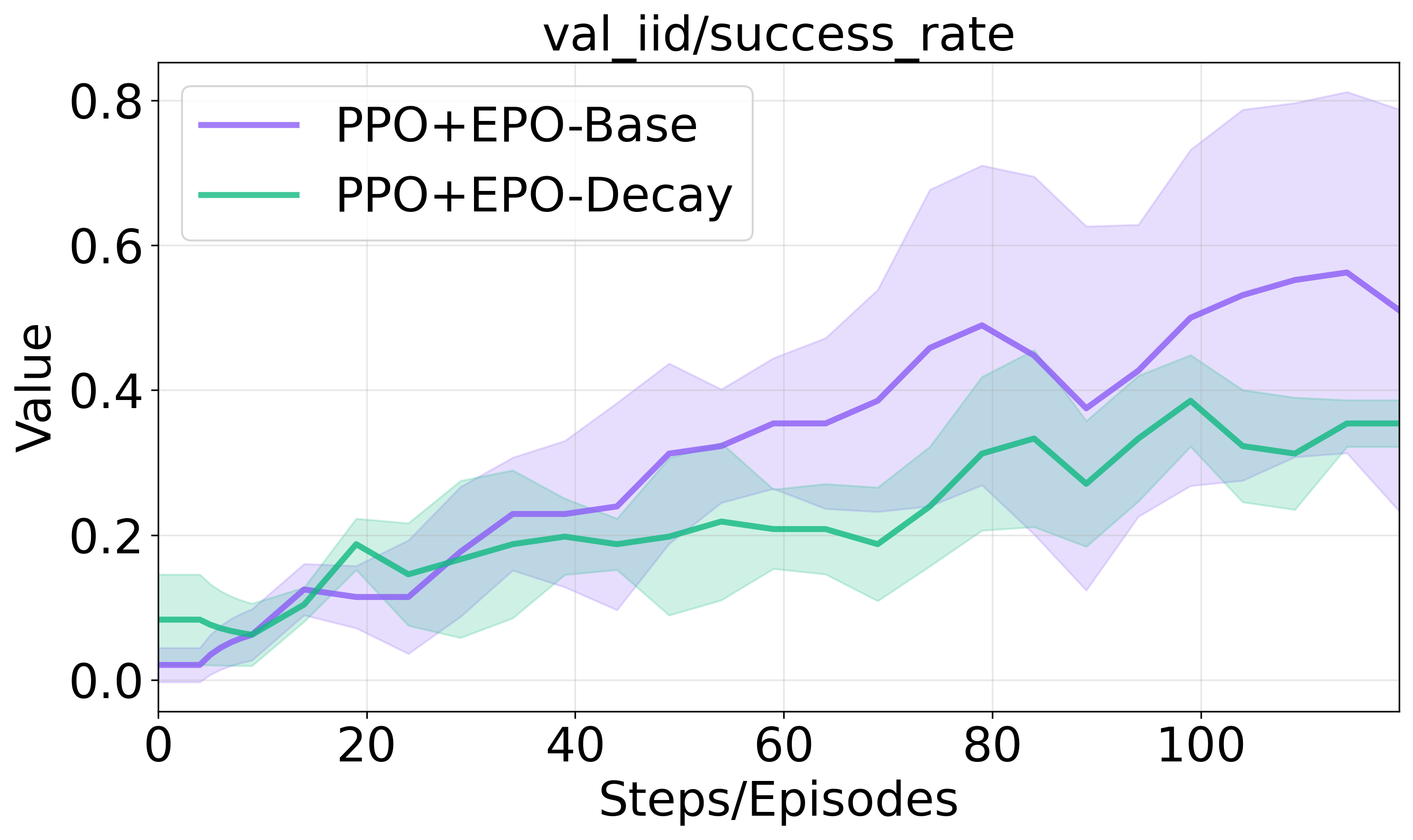}
%\caption{fig2}
\end{minipage}%
}% 
\\
\subfigure[\scriptsize OOD Success Rate (ScienceWorld)]{
\begin{minipage}[t]{0.49\linewidth}
\centering
\includegraphics[width=0.99\linewidth]{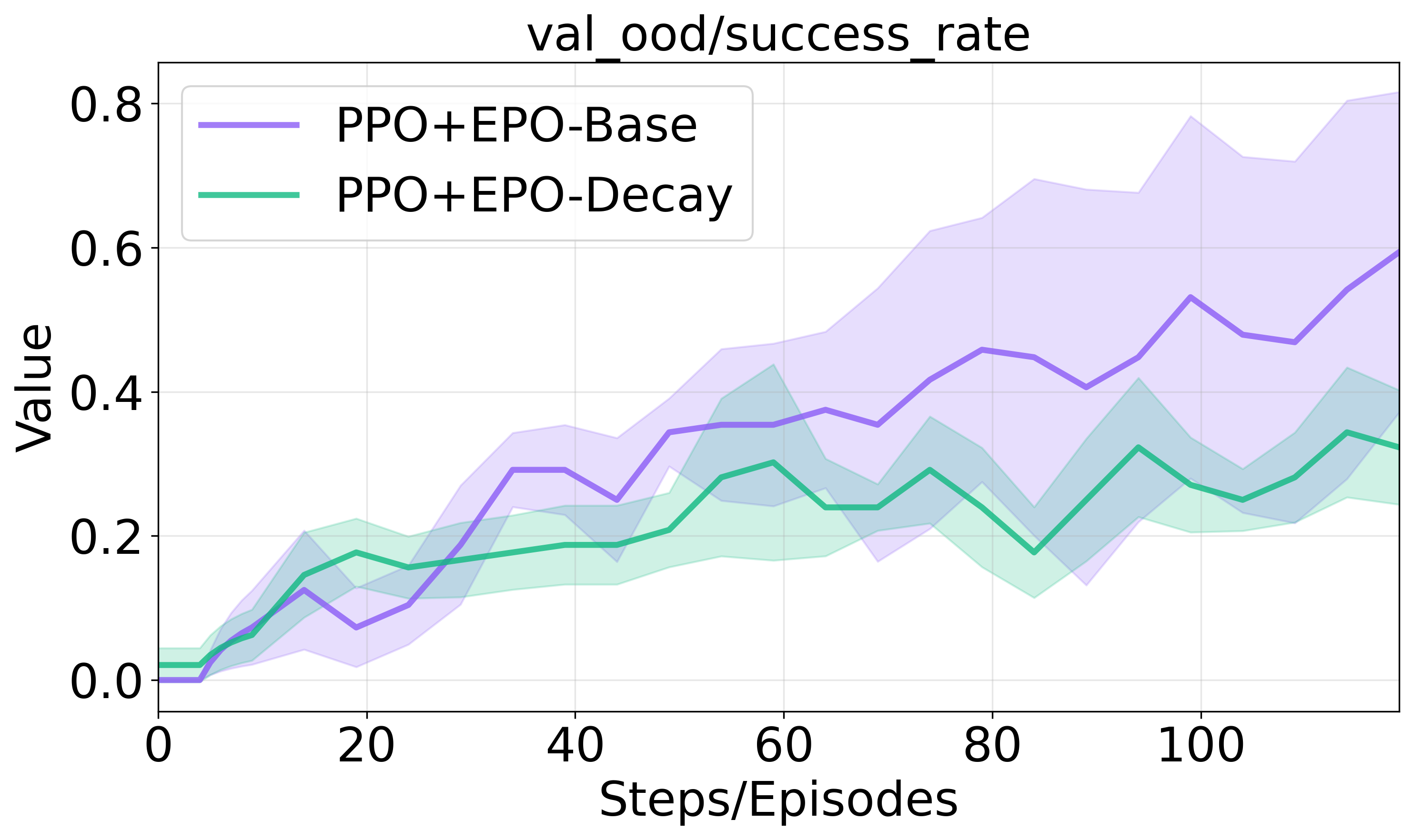}
%\caption{fig2}
\end{minipage}%
}% 
\hfill
\subfigure[\scriptsize Entropy Comparison (ScienceWorld)]{
\begin{minipage}[t]{0.49\linewidth}
\centering
\includegraphics[width=0.99\linewidth]{figures/exper/ppo_sciworld_PPO+EPO-Base_vs_PPO+EPO-Decay_entropy_group_comparison.png}
%\caption{fig1}
\end{minipage}%
}%
\centering
\vspace{-10pt}
% \captionsetup{font={small}}
\caption{Performance comparison of our standard PPO+EPO-Base against PPO+EPO-Decay, which uses a decaying entropy coefficient, on the ScienceWorld benchmark. Panels (a-c) demonstrate that the dynamic decay schedule consistently degrades performance across episodic rewards and success rates. Panel (d) analyzes the intra-episode entropy for early versus late tokens, revealing that the decay schedule prematurely suppresses crucial early-turn exploration, which negatively impacts overall performance.}
\label{app:fig:model_study:ER}
\vspace{-10pt}
\end{figure*}

Contrary to our hypothesis, the empirical results in ~\autoref{app:fig:model_study:ER} show this strategy is counterproductive. The PPO+EPO-Decay variant consistently underperforms the baseline across all metrics, including episodic reward (a), in-distribution success rate (b), and out-of-distribution success rate (c).

Panel (d) provides insight into this failure by analyzing the intra-episode entropy, comparing the average entropy of the first 10 tokens (``Early Steps'') with the last 10 tokens (``Late Steps''). While the decay schedule successfully reduces the policy's entropy in the later stages of training, it does so at a significant cost. The schedule prematurely suppresses exploration in the crucial initial turns of each episode. This insufficient early exploration locks the agent into suboptimal strategies from which it cannot recover, even as the policy becomes more deterministic. This finding underscores that for complex, multi-turn tasks, maintaining a robust and consistent exploration pressure is more effective than manually scheduling a transition towards exploitation.

\subsubsection{Study of Entropy-Shaped Advantage}
\begin{figure*}[tb!]
\centering
% 第一行的两个子图
\subfigure[ScienceWorld]{
\begin{minipage}[t]{0.49\linewidth}
\centering
\includegraphics[width=0.99\linewidth]{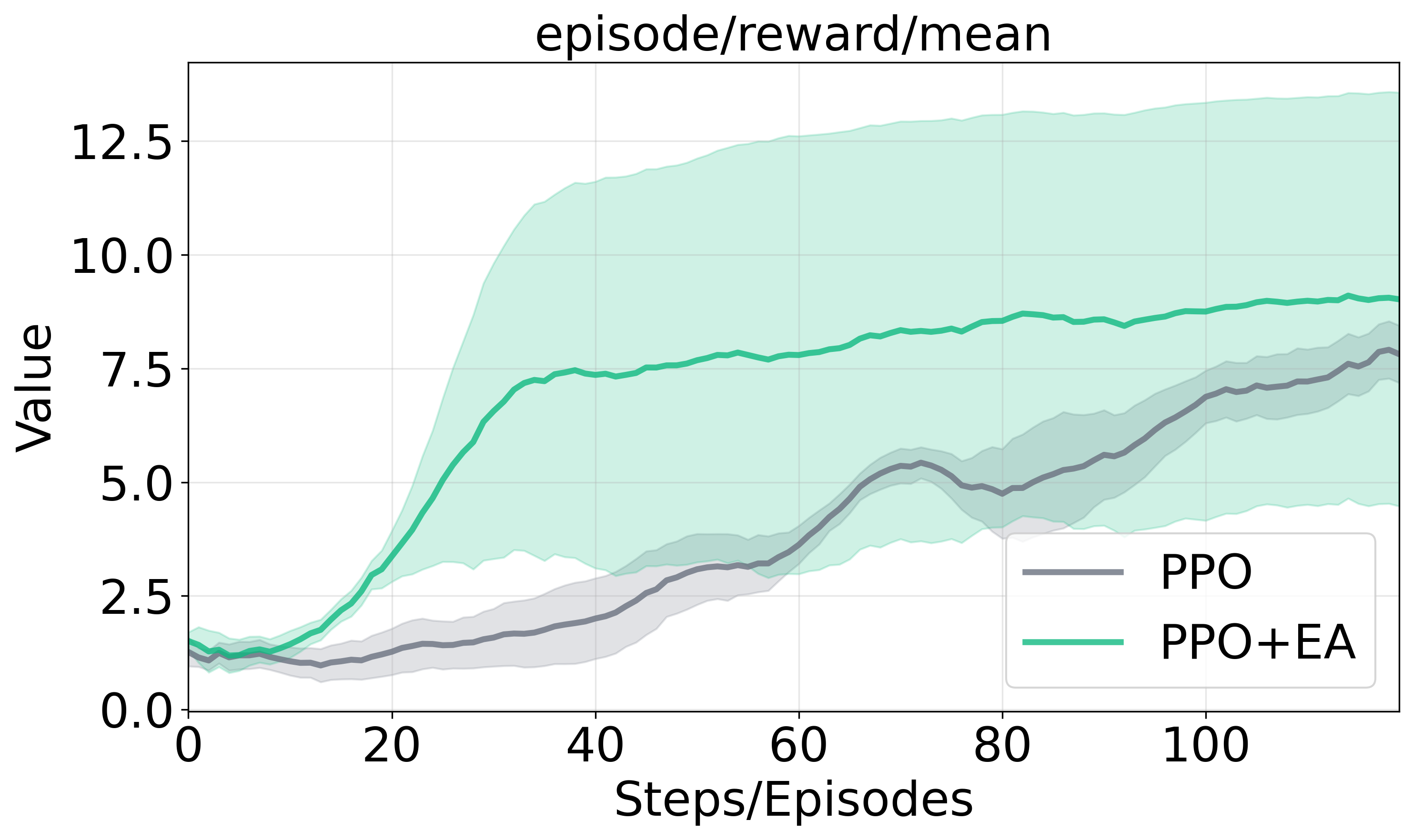}
%\caption{fig1}
\end{minipage}%
}%
\hfill % 填充所有可用的水平空间
\subfigure[ScienceWorld]{
\begin{minipage}[t]{0.49\linewidth}
\centering
\includegraphics[width=0.99\linewidth]{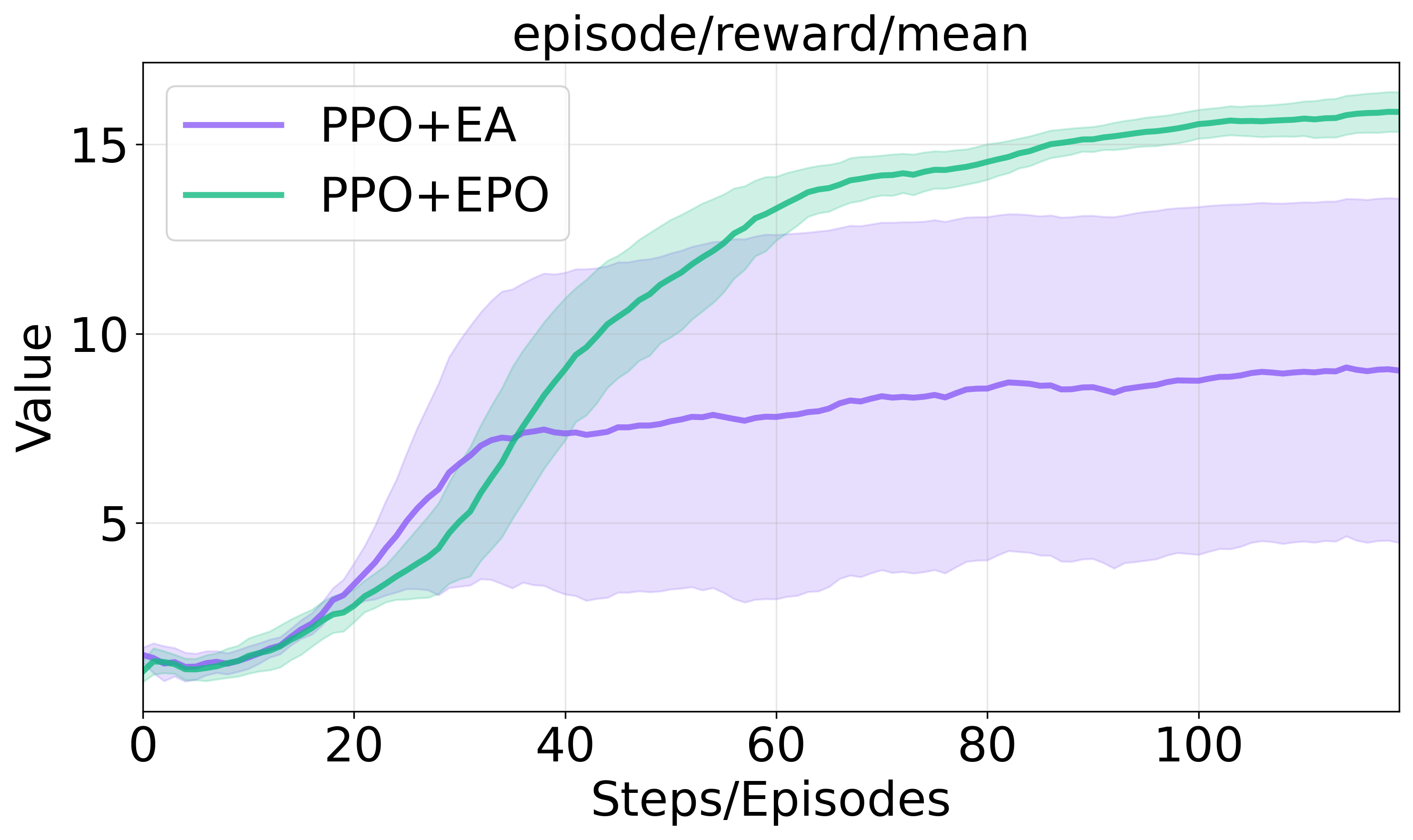}
%\caption{fig2}
\end{minipage}%
}% 
\\
\subfigure[ScienceWorld]{
\begin{minipage}[t]{0.49\linewidth}
\centering
\includegraphics[width=0.99\linewidth]{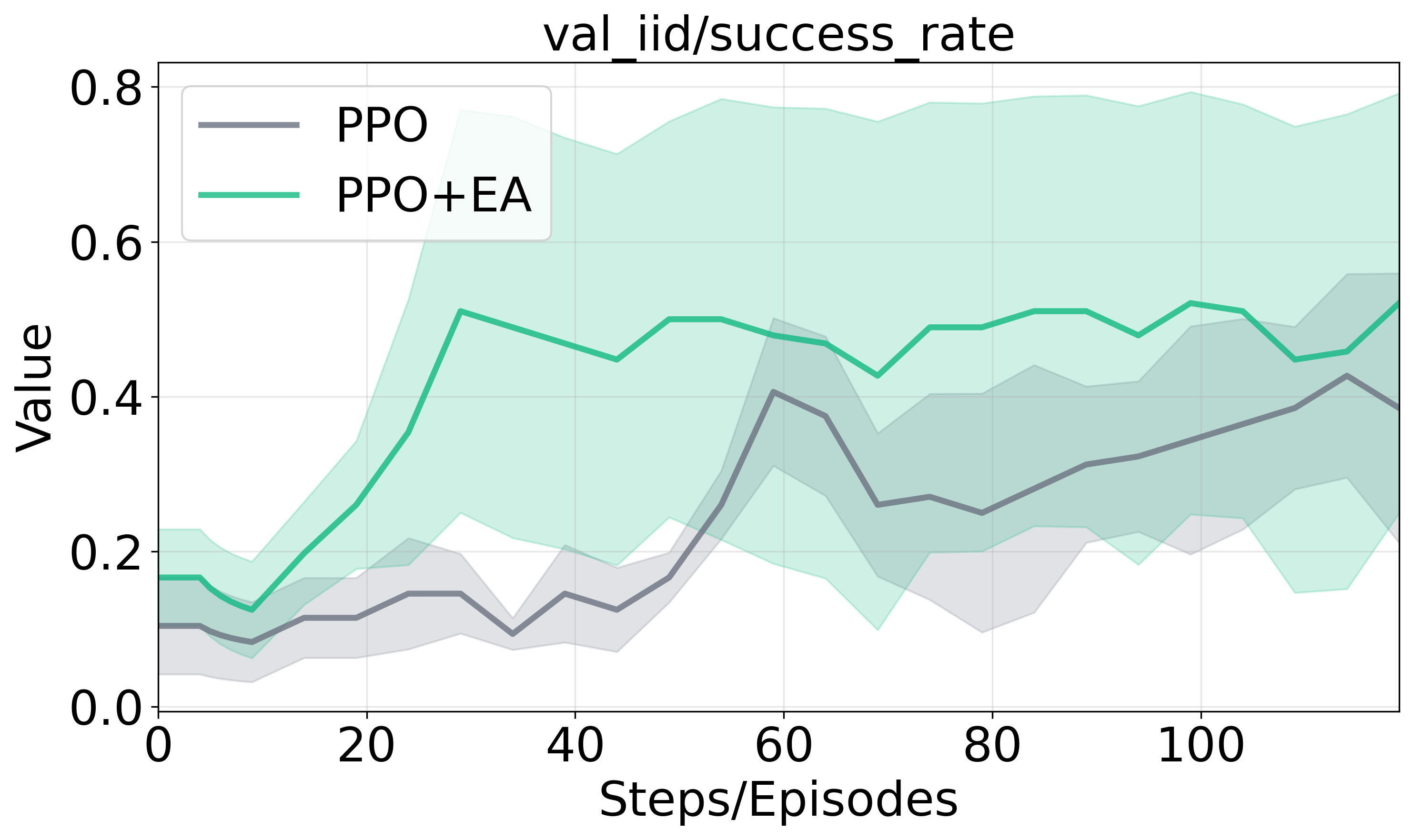}
%\caption{fig2}
\end{minipage}%
}% 
\hfill
\subfigure[ScienceWorld]{
\begin{minipage}[t]{0.49\linewidth}
\centering
\includegraphics[width=0.99\linewidth]{figures/exper/ppo_sciworld_PPO+EA_vs_PPO+EPO_val_l0_success_rate.png}
%\caption{fig1}
\end{minipage}%
}%
\\
\subfigure[ScienceWorld]{
\begin{minipage}[t]{0.49\linewidth}
\centering
\includegraphics[width=0.99\linewidth]{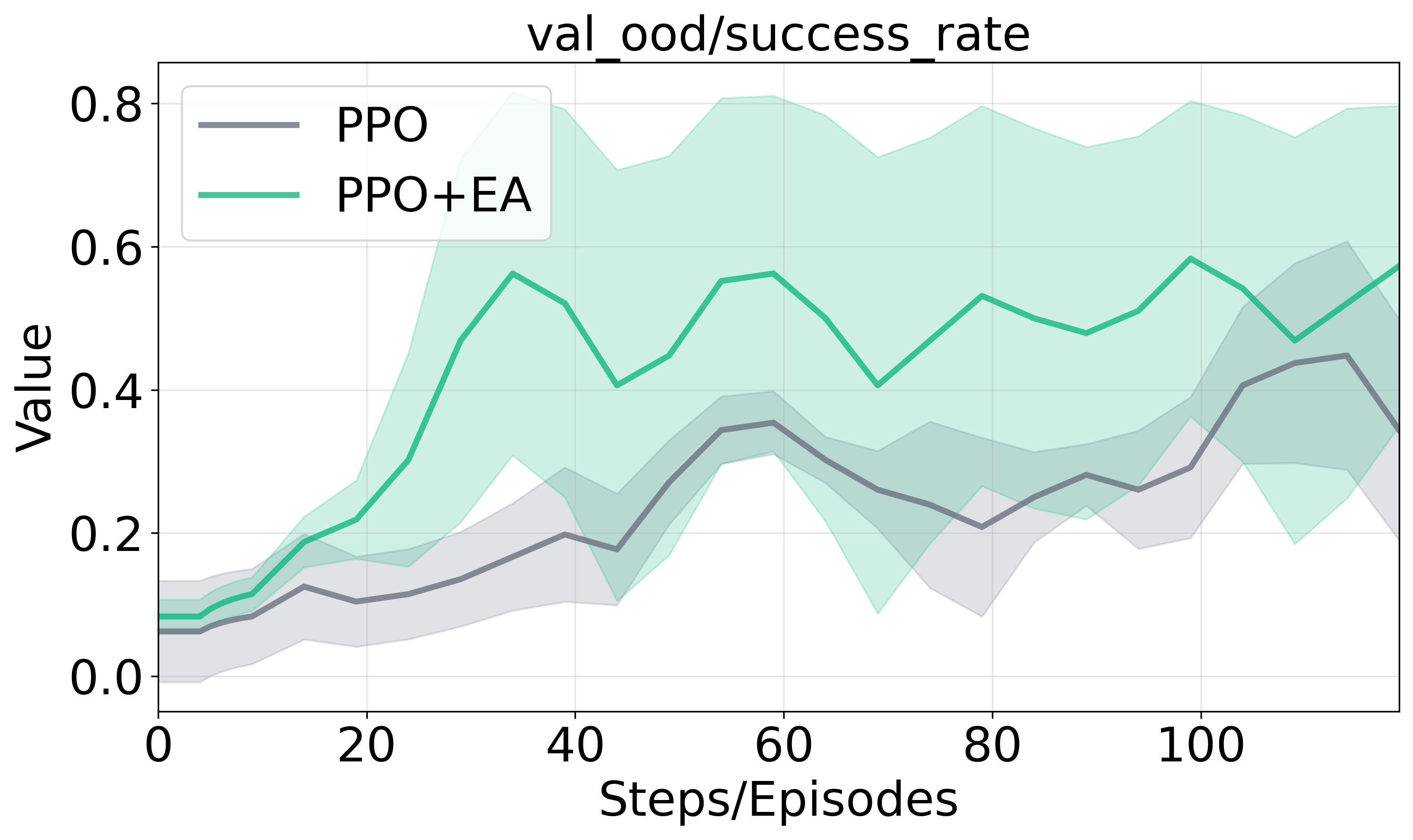}
%\caption{fig2}
\end{minipage}%
}% 
\hfill % 填充所有可用的水平空间
\subfigure[ScienceWorld]{
\begin{minipage}[t]{0.49\linewidth}
\centering
\includegraphics[width=0.99\linewidth]{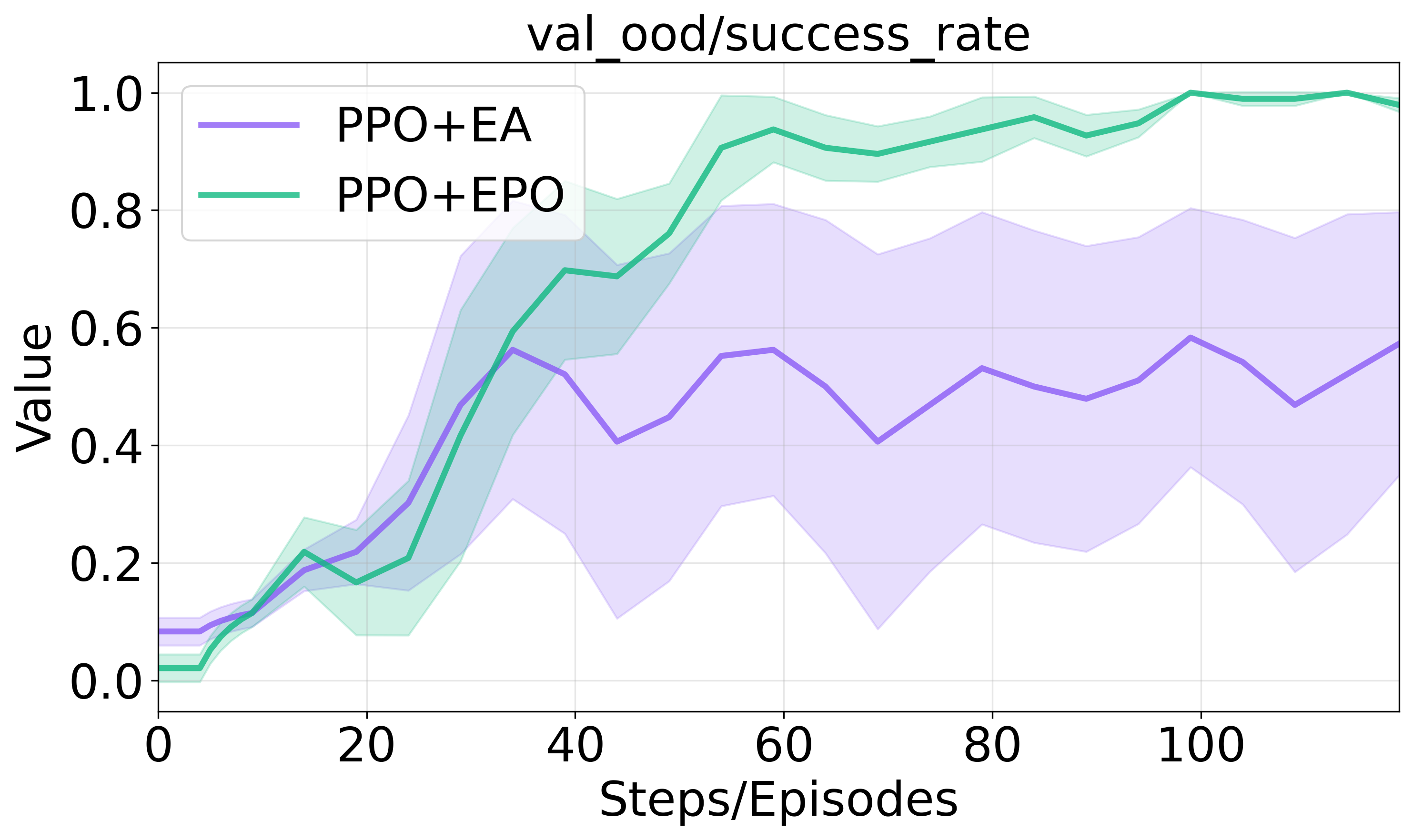}
%\caption{fig2}
\end{minipage}%
}% 

\centering
\vspace{-10pt}
% \captionsetup{font={small}}
\caption{Performance comparison on ScienceWorld environment: vanilla PPO, PPO with Entropy-based Advantage shaping (PPO+EA) from Cheng et al.~\citep{cheng2025reasoning}, and our PPO with Entropy-smoothed Policy Optimization (PPO+EPO). Results show episodic rewards (a,b), validation IID success rates (c,d), and OOD success rates (e,f). PPO+EPO consistently outperforms both baselines, achieving near-perfect success rates ($\sim$1.0) compared to PPO+EA's plateau at 0.5-0.6. Curves show mean values with shaded standard error across multiple seeds.}
\label{app:fig:model_study:ea}
\vspace{-10pt}
\end{figure*}

We compare our Entropy-smoothed Policy Optimization (EPO) with the Entropy-based Advantage (EA) shaping method from Cheng et al.~\citep{cheng2025reasoning}. As shown in ~\autoref{app:fig:model_study:ea}, while PPO+EA improves over the baseline, our PPO+EPO is substantially superior in both final performance and convergence speed.

The primary difference lies in the gradient signal. The EA method uses a detached entropy term , which acts as an indirect intrinsic reward rather than a direct, optimizable objective. Consequently, the policy receives no gradient signal to explicitly increase its entropy. In contrast, our EPO formulation integrates entropy directly into the policy loss, enabling a direct gradient $\nabla_\theta L^H(\theta)$ to explicitly guide the policy towards more exploratory behavior. Furthermore, EA's hard clipping on the advantage bonus can induce training instability, and its myopic nature considers only instantaneous entropy. Our EPO method promotes smoother and more consistent updates by using a continuous smoothing regularizer that leverages a historical entropy window. This temporal consistency is critical for long-horizon reasoning tasks.

These theoretical advantages explain the empirical gap: PPO+EPO converges to a near-optimal success rate of almost 1.0, while PPO+EA plateaus far lower at 0.5-0.6. We posit that EA's direct advantage modification distorts the credit assignment process. In contrast, EPO's decoupled regularization preserves the integrity of the value signal, leading to more robust and effective learning.

\clearpage

\section{Core Implementation Pseudocode}\label{app:pseudocode}
\autoref{alg:pytorch_impl} presents a PyTorch-style pseudocode outlining the core computational steps for our proposed Entropy-Smoothed Policy Optimization (EPO) loss. This implementation directly corresponds to the methodology described in Section 3, detailing how the base policy loss, entropy regularization, and the entropy smoothing regularizer are combined to form the final training objective for a single policy update step.

\begin{algorithm}[H]
    \caption{PyTorch-style Pseudocode for EPO Loss Calculation}
    \label{alg:pytorch_impl}
    \begin{lstlisting}[style=mystyle]
# Given: policy `policy_pi`, batch `data`, current epoch `k`
# `data` contains: old_log_probs, advantages, response_mask, entropy_history
# Hyperparameters: lambda_, kappa_l, kappa_r, alpha, K

# 1. Forward pass for current log probabilities and token-level entropy
logits = policy_pi(data.input_ids, data.attention_mask)
log_prob, entropy = get_logprob_and_entropy(logits, data.responses)

# 2. Compute the base multi-turn policy loss L^MT (e.g., PPO objective)
pg_loss = compute_policy_loss(
    log_prob=log_prob,
    old_log_prob=data.old_log_probs,
    advantages=data.advantages,
    response_mask=data.response_mask,
    clip_ratio=0.2
)

# 3. Compute the entropy regularization loss L^H (Eq. 6)
entropy_loss = agg_loss(entropy, data.response_mask)

# 4. Compute the entropy smoothing regularizer
# 4a. Calculate historical average entropy from window W_k
historical_avg_entropy = data.entropy_history.mean()

# 4b. Generate token-wise penalty mask P based on historical avg (Eq. 8)
penalty_mask = generate_entropy_penalty(
    current_entropy=entropy,
    historical_avg_entropy=historical_avg_entropy,
    min_ratio=kappa_l, max_ratio=kappa_r, penalty_weight=alpha
)
    
# 4c. Calculate smoothing loss L^smooth by aggregating penalties (Eq. 9)
smoothing_loss = agg_loss(penalty_mask, data.response_mask)

# 4d. Get dynamic coefficient beta_k for the current step k (Eq. 11)
beta_k = calculate_dynamic_beta(current_step=k, total_steps=K)
    
# 5. Combine entropy and smoothing terms
# Corresponds to [L^H(theta) - beta_k * L^smooth(theta)]
entropy_term = entropy_loss - beta_k * smoothing_loss

# 6. Compute the final EPO loss (Eq. 10)
# L^EPO = L^MT - lambda * [L^H - beta_k * L^smooth]
final_loss = pg_loss - lambda_ * entropy_term
    \end{lstlisting}
\end{algorithm}

\clearpage

\section{System Prompts}~\label{app:system prompts}

This appendix details the system prompts used to guide the language model agents in the ALFWorld and ScienceWorld environments. For each environment, we provide two versions of the prompt: one that includes historical context (previous actions and observations) and one that omits it for the initial turn. The placeholders in curly braces, such as \texttt{\textcolor{placeholderred}{\{current\_observation\}}}, are dynamically replaced with environment-specific information at runtime.

\subsection{ALFWorld Prompts}

\begin{lstlisting}[language=, caption={ALFWorld prompt (without history)}, label=lst:alfworld_no_his, style=promptstyle]
You are an expert agent operating in the ALFRED Embodied Environment.
Your current observation is: {current_observation}
Your admissible actions of the current situation are: [{admissible_actions}].

Now it's your turn to take an action.
You should first reason step-by-step about the current situation. This reasoning process MUST be enclosed within <think> </think> tags. 
Once you've finished your reasoning, you should choose an admissible action for current step and present it within <action> </action> tags.
\end{lstlisting}

\begin{lstlisting}[language=, caption={ALFWorld prompt (with history)}, label=lst:alfworld_his, style=promptstyle]
You are an expert agent operating in the ALFRED Embodied Environment. Your task is to: {task_description}
Prior to this step, you have already taken {step_count} step(s). Below are the most recent {history_length} observations and the corresponding actions you took: {action_history}
You are now at step {current_step} and your current observation is: {current_observation}
Your admissible actions of the current situation are: [{admissible_actions}].

Now it's your turn to take an action.
You should first reason step-by-step about the current situation. This reasoning process MUST be enclosed within <think> </think> tags. 
Once you've finished your reasoning, you should choose an admissible action for current step and present it within <action> </action> tags.
\end{lstlisting}

\subsection{ScienceWorld Prompts}

\begin{lstlisting}[language=, caption={ScienceWorld prompt (without history)}, label=lst:sciworld_no_his, style=promptstyle]
You are an expert agent operating in the ScienceWorld environment, which is a text-based virtual environment centered around accomplishing tasks from the elementary science curriculum.
Your current task is: {task_description}

Your current observation is: {current_observation}
Here are the actions you may take:
[
{"action": "open OBJ", "description": "open a container"},
{"action": "close OBJ", "description": "close a container"},
{"action": "activate OBJ", "description": "activate a device"},
{"action": "deactivate OBJ", "description": "deactivate a device"},
{"action": "connect OBJ to OBJ", "description": "connect electrical components"},
{"action": "disconnect OBJ", "description": "disconnect electrical components"},
{"action": "use OBJ [on OBJ]", "description": "use a device/item"},
{"action": "look around", "description": "describe the current room"},
{"action": "look at OBJ", "description": "describe an object in detail"},
{"action": "look in OBJ", "description": "describe a container's contents"},
{"action": "read OBJ", "description": "read a note or book"},
{"action": "move OBJ to OBJ", "description": "move an object to a container"},
{"action": "pick up OBJ", "description": "move an object to the inventory"},
{"action": "put down OBJ", "description": "drop an inventory item"},
{"action": "pour OBJ into OBJ", "description": "pour a liquid into a container"},
{"action": "dunk OBJ into OBJ", "description": "dunk a container into a liquid"},
{"action": "mix OBJ", "description": "chemically mix a container"},
{"action": "go to LOC", "description": "move to a new location"},
{"action": "eat OBJ", "description": "eat a food"},
{"action": "flush OBJ", "description": "flush a toilet"},
{"action": "focus on OBJ", "description": "signal intent on a task object"},
{"action": "wait", "description": "take no action for 10 iterations"},
{"action": "wait1", "description": "take no action for 1 iteration"},
{"action": "task", "description": "describe current task"},
{"action": "inventory", "description": "list your inventory"}
]

Current available actions:
{available_actions}

Now it's your turn to take an action.
You should first reason step-by-step about the current situation. This reasoning process MUST be enclosed within <think> </think> tags. 
Once you've finished your reasoning, you should choose an appropriate action for the current step and present it within <action> </action> tags.
\end{lstlisting}

\begin{lstlisting}[language=, caption={ScienceWorld prompt (with history)}, label=lst:sciworld_his, style=promptstyle]
You are an expert agent operating in the ScienceWorld environment, which is a text-based virtual environment centered around accomplishing tasks from the elementary science curriculum.
Your current task is: {task_description}

Prior to this step, you have already taken {step_count} step(s). Below are the most recent {history_length} observations and the corresponding actions you took: {action_history}
You are now at step {current_step} and your current observation is: {current_observation}
Here are the actions you may take:
[
{"action": "open OBJ", "description": "open a container"},
{"action": "close OBJ", "description": "close a container"},
{"action": "activate OBJ", "description": "activate a device"},
{"action": "deactivate OBJ", "description": "deactivate a device"},
{"action": "connect OBJ to OBJ", "description": "connect electrical components"},
{"action": "disconnect OBJ", "description": "disconnect electrical components"},
{"action": "use OBJ [on OBJ]", "description": "use a device/item"},
{"action": "look around", "description": "describe the current room"},
{"action": "look at OBJ", "description": "describe an object in detail"},
{"action": "look in OBJ", "description": "describe a container's contents"},
{"action": "read OBJ", "description": "read a note or book"},
{"action": "move OBJ to OBJ", "description": "move an object to a container"},
{"action": "pick up OBJ", "description": "move an object to the inventory"},
{"action": "put down OBJ", "description": "drop an inventory item"},
{"action": "pour OBJ into OBJ", "description": "pour a liquid into a container"},
{"action": "dunk OBJ into OBJ", "description": "dunk a container into a liquid"},
{"action": "mix OBJ", "description": "chemically mix a container"},
{"action": "go to LOC", "description": "move to a new location"},
{"action": "eat OBJ", "description": "eat a food"},
{"action": "flush OBJ", "description": "flush a toilet"},
{"action": "focus on OBJ", "description": "signal intent on a task object"},
{"action": "wait", "description": "take no action for 10 iterations"},
{"action": "wait1", "description": "take no action for 1 iteration"},
{"action": "task", "description": "describe current task"},
{"action": "inventory", "description": "list your inventory"}
]

Current available actions:
{available_actions}

Now it's your turn to take an action. You should first reason step-by-step about the current situation. This reasoning process MUST be enclosed within <think> </think> tags.
Once you've finished your reasoning, you should choose an appropriate action for the current step and present it within <action> </action> tags.
\end{lstlisting}

\section{Limitation and Future Work}
While EPO effectively addresses the exploration-exploitation cascade failure in multi-turn sparse-reward environments, our approach does not fully leverage memory systems~\citep{amem} to enhance learning from past trajectories. Currently, EPO uses historical entropy information solely for regularization, but does not incorporate explicit memory mechanisms that could help agents recall and reuse successful behavioral patterns from previous episodes. In multi-turn settings where sparse rewards make successful trajectories particularly valuable, a memory-augmented approach could potentially accelerate learning by allowing agents to explicitly store and retrieve relevant past experiences, especially those leading to rare positive rewards.

Future work could extend EPO to vision-language model (VLM) agents operating in multi-turn visual environments, where the cascade failure may manifest differently due to the multimodal nature of observations and actions. The interplay between visual and textual entropy in VLM agents presents unique challenges—visual observations might require different entropy bounds than textual responses, and the temporal dependencies across modalities could amplify or dampen the cascade failure. 

\section{Use of Large Language Models}
We utilized Large Language Models (LLMs), such as Claude, exclusively for ancillary support in two main areas: (i) language editing and polishing of the manuscript, and (ii) coding assistance for minor boilerplate tasks, such as generating plotting scripts and small utilities. All model-generated outputs were thoroughly reviewed, modified, and rigorously tested by the authors to ensure their accuracy and appropriateness.

The core intellectual contributions of this work—including all research ideas, algorithmic designs, experimental methodologies, data analysis, and conclusions—were conceived and validated entirely by the authors. Critically, LLMs were \textbf{not} used to generate any experimental results, create annotations or ground truth data, or influence methodological decisions. The authors assume full and sole responsibility for all content presented in this paper.

\end{document}